%% file: ms.tex
\documentclass[10pt,twocolumn,letterpaper]{article}
\pdfoutput=1

\usepackage{cvpr}
\usepackage{times}
\usepackage{epsfig}
\usepackage{graphicx}
\usepackage{amsmath}
\usepackage{amssymb}

\usepackage[breaklinks=true,bookmarks=false]{hyperref}
\cvprfinalcopy 
\def\cvprPaperID{2466} 

\ifcvprfinal\fi
\input{tex/preamble}
\input{tex/title}

\author{Yuxian Qiu$^\S$ \quad Jingwen Leng$^\S$\footnotemark \quad Cong Guo$^\S$\\Quan Chen$^\S$ \quad Chao Li$^\S$ \quad Minyi Guo$^\S$\footnotemark[\value{footnote}] \quad Yuhao Zhu$^\dagger$\\
$^\S$\textit{Department of Computer Science and Engineering, Shanghai Jiao Tong University}\\
$^\dagger$\textit{Department of Computer Science, University of Rochester}\\
{\tt\small \{qiuyuxian,leng-jw,guocong\}@sjtu.edu.cn,}\\{\tt\small \{chen-quan,lichao,guo-my\}@cs.sjtu.edu.cn, yzhu@rochester.edu}}

\begin{document}

\clearpage\maketitle
\thispagestyle{empty}

\footnotetext{$^*$ Jingwen Leng and Minyi Guo are co-corresponding authors of this paper.}

\input{tex/abstract}
\input{tex/introduction}
\input{tex/path}

\input{tex/differentiation}

\input{tex/defense}

\input{tex/evaluation}
\input{tex/related}

\input{tex/conclusion}
\input{tex/acknowledgment}

{\small
\bibliographystyle{ieee_fullname}
\bibliography{mybib}
}

\ifnum\pdfstrcmp{\jobname}{ms}=0
    \clearpage
    \appendix
    \input{tex/appendix}
\fi

\end{document}

%% file: tex/preamble.tex
\input{tex/math_commands.tex}

\usepackage[utf8]{inputenc} 
\usepackage[T1]{fontenc}    
\usepackage{url}            
\usepackage{booktabs}       
\usepackage{amsfonts}       
\usepackage{nicefrac}       
\usepackage{microtype}      
\usepackage{graphicx}
\usepackage{amsmath}
\usepackage{csvsimple}
\usepackage{subfig}
\usepackage{makecell} 
\usepackage{multirow}

\usepackage{xargs}                      
\usepackage{mathrsfs}

\newcommandx{\unsure}[2][1=]{\todo[linecolor=red,backgroundcolor=red!25,bordercolor=red,#1]{#2}}
\newcommandx{\change}[2][1=]{\todo[linecolor=blue,backgroundcolor=blue!25,bordercolor=blue,#1]{#2}}
\newcommandx{\info}[2][1=]{\todo[linecolor=OliveGreen,backgroundcolor=OliveGreen!25,bordercolor=OliveGreen,#1]{#2}}
\newcommandx{\improvement}[2][1=]{\todo[linecolor=Plum,backgroundcolor=Plum!25,bordercolor=Plum,#1]{#2}}
\newcommandx{\thiswillnotshow}[2][1=]{\todo[disable,#1]{#2}}

\newcommand{\Fig}[1]{Fig.~\ref{#1}}

\newcommand{\Tbl}[1]{Tbl.~\ref{#1}}
\newcommand{\Sec}[1]{Sec.~\ref{#1}}


%% file: tex/math_commands.tex
\usepackage{amsmath,amsfonts,bm}

\def\eqref#1{equation~\ref{#1}}

\def\1{\bm{1}}

\DeclareMathAlphabet{\mathsfit}{\encodingdefault}{\sfdefault}{m}{sl}
\SetMathAlphabet{\mathsfit}{bold}{\encodingdefault}{\sfdefault}{bx}{n}



%% file: tex/title.tex
\title{Adversarial Defense Through Network Profiling Based Path Extraction}



%% file: tex/abstract.tex
\begin{abstract}

    Recently, researchers have started decomposing deep neural network models according to their semantics or functions. Recent work has shown the effectiveness of decomposed functional blocks for defending adversarial attacks, which add small input perturbation to the input image to fool the DNN models. This work proposes a profiling-based method to decompose the DNN models to different functional blocks, which lead to the effective path as a new approach to exploring DNNs' internal organization. Specifically, the per-image effective path can be aggregated to the class-level effective path, through which we observe that adversarial images activate effective path different from normal images. We propose an effective path similarity-based method to detect adversarial images with an interpretable model, which achieve better accuracy and broader applicability than the state-of-the-art technique.





\end{abstract}

%% file: tex/introduction.tex
\section{Introduction}

Deep learning (DL) has revolutionized the key application domains such computer vision~\cite{DBLP:conf/nips/KrizhevskySH12}, natural-language processing~\cite{DBLP:conf/nips/SutskeverVL14}, and automatic speech recognition~\cite{AbdelHamid2014ConvolutionalNN}. DL models have outperformed traditional machine learning approaches and even outperformed human beings.
Although most of the current research efforts have been in improving the efficiency and accuracy of DL models, interpretability has recently become an increasingly important topic. This is because many DL-enabled or DL-based systems are mission-critical systems, such as ADAS~\cite{adas} and online banking systems~\cite{fiore2017using}. However, to date, there is no theoretical understanding of how DL models work, which is a significant roadblock in pushing DL into mission-critical systems.

Owing to the lack of interpretability, DL models usually do not have a clear decision boundary and are vulnerable to the input perturbation. Researches have recently been proposed~\cite{Pei:2017:DAW:3132747.3132785, DBLP:journals/corr/Moosavi-Dezfooli15,DBLP:journals/corr/KurakinGB16,DBLP:journals/corr/CarliniW16a}, which can all successfully find a small perturbation on the input image to fool the DNN based classifier. There is also prior work that demonstrates the physical attack feasibility by putting a printed image in front of a stop sign to mislead a real DNN based traffic sign detector~\cite{eykholt2018robust}. Last but not least, a DNN model often fails for inputs that are dramatically different from the training samples. For example, the classification model used in Tesla's autopilot system that incorrectly classified a white truck to cloud~\cite{tesla_crash} and caused the crash accident.

To address the vulnerability challenge in DL models, this work proposes the \emph{effective path} as a new approach to explore the internal organization of neural networks. The effective path for an image is a critical set of synapses and neurons that together lead to the final predicted class. The concept is similar to the execution path of a control-flow based program~\cite{Ball:1996:EPP:243846.243857}. We propose an activation based back-propagation algorithm to extract the image's effective path, which preserves the critical information in the neural network and allows us to analyze the inner structures of DNNs.

The derived per-image effective path has direct aggregation capability. For example, we get per-class effective path by aggregating the effective path from all training images in the same class.
We can then decompose the entire DNN into multiple components, each pertaining to an inference class.
We perform similarity analysis and find the phenomenon called \emph{path specialization} that different classes activate distinctive portions of the neural network in the inference task. On the basis of the observation, we analyze the path similarity between normal and adversarial images, we uncover that when an adversarial image successfully alters the prediction result by small perturbation, the network activates a significantly distinctive set of effective path compared to the training samples, which lays the foundation for defending the DNN using the effective path.

\begin{figure*}[t]
    \vspace*{-0.2cm}
    \centering
    \subfloat[Linear layer.]{
        \includegraphics[trim=0mm 0mm 0mm 0mm,clip,height=.12\linewidth]{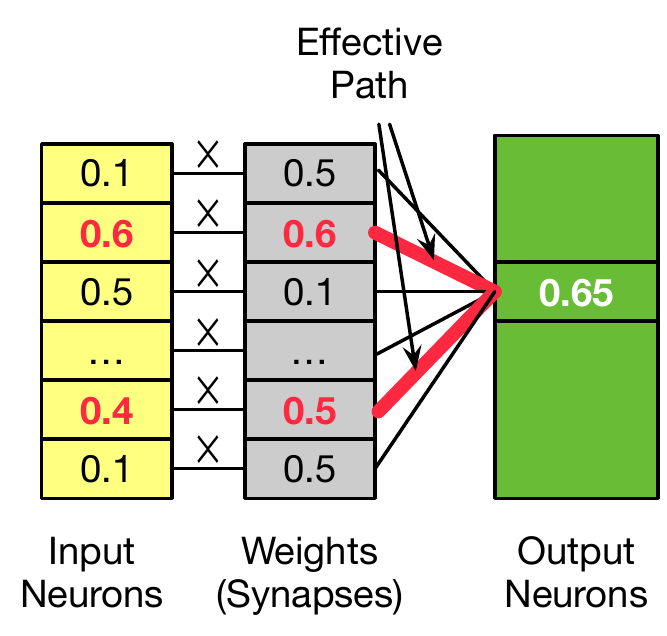}
        \label{fig:effective_path_example_linear}
    }%
    \hspace{.8cm}
    \subfloat[Convolutional layer.]{
        \includegraphics[trim=0mm 0mm 0mm 0mm,clip,height=.12\linewidth]{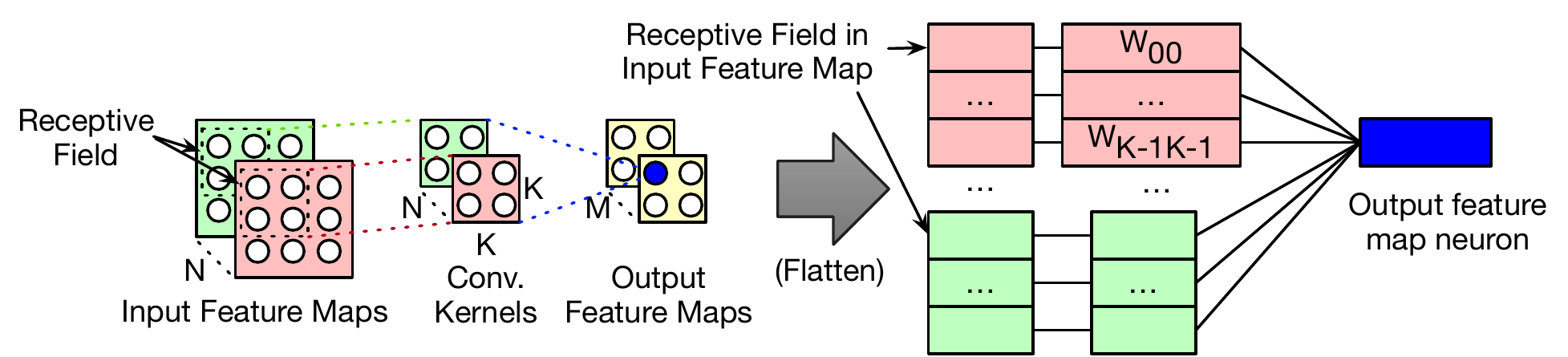}
        \label{fig:effective_path_example_conv}
    }
    \caption{Examples that illustrate the process of using profiling to extract effective path.}
    \vspace*{-0.4cm}
    \label{fig:effective_path_example}
\end{figure*}

We propose to use the simple linear combination of an image's per-layer effective path similarity to detect adversarial images.
Our work can use a simpler linear model to outperform the state-of-the-art work~\cite{wang2018interpret} for six representative attack methods.
Besides, we also show that our detection approach generalizes well to those attacks, meaning it can detect adversarial samples from methods that are not used in its training process, while the prior work does not have this level of generalization ability.
Moreover, the overhead of using our approach is also much smaller (up to 500 $\times$) than prior work.
In the end, we show that the effective path can not only be used for adversarial image detection but also for explaining the impact of the training process and network structure on the DNN's inference capability.





%% file: tex/path.tex
\section{Effective Path via Profiling}

Prior work~\cite{wang2018interpret} has proposed a method to extract critical data routing path (CDRP) for DNN models and demonstrated its usefulness of defending against adversarial samples.
However, deploying it in practice has two major limitations.
The first is the extraction process, which requires inserting control gates for each layer’s output channel and learning those gates through retraining.
This retraining process needs hyper-parameter tuning and takes a long time to process a single image.
The second disadvantage is the extracted path representation is still high dimensional (1152 for AlexNet and 15104 for ResNet-50).
The high dimensional representation weakens its interpretability and generalization ability to different adversarial attacks, which we will discuss later.

To overcome those limitations, we propose a novel method to extract the DNN's path information for an image. The method is inspired from path profiling in program analysis~\cite{Ball:1996:EPP:243846.243857}: a program is represented in the form of control flow graph, where a node is a basic block and an edge is the control flow between basic blocks. Compilers use path profiling to identify the sequences of frequently executed basic blocks, i.e., execution paths in the program. A program's path profiling provides useful insights on its execution and facilitates the understanding of the program.
In this work, we treat a neural network as a dataflow graph where a node is a neuron and an edge is a synapse (weight) between two neurons, and apply the profiling technique to extract its execution path, which we call as \textbf{\emph{effective path}} to distinguish from the prior work.
In the high level, both them represent the critical dataflow inside a DNN but our method doesn't rely on retraining and the derived representation is low-dimensional and generic.

\subsection{Single Image Extraction}
\label{sec:single-image-extraction}
We first explain how to extract the effective path for a single image,  denoted as $\mathcal{P} = (\mathcal{N}, \mathcal{S}, \mathcal{W})$, which represents the collection of critical neurons $\mathcal{N}$, synapses $\mathcal{S}$ and weights $\mathcal{W}$. It can be further broken down to the per-layer form $\mathcal{N} = (\mathcal{N}^{1},\ldots, \ \mathcal{N}^{L}), \  \mathcal{S} = (\mathcal{S}^{1},\ldots,\mathcal{S}^{L}), \mathcal{W} = (\mathcal{W}^{1},\ldots,\mathcal{W}^{L})$,  where  $\mathcal{N}^{l}$ represents the important output neurons of layer $l$, while $\mathcal{S}^{l}$ and $\mathcal{W}^{l}$ represent important synapses and weights.

The extraction process starts at the last layer $L$ and moves backward to the first layer. In the last layer $L$, only the neuron corresponding to the predicted class $n_{p}^{L}$ is active and thus is included in the effective path, i.e., $\mathcal{N}^{L} = \{n_{p} ^{L}\}$. The important weights form the minimum set of weights that can contribute more than $\theta$ ratio of the output neuron $n_{p}^{L}$. Equation~\ref{eq:eq1} formalizes the process, where $\tilde{K}_{p}^{L}$ is a selected set of weight indices with pre-nonlinearity neuron $n_{p}^{L}$ as the output, $w_{k,p}^{L}$ is the weight value, and $n_{k}^{L-1}$ is the corresponding input neuron value (also the output neuron of layer $l-1$). To find the minimum $\tilde{K}_{p}^{L}$, we can rank the weight and input neuron pairs by the value of their product and choose the minimum number of pairs that contribute to more than threshold $\theta \times n_{p} ^{L}$.

{
\vspace*{-0.2cm}
\begin{align}
    \begin{split}
        & \min_{\tilde{K}_{p}^{L}} |\tilde{K}_{p}^{L}|, s.t. \sum_{k \in \tilde{K}_{p}^{L}} n_{k}^{L-1} \times w_{k,p}^{L}  \geq  \theta \times n_{p} ^{L}
        \label{eq:eq1}
    \end{split}                                          \\
     & \mathcal{W}^{L} = \{w_{k,p}^{L} | k \in \tilde{K}_{p}^{L} \}
    \label{eq:eq2}                                                     \\
     & \mathcal{N}^{L-1} =  \{n_{k}^{L-1} | k \in \tilde{K}_{p}^{L} \}
    \label{eq:eq22}
\end{align}
}

After deriving the weight indices set $\tilde{K}_{p}^{L}$, we can get the $\mathcal{W}^{L}$ set using  Equation~\ref{eq:eq2}. Since the last layer is the fully connected layer and there is a one-to-one mapping between weight and synapses, $\mathcal{S}^{L}$ can also be derived. Meanwhile, since the output neurons of layer $L-1$ are the input neurons of layer $L$, it is straightforward to derive $\mathcal{N}^{L-1}$ in Equation~\ref{eq:eq22}. We then can repeat the process in Equation~\ref{eq:eq1} for every active neuron in $\mathcal{N}^{L-1}$: each active neuron will result in a set of weights and their union form the  $\mathcal{W}^{L-1}$. The process repeats backward until the first layer, and yields the whole neuron set $\mathcal{N}$, synapse set $\mathcal{S}$, and weight set $\mathcal{W}$ for the input image.

\begin{figure}[t]
    \begin{minipage}[c]{0.49\linewidth}
        \centering
        \includegraphics[trim=0mm 2mm 0mm 0mm,clip,width=\columnwidth]{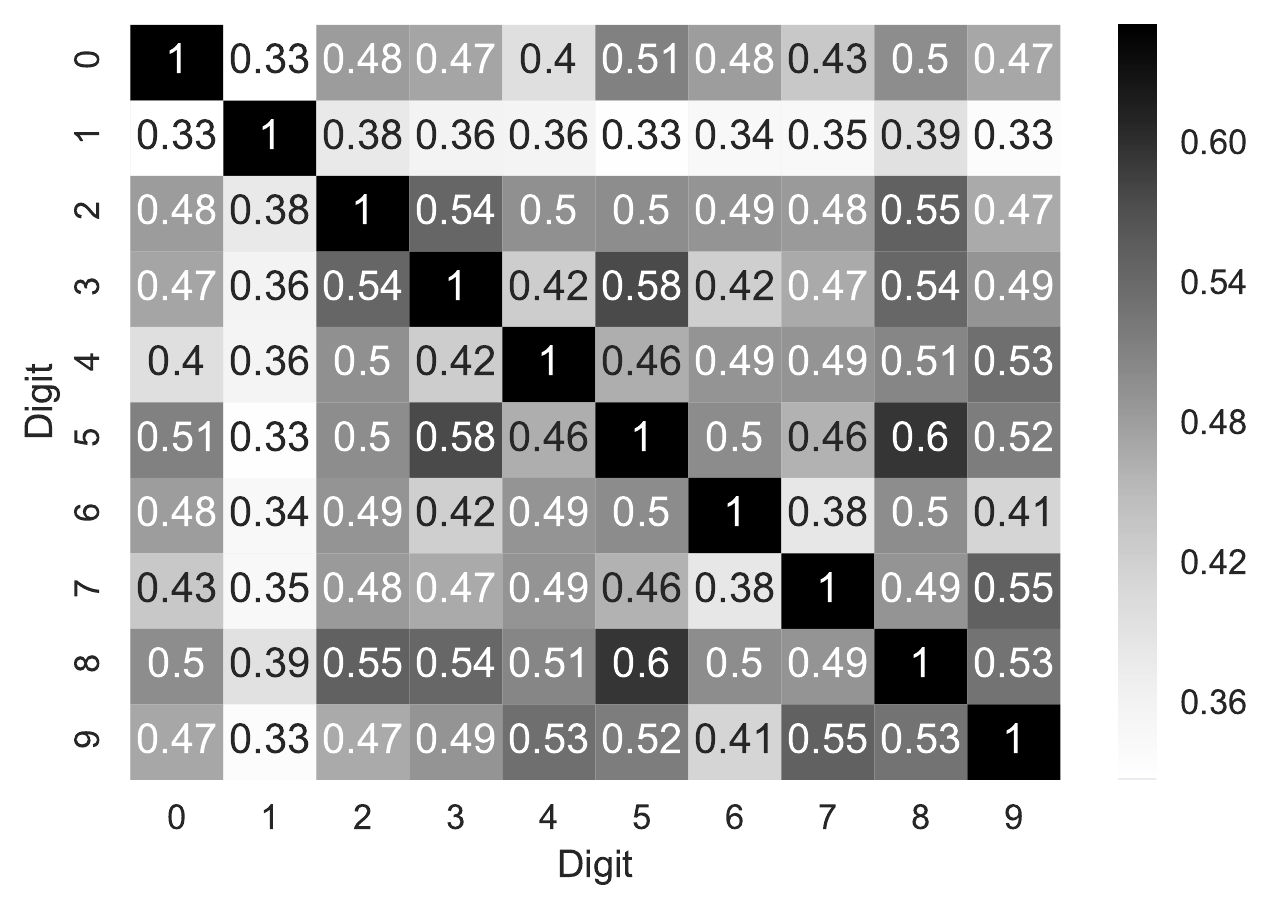}
        \vspace*{-0.3cm}
        \caption{Class-wise path similarity in LeNet.}
        \label{fig:last_layer_trace}
    \end{minipage}\hfill
    \begin{minipage}[c]{0.49\linewidth}
        \centering
        \includegraphics[trim=0mm 2mm 0mm 0mm,clip,width=\columnwidth]{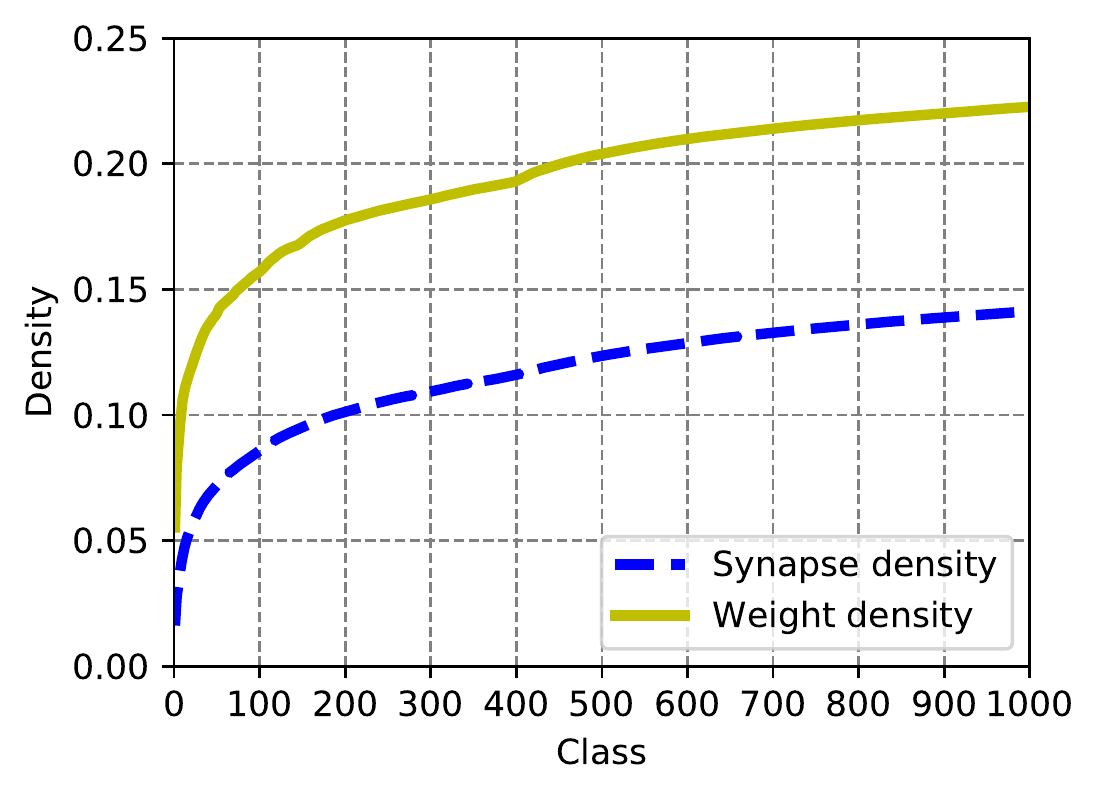}
        \vspace*{-0.3cm}
        \caption{Density growth when merging per-class path.}
        \label{fig:trace_growth}
    \end{minipage}
    \vspace*{-0.4cm}
\end{figure}

Note that the above process addresses the fully connected layer.
To process the convolutional layer, we need to convert it to FC layer according to each output neuron's receptive field as \Fig{fig:effective_path_example_conv} shows. There are two caveats on handling the convolutional layer. First, solving of Equation~\ref{eq:eq1} does not require the ranking of all input neurons but only the neurons in the receptive field of the output neuron. Second, there is no one-to-one mapping between synapse and weight because of weight sharing. As a result, multiple synapses can have the same active weights in the effective path.

\subsection{Multi-Image Aggregation}

The derived effective path is a binary mask that indicates whether a neuron or synapse contributes to the image inference.
As such, we can simply aggregate effective paths from an image group, e.g. images of the same class, to obtain a larger effective path that provides a higher level perspective of the whole group.
Aggregating the effective path of two images $\mathcal{P}(i)$ and $\mathcal{P}(j)$ is essentially taking the union of $\mathcal{N}$, $\mathcal{S}$ and $\mathcal{W}$ on each layer, represented by $
    \mathcal{P}(i) \cup \mathcal{P}(j) = (
    \mathcal{N}(i) \cup \mathcal{N}(j),
    \mathcal{S}(i) \cup \mathcal{S}(j),
    \mathcal{W}(i) \cup \mathcal{W}(j)
    )
$
, where $\mathcal{N}(i) \cup \mathcal{N}(j) = ( \mathcal{N}^{1}(i) \cup \mathcal{N}^{1}(j), \ldots, \mathcal{N}^{L}(i) \cup \mathcal{N}^{L}(j) )$  ($\mathcal{N}$ and $\mathcal{W}$ are similar).
This approach can create a meaningful representation for an image group without increasing its dimension.
In contrast, the feature dimension of CDRPs~\cite{wang2018interpret} increase linearly with the number of images in the class because each element in CDRP is a continuous number between 0 and 10 instead of a binary mask.


In this work, we use two types of aggregated effective path for the neural network interpretation and defense. For the class-level perspective, we aggregate all correctly predicted training images from the class $c$, denoted by $\tilde{\mathcal{X}}_{c}$, to get the \textbf{\emph {per-class effective path}} $\tilde{\mathcal{P}}_{c} = \bigcup_{x \in \tilde{\mathcal{X}}_{c}} \mathcal{P}(x)$; for the network-level perspective, we aggregate images from the whole training set $\tilde{\mathcal{X}}$ to get the \textbf{\emph{overall effective path}} $\tilde{\mathcal{P}} = \bigcup_{x \in \tilde{\mathcal{X}}} \mathcal{P}(x)$.

\vspace*{-0.2cm}
\paragraph{Path Sparsity}
The derived overall effective path is highly sparse compared to the full model, indicating that critical information is reserved.
We define the weight (synapse) density of the effective path $\mathcal{D}_{\mathcal{W}}$ ($\mathcal{D}_{\mathcal{S}}$) as the ratio of its weights (synapses) over the entire weights (synapses). They can be calculated in Equation~\ref{eq:eq_density}, where $\mathbb{W}^{l}$ and $\tilde{\mathcal{W}}^{l}$  ($\mathbb{S}^{l}$ and $\tilde{\mathcal{S}}^{l}$) is the layer $l$'s entire weight (synapse) set and weight (synapse) set in overall effective path, respectively.

{
\vspace*{-0.4cm}
\begin{align}
     & \mathcal{D}_{\mathcal{W}} = \frac{\sum_{l=1}^{L} \left | \tilde{\mathcal{W}}^{l} \right | }{\sum_{l=1}^{L} \left | \mathbb{W}^{l} \right | }, \mathcal{D}_{\mathcal{S}} = \frac{\sum_{l=1}^{L} \left | \tilde{\mathcal{S}}^{l} \right | }{\sum_{l=1}^{L} \left | \mathbb{S}^{l} \right | }
    \label{eq:eq_density}
\end{align}
\vspace*{-0.2cm}
}

We extracted the overall effective path for popular DNN models including LeNet-5~\cite{DBLP:journals/neco/LeCunBDHHHJ89}, AlexNet~\cite{Krizhevsky:2012:ICD:2999134.2999257}, ResNet-50~\cite{DBLP:journals/corr/HeZRS15}, Inception-v4~\cite{DBLP:journals/corr/SzegedyIV16}, and VGG-16~\cite{DBLP:journals/corr/SimonyanZ14a}. With $\theta = 0.5$, their synapse densities are 13.8\%, 20.5\%, 22.2\%, 41.7\%, 17.2\%, respectively.
Note that those values are calculated after aggregating all training samples (i.e. overall effective path).
Prior work CDRP~\cite{wang2018interpret} reported similar sparsity values, which, however, was calculated based on an individual image because different images' CDRPs cannot be aggregated directly while effective path can.
We also conduct an experiment which shows the DNN accuracy drops immediately when we start deactivating portions of the effective path, which indicates that the extracted path is not only sparse but also representative.

%% file: tex/differentiation.tex
\section{Effective Path Visualization}
\label{sec:similarity}


The per-class path dissects the network to different components and can be used to understand why the neural network can distinguish different classes and study the impact of changing the network structure.
We perform the path similarity analysis among different classes, which leads to a finding called \textbf{\emph{path specialization}}.
Different classes activate not only sparse but also a distinctive set of neurons and synapses for the inference task.

We first study the similarity of per-class effective paths. The similarity between class $c_1$ and $c_2$ is calculated by the Jaccard coefficient of their synapse set as in Equation~\ref{eq:similarity}.

{
\vspace*{-0.2cm}
\begin{align}
    J_{c_1,c_2} = J(\tilde{\mathcal{S}}_{c_1}, \tilde{\mathcal{S}}_{c_2}) = \frac{ \left | \tilde{\mathcal{S}}_{c_1} \bigcap \tilde{\mathcal{S}}_{c_2} \right | }{ \left | \tilde{\mathcal{S}}_{c_1} \bigcup \tilde{\mathcal{S}}_{c_2} \right | }
    \label{eq:similarity}
\end{align}
}

\Fig{fig:last_layer_trace} shows the class-wise path similarity in LeNet, unveiling the existence of path specialization: the averaged similarity between two classes is low (around 0.5). On average, two classes activate about 50\% common paths, as well as 50\% distinctive paths.
We can also conjecture that the degree of path specialization reflects the visual similarity between the two classes.
For example, in \Fig{fig:last_layer_trace}, digit `1' has the highest degree of specialization (i.e., lowest path similarity against other digits): its average similarity with other classes is around 0.35 (compared to the 0.5 average value).
The reason is most likely attributed to its unique shape.
In contrast, digit `5' and `8' have the highest path similarity of 0.6, also likely owing to their similar shapes.

\begin{figure*}[t]
    \centering
    \includegraphics[width=.95\linewidth]{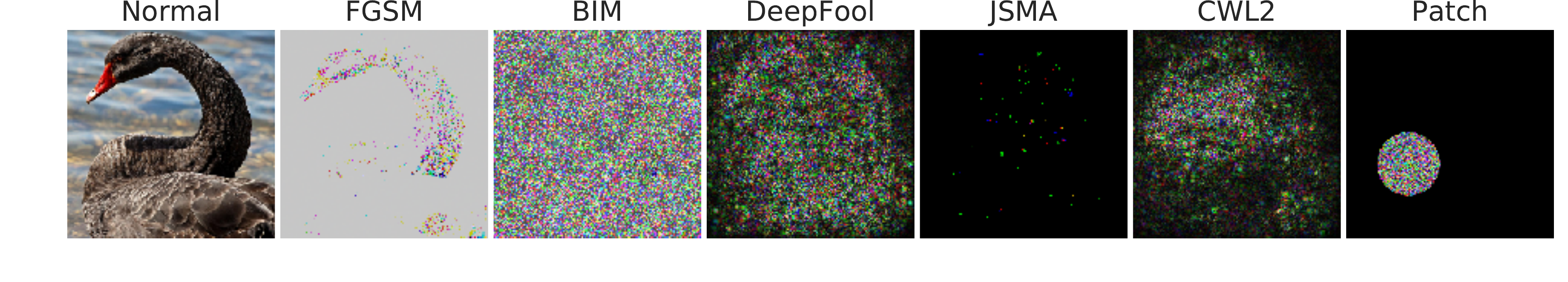}
    \vspace*{-0.2cm}
    \caption{Normal example and perturbations from different attacks. The perturbations are enhanced by 100 times to highlight the differences.}
    \label{fig:attack_examples}
    \vspace*{-0.2cm}
\end{figure*}


\begin{figure*}[t]
    \vspace*{-0.3cm}
    \centering
    \subfloat[]{
        \includegraphics[trim=0mm 0mm 0mm 0mm,clip,height=.2\linewidth]{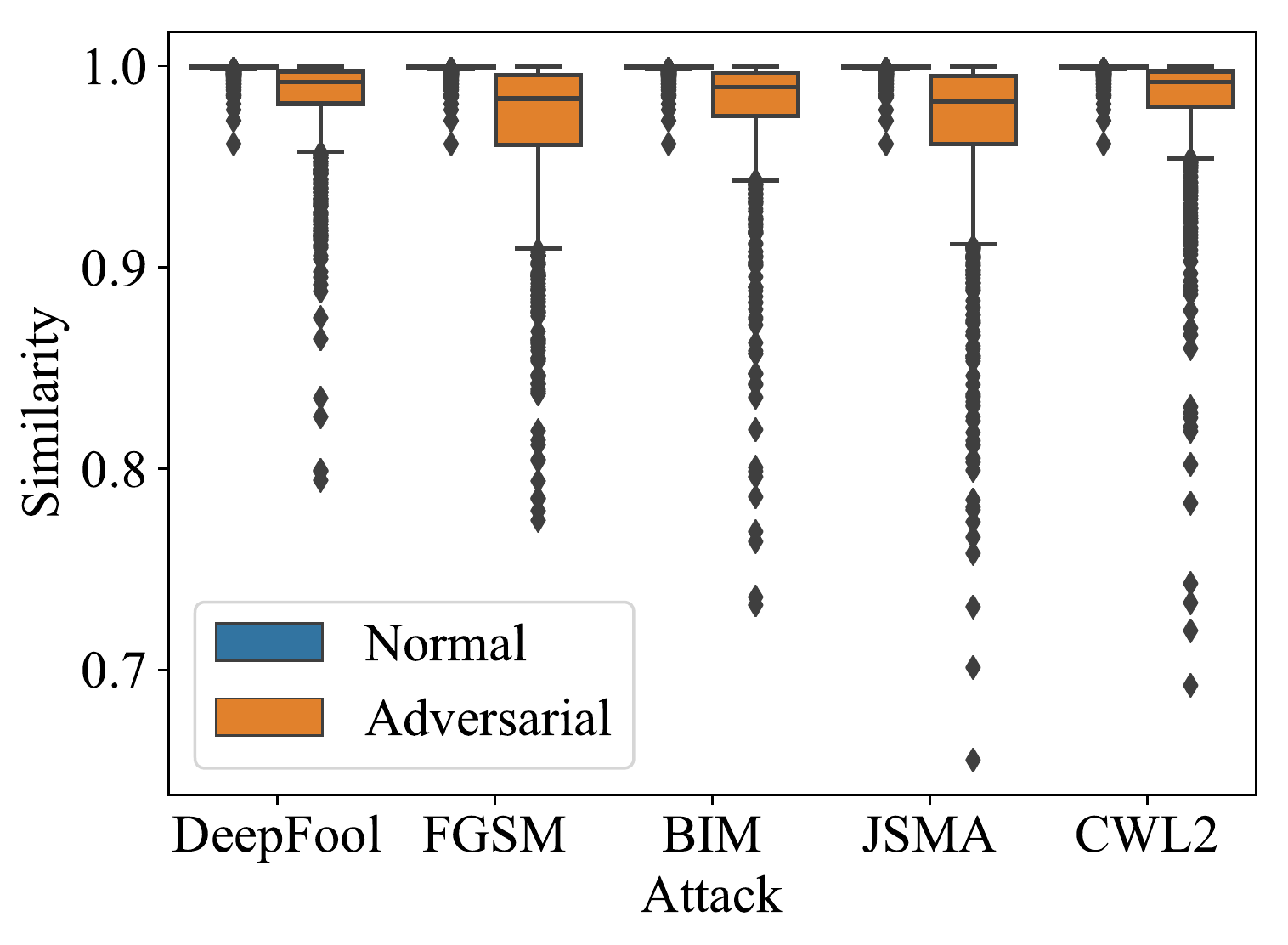}
        \label{fig:lenet_mnist_box_plot}
    }
    \subfloat[]{
    \includegraphics[trim=0mm 0mm 53mm 0mm,clip,height=.2\linewidth]{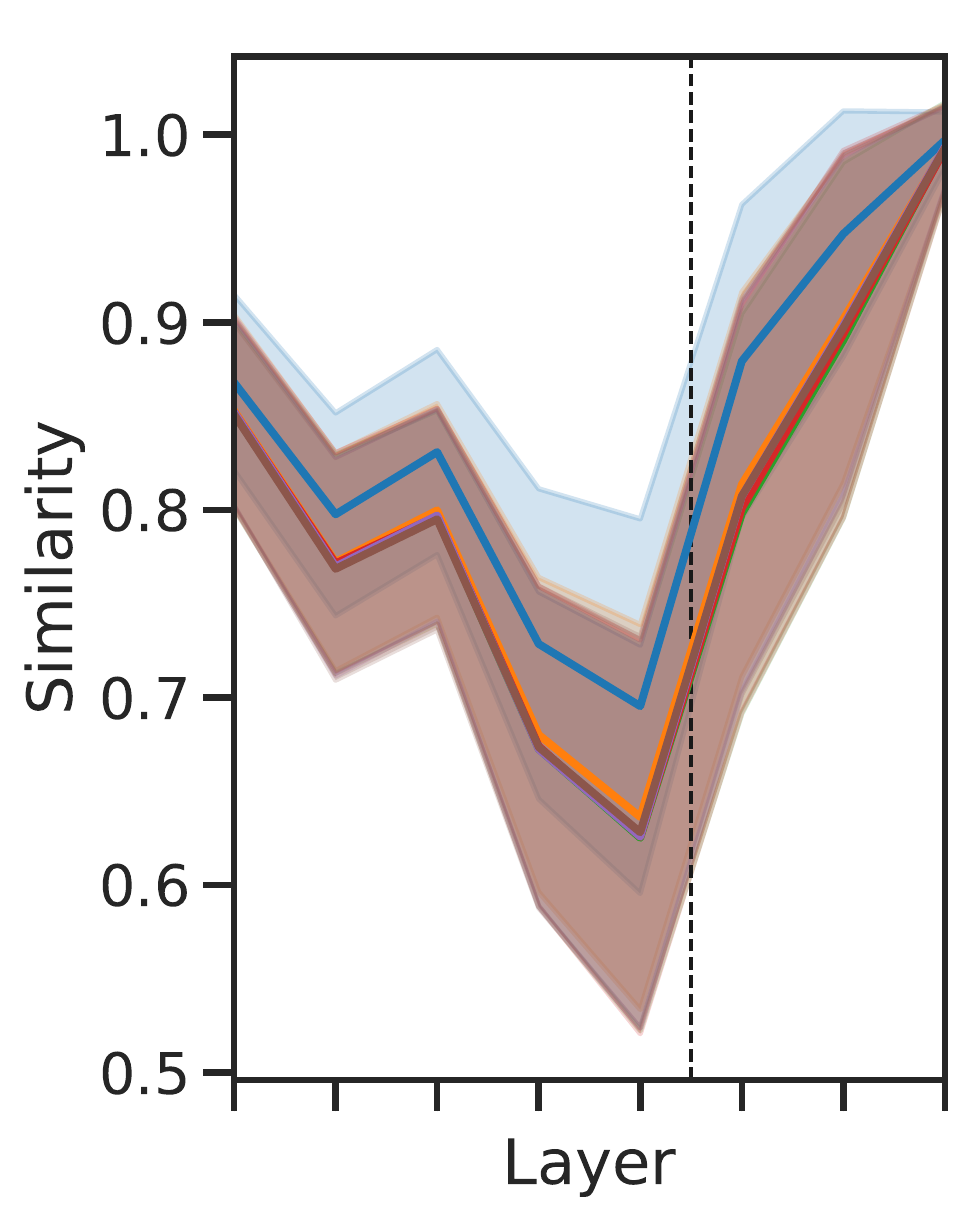}
    \label{fig:alexnet_similarity_rank1}
    }%
    \subfloat[]{
    \includegraphics[trim=0mm 0mm 53mm 0mm,clip,height=.2\linewidth]{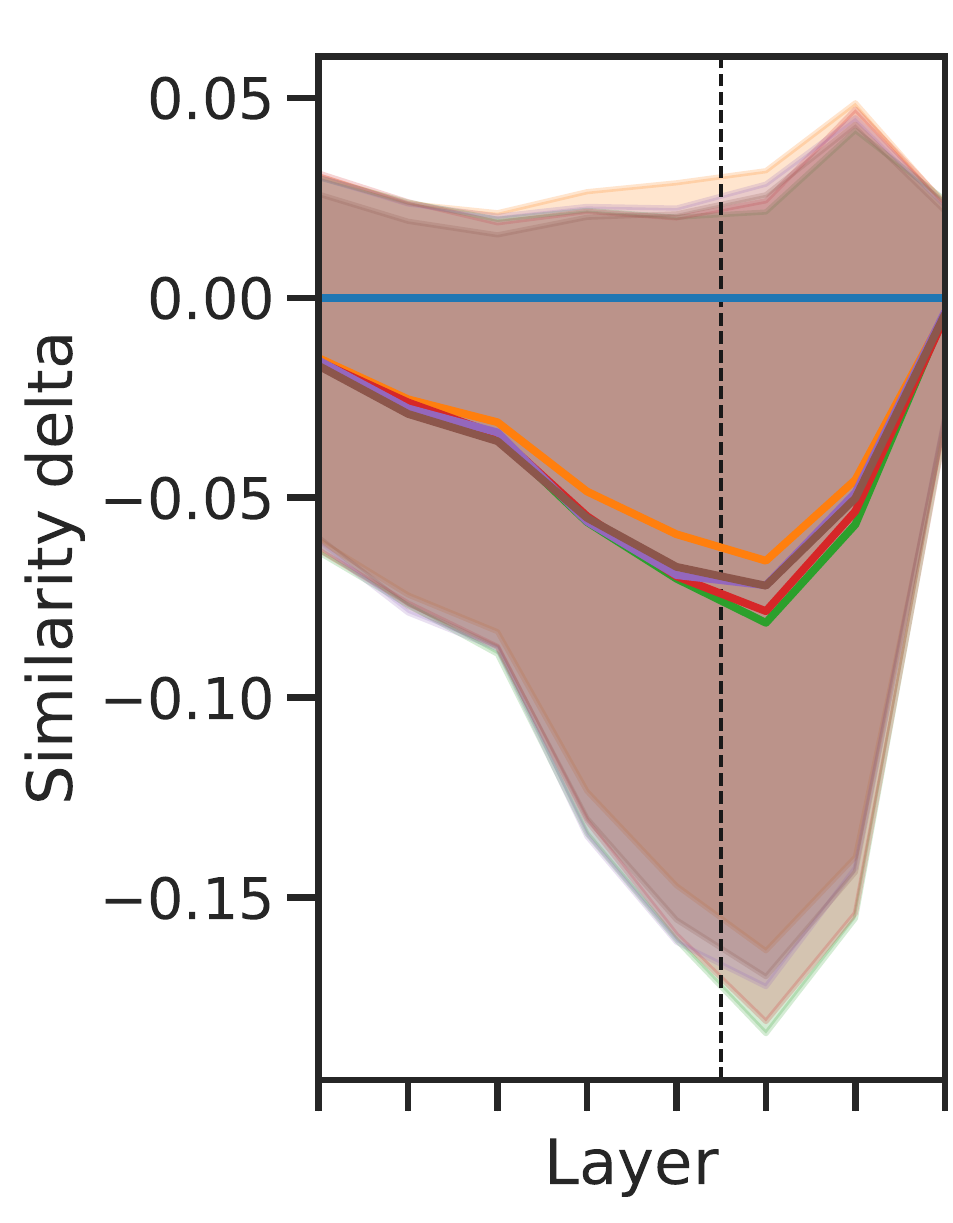}
    \label{fig:alexnet_similarity_delta_rank1}
    }%
    \subfloat[]{
    \includegraphics[trim=0mm 0mm 53mm 0mm,clip,height=.2\linewidth]{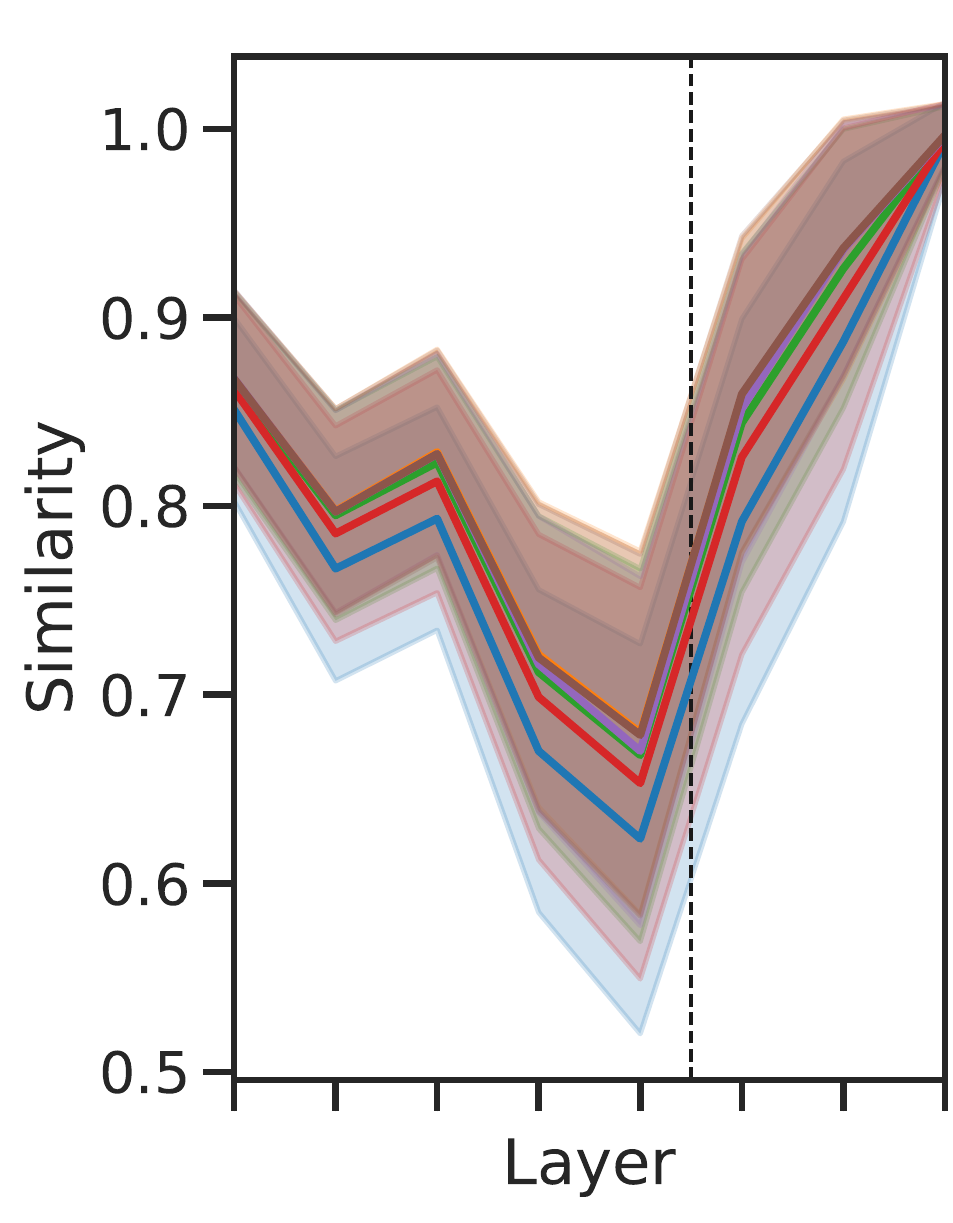}
    \label{fig:alexnet_similarity_rank2}
    }%
    \subfloat[]{
    \includegraphics[trim=0mm 0mm 0mm 0mm,clip,height=.2\linewidth]{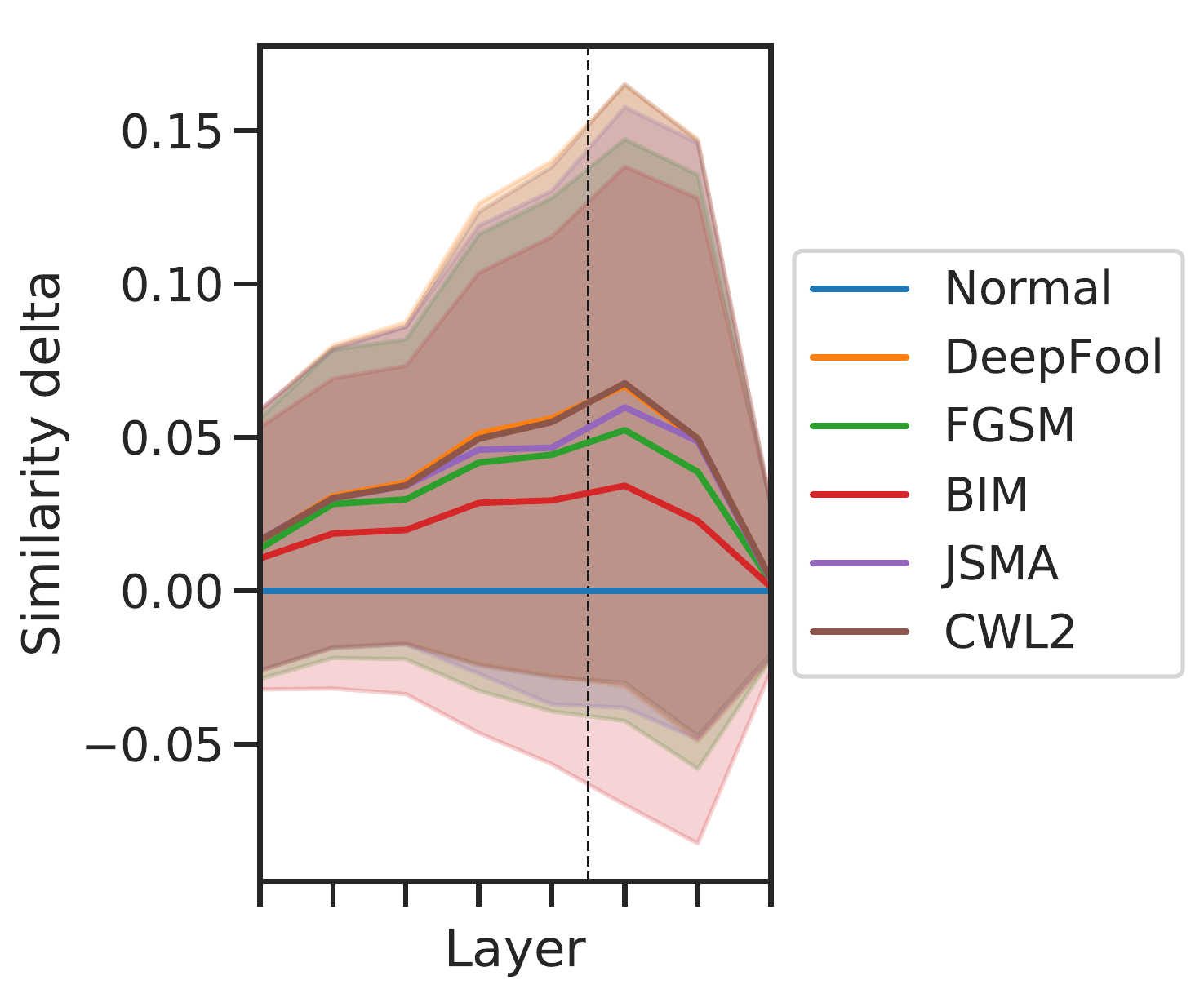}
    \label{fig:alexnet_similarity_delta_rank2}
    }%
    \vspace*{-0.3cm}
    \caption{(a) Path similarity for LeNet. (d-e): Distribution of per-layer similarity for AlexNet on ImageNet. Each line plot represents the mean of each kind of adversarial examples' similarity, with the same-color band around to show the standard deviation. The dashed line split convolutional layers and FC layers. (b): Rank-1 similarity. (c): Rank-1 similarity delta. (d): Rank-2 similarity. (e): Rank-2 similarity delta.}
    \label{fig:alexnet_similarity}
    \vspace*{-0.1cm}
\end{figure*}

We observe the existence of the path specialization in other datasets and networks. \Fig{fig:trace_growth} shows the path density growth when merging per-class (ImageNet) paths for ResNet-50. The growth of both weight and synapse follow the same trend (weight density is greater owing to weight sharing). The density increases rapidly initially, indicating the high degree of path specialization. After 50 classes, the density still increases but at a much slower pace. This matches the class hierarchy in the ImageNet dataset, which has around 100 basic categories:  different categories have a larger degree of path specialization while classes in the same categories have a smaller specialization degree.

In summary, we find the existence of path specialization phenomenon in trained DNNs, which unveils that DNNs activate different blocks when handling different classes.
Inspired by the observation, we study the possibility of using the effective path to detect adversarial samples.

%% file: tex/defense.tex
\section{Adversarial Samples Defense}
\label{sec:defense}

In this section, we study how to exploit the observed path specialization phenomenon to detect adversarial samples.
Adversarial samples are generated by adding a small perturbation to normal images.
The perturbation is small and imperceptible by human beings but can lead to an incorrect prediction of the neural network.
We evaluate 6 different attacks (i.e. methods to generate misleading perturbation for a given input image), whose examples are shown in \Fig{fig:attack_examples}.

For each attack, we always choose the canonical implementation. We use Foolbox~\cite{rauber2017foolbox} implementations and its default parameters in version 1.3.2 for Fast Gradient Sign Method (FGSM)~\cite{DBLP:journals/corr/GoodfellowSS14}, Basic Iterative Method(BIM)~\cite{DBLP:journals/corr/KurakinGB16}, DeepFool~\cite{DBLP:journals/corr/Moosavi-Dezfooli15}, Jacobian-based Saliency Map Attack(JSMA)~\cite{DBLP:journals/corr/PapernotMJFCS15}. For Carlini and Wagner(C\&W) attacks~\cite{DBLP:journals/corr/CarliniW16a}, we use the open-source code released by the paper authors. We use adversarial patch~\cite{DBLP:journals/corr/abs-1712-09665} implementation provided in CleverHans~\cite{papernot2018cleverhans} and extend it to support AlexNet without modification to its settings.

We first explore the distribution of effective path for normal and adversarial examples, and show that adversarial images activate distinctive effective path to fool the DNN.
Our further analysis indicates that effective path similarity provides a generic detection metric across all studied adversarial attacks.
Based on the analysis result, we propose a low-dimensional and uniform metric to detect adversarial samples from different kinds of attacks.


\subsection{Adversarial Samples Similarity Analysis}

On the basis of path specialization, we study the similarity of the effective path between normal images and adversarial images.
We introduce another similarity metric called \textbf{\emph{image-class path similarity}}, which indicates how many synapses in the image's effective path come from the predicted class's effective path. It can be calculated as $J_{\mathcal{P}} = J( \mathcal{S}, \mathcal{S} \cap \tilde{\mathcal{S}}_{p} ) = |\mathcal{S} \cap \tilde{\mathcal{S}}_{p}| / | \mathcal{S} |$, where $p$ is  the image's predicted class, $\mathcal{S}$ is the synapse set of image effective path, and $\tilde{\mathcal{S}}_{p}$ is the synapse set of class $p$'s effective path.
Because the per-class effective path is far larger than the image's effective path, their Jaccard coefficient will be nearly zero. As such, the image-class path similarity is essentially the Jaccard coefficient between the image's effective path and the intersection set of effective path between the image and predicted class.

\Fig{fig:lenet_mnist_box_plot} shows the distribution of image-class path similarity for both normal images and a rich set of adversarial images in MNIST for LeNet. The similarity values for normal images are almost all 1, and note that they are not used in the training and per-class path extraction. In contrast, the similarity values for adversarial images are mostly smaller than 1,  indicating effective path as a great metric to distinguish between normal and adversarial images.

For deeper and more complicated DNNs, we breakdown the image-class path similarity metric to different layers. It can be calculated as $J_{\mathcal{P}}^{l} = |\mathcal{S}^{l} \cap \tilde{\mathcal{S}}_{p}^{l}|/|\mathcal{S}^{l} |$ for layer $l$.
\Fig{fig:alexnet_similarity_rank1} compares the per-layer similarity for normal images (from test set) and adversarial images on AlexNet, and show that normal images demonstrate a higher similarity degree than adversarial images.
We further calculate the similarity delta, which equals to the similarity value of a normal image minus the similarity value of its corresponding adversarial image.
\Fig{fig:alexnet_similarity_delta_rank1} shows that all adversarial attacks cause almost identical similarity decrease pattern, where the largest decrease occurs in the middle layers, i.e. boundary between convolutional layers and fully connected layers.

Recall that we extract the effective path starting from the predicted class, i.e. rank-1 class, which we call rank-1 effective path. We also study the rank-2 effective path which starts from the rank-2 class.
\Fig{fig:alexnet_similarity_rank2} compares the rank-2 effective path similarity for normal and adversarial images. Different from the rank-1 effective path,  adversarial images demonstrate a higher similarity degree than normal images. The reason is that the predicted rank-2 class for an adversarial image is often the rank-1 class of its corresponding normal image (i.e. without adding the perturbation).
In comparison, the predicted rank-2 class for a normal image has no such relationship, and therefore has a lower degree of similarity than adversarial image.
Moreover, different adversarial attack methods cause a similar pattern as \Fig{fig:alexnet_similarity_delta_rank2} shows.

In summary,  extending class-wise path similarity to the image-class case opens the door of using effective path to detect adversarial images: mainstream adversarial attacks modify the normally inactive path to fool the DNN and their impact indicates a uniform pattern.
In the next subsection, we propose a simple and highly interpretable method to exploit these observations for detecting adversarial samples.



\subsection{Defense Model}
\label{sec:defense_model}

Based on the per-layer similarity analysis, we propose to use the rank-1 and rank-2 effective path similarity to detect adversarial samples.
We study four different detection models, including linear model, random forest, AdaBoost, and gradient boosting.
Among them, the linear model is the simplest one with the strongest interpretability.
As we will show later, the linear model also achieves similar accuracy of other more complex models, proving that the selected input features (effective path similarity values) are strong indicators for detecting adversarial images.

\vspace*{-0.5cm}
\paragraph{Linear Model}
For the linear model, we propose \textbf{\emph{jointed similarity}} as the defense metric. It can be calculated as $\tilde{J}_{\mathcal{P}} = \sum_{l=1}^{L} \omega^{l} J_{\mathcal{P}}^{l} - \sum_{l=1}^{L} {\omega^{l}}' {J_{\mathcal{P}}^{l}}'$, where $J_{\mathcal{P}}^{l}$ and ${J_{\mathcal{P}}^{l}}'$ are respectively rank-1 and rank-2 similarity for layer $l$, $\omega^{l}$ and ${\omega^{l}}'$ are their coefficients that satisfy $\omega^{l} \geq 0, {\omega^{l}}' \geq 0$.
The joint similarity reflects the low rank-1 similarity degree and high rank-2 similarity degree of adversarial images.  An image is detected as an adversarial image if its joint similarity is less than a threshold.
The simple linear model avoids overfitting and offers strong interpretability.

We use LeNet-5 on MNIST, AlexNet on ImageNet and ResNet-50 v2 on ImageNet for evaluation. For each dataset, effective path extraction is performed on the overall training set with $\theta = 0.5$. For each model, adversarial examples from all evaluated attacks are aggregated, shuffled and split into 10\% for the training of joint similarity's coefficients and 90\% for defense performance evaluation. Notice that we only generate adversarial examples for the first test image in each class of ImageNet due to the significant computational cost. The training of joint similarity's coefficients is performed by SGD running 10000 epochs, with elastic-net regularization ($l_1$ ratio is 0.5) for sparsity.

\vspace*{-0.5cm}
\paragraph{Other Models}
We also study to use other more complex models, including random forest, AdaBoost, and gradient boosting.
These three approaches are also used to construct models based on CDRPs~\cite{wang2018interpret}.
However, our input features for those approaches are a vector formed by each layer's rank-1 and rank-2 effective path similarity while prior work's input features have a much larger dimension (e.g. 1152 for AlexNet and 15104 for ResNet-50). For sake of consistency, we apply the same adversarial example preprocessing with the linear model. We use 100 estimators for random forest and gradient boosting, while AdaBoost is limited to 50 estimators. All unmentioned configurations of these models stay the same with the default values in scikit-learn v0.19.2.



%% file: tex/evaluation.tex
\section{Evaluation}

In this section, we evaluate the adversarial sample detection accuracy based on effective path.
We first focus on the highly interpretable linear detection model and show its detection performance on a wide range of different attacks, datasets, and models.
We then compare our approach with prior work CDRP~\cite{wang2018interpret} and show that our approach achieves better accuracy, requires less training samples, and generalizes well to different types of adversarial attacks.



\begin{figure}[t]
\vspace*{-0.3cm}
\centering

\begingroup
\captionsetup[subfigure]{width=.33\linewidth}
\subfloat[LeNet.]{
    \includegraphics[width=.33\linewidth]{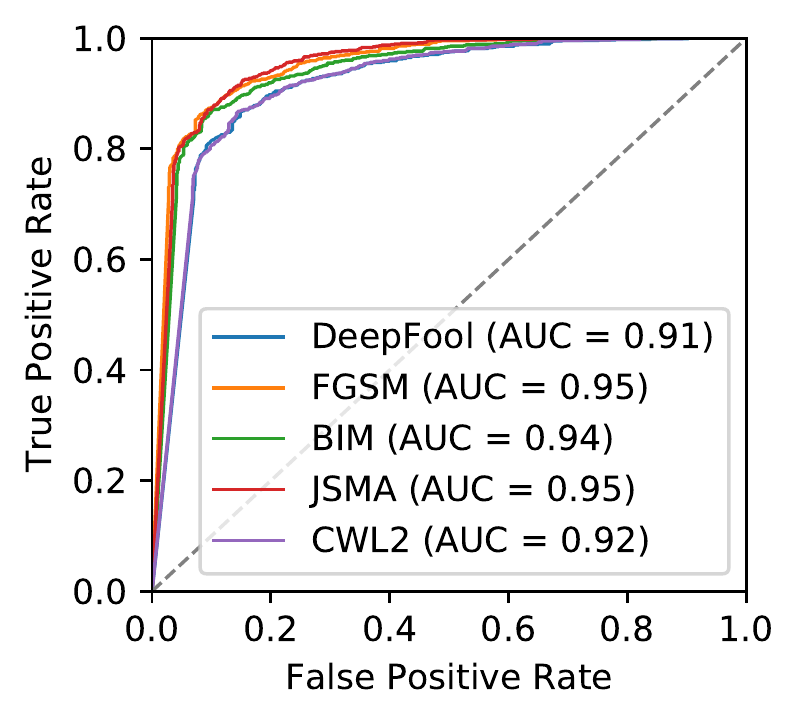}
    \label{fig:lenet_mnist_roc}
}
\subfloat[AlexNet.]{
\includegraphics[width=.33\linewidth]{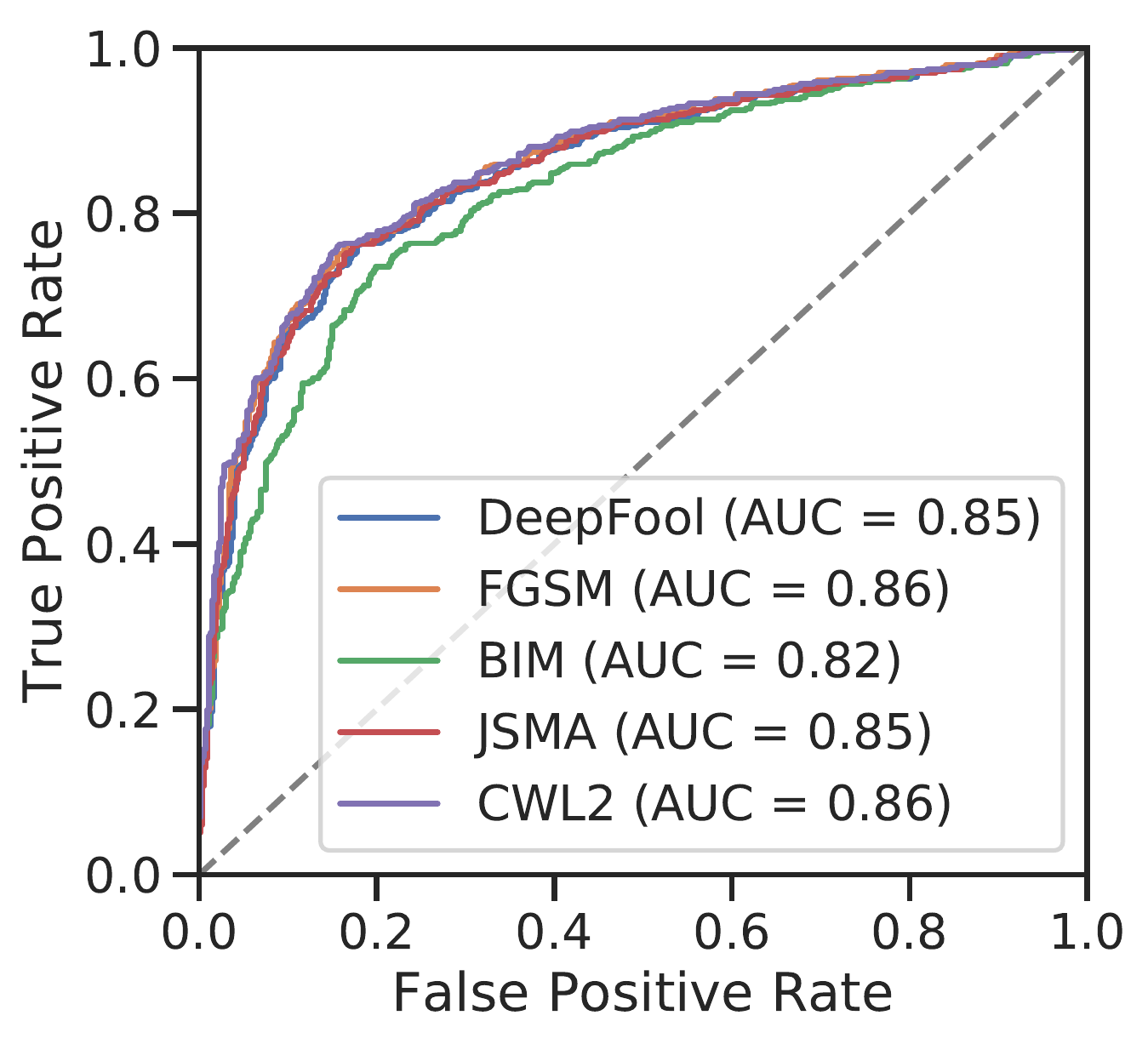}
\label{fig:alexnet_roc_basic_u}
}%
\subfloat[ResNet-50.]{
\includegraphics[width=.33\linewidth]{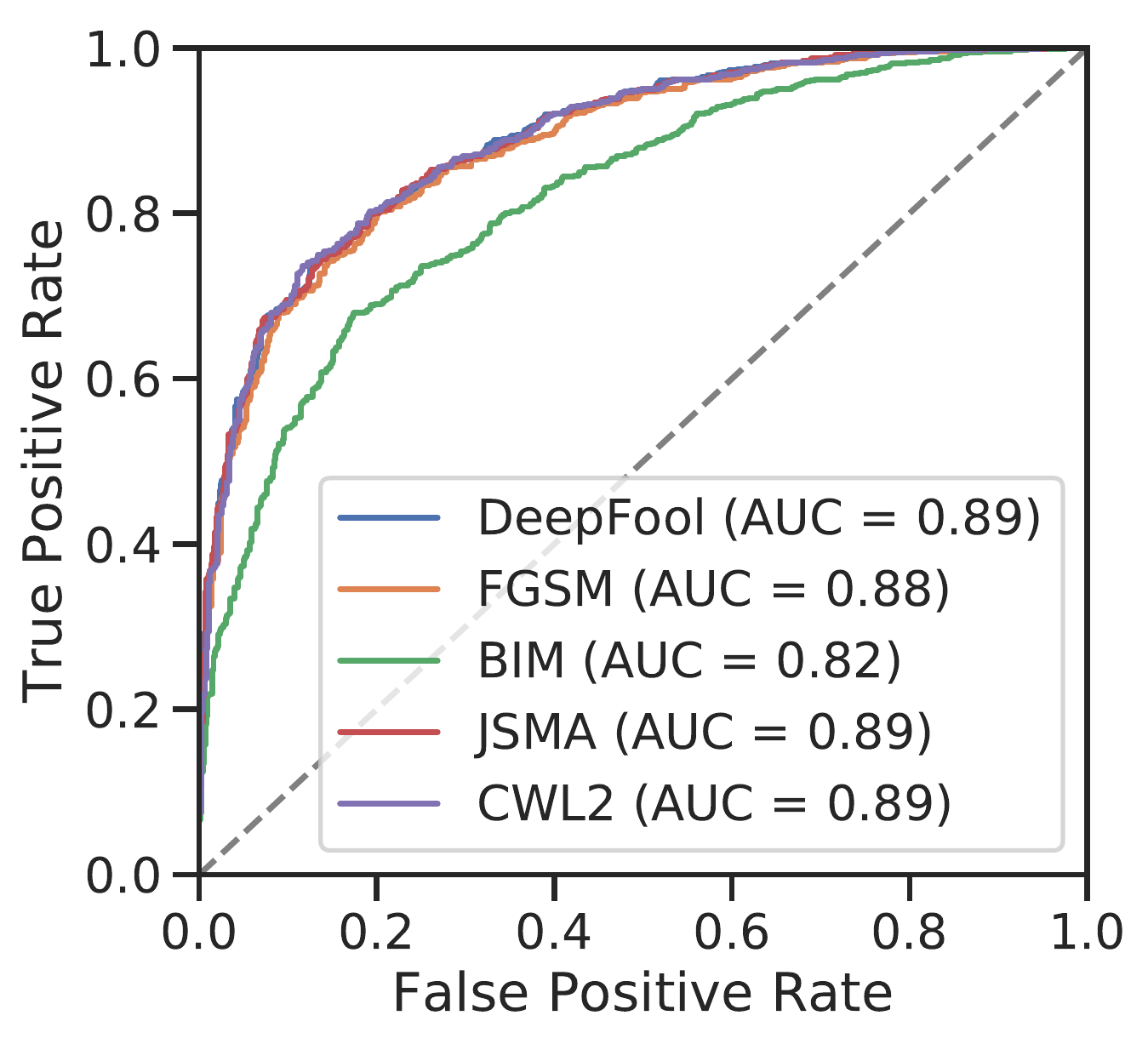}
\label{fig:resnet_roc_basic_u}
}
\vspace*{-0.3cm}
\endgroup
\caption{Detection results for LeNet (on MNIST), AlexNet (on ImageNet), and ResNet-50 with joint similarity.}
\label{fig:imagenet_detection_basic}
\vspace*{-0.3cm}
\end{figure}

\subsection{Linear Model}

We first evaluate the detection accuracy of the linear model in \Sec{sec:defense_model} for a wide range of adversarial attacks.

\vspace*{-0.5cm}
\paragraph{Non-targeted Attacks}
We evaluate non-targeted attacks, which are free to use any class as the adversarial image's inference result, with three different norms: FGSM and BIM with $l_\infty$ norm, DeepFool and C\&W $l2$ (CWL2) attack with $l_2$ norm, and JSMA with $l_0$ norm.
For LeNet, we achieves an area under the curve (AUC) value up to 0.95 in \Fig{fig:lenet_mnist_roc}. Even the lowest AUC value is 0.92, because of significant path similarity distinction between adversarial and normal images of MNIST. On ImageNet, we achieve AUC of 0.85\textasciitilde0.86 for AlexNet and AUC of 0.88\textasciitilde0.89 for ResNet-50, which has more layers to provide richer information for detection, leading to better accuracy.
The BIM has a low AUC value of 0.82. The reason is that BIM iteratively modifies all pixels (\Fig{fig:attack_examples}), which makes its rank-2 effective path behave slightly different from other attacks.

\begin{figure}[t]
\vspace*{-0.3cm}
\centering

\begingroup
\captionsetup[subfigure]{width=.33\linewidth}
\subfloat[Targeted attack.]{
\includegraphics[width=.333\linewidth]{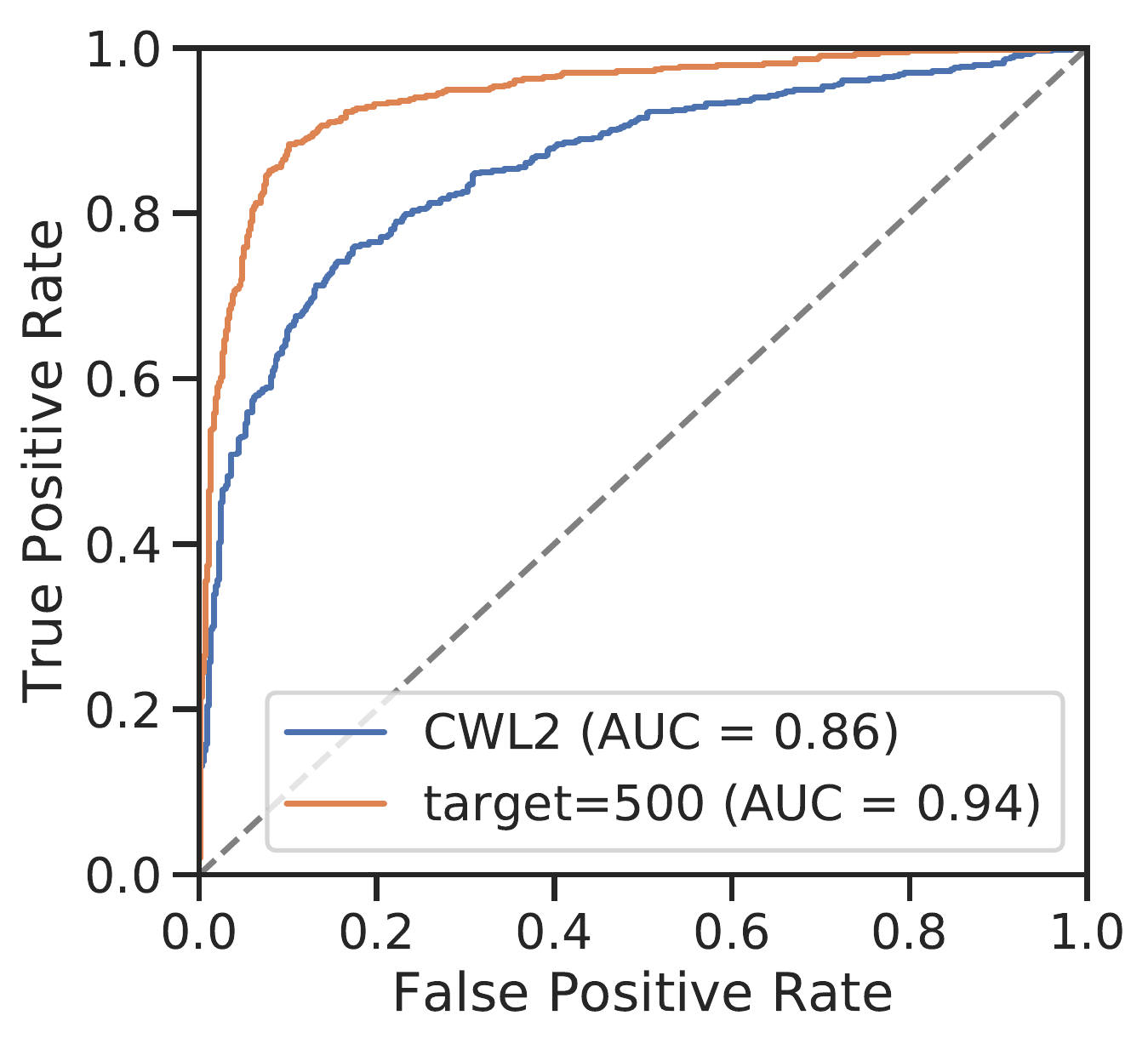}
\label{fig:roc_targeted}
}%
\subfloat[Universal perturbation attack.]{
\includegraphics[width=.333\linewidth]{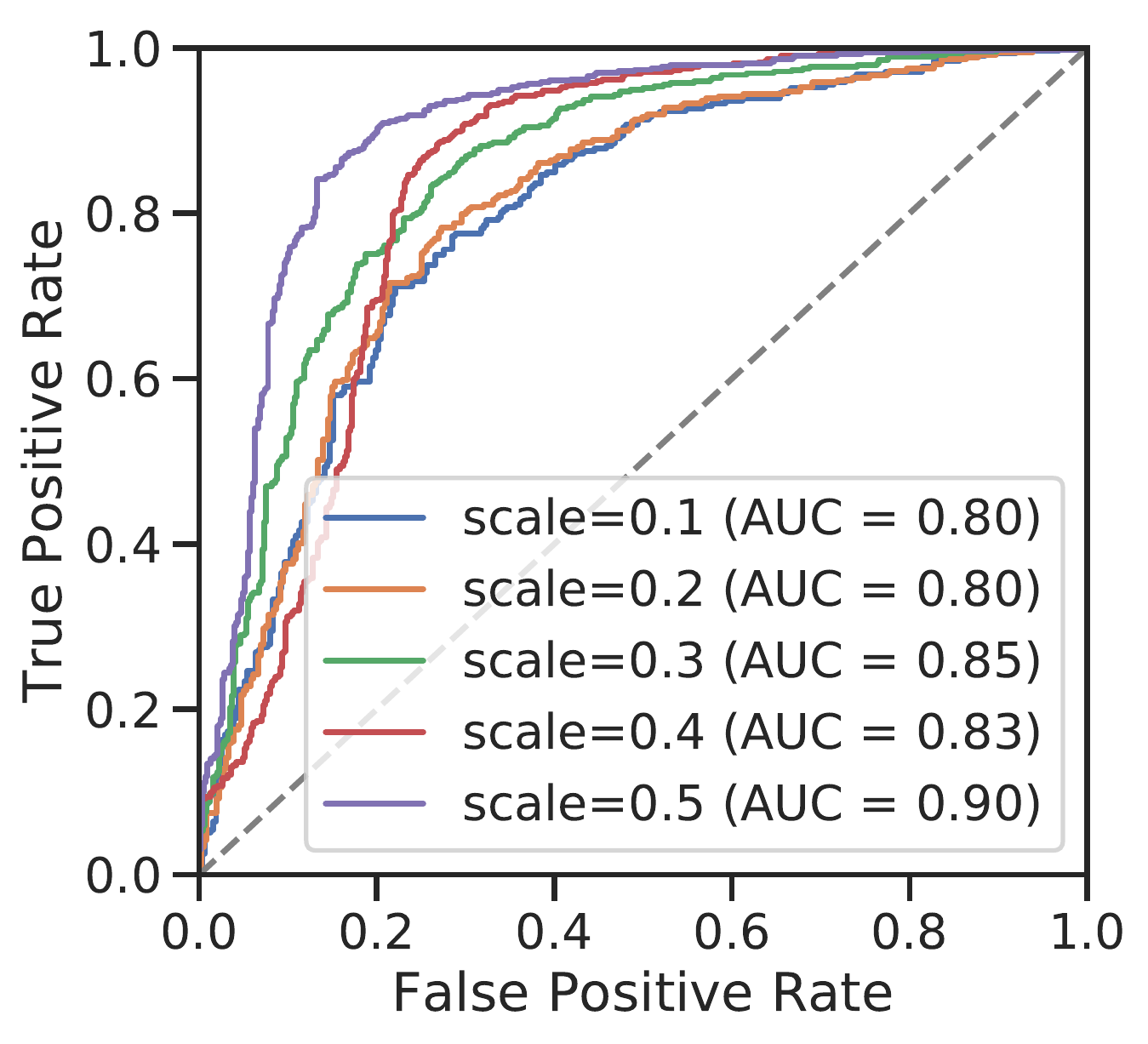}
\label{fig:roc_universal}
}%
\subfloat[Unrecognizable examples.]{
\includegraphics[width=.333\linewidth]{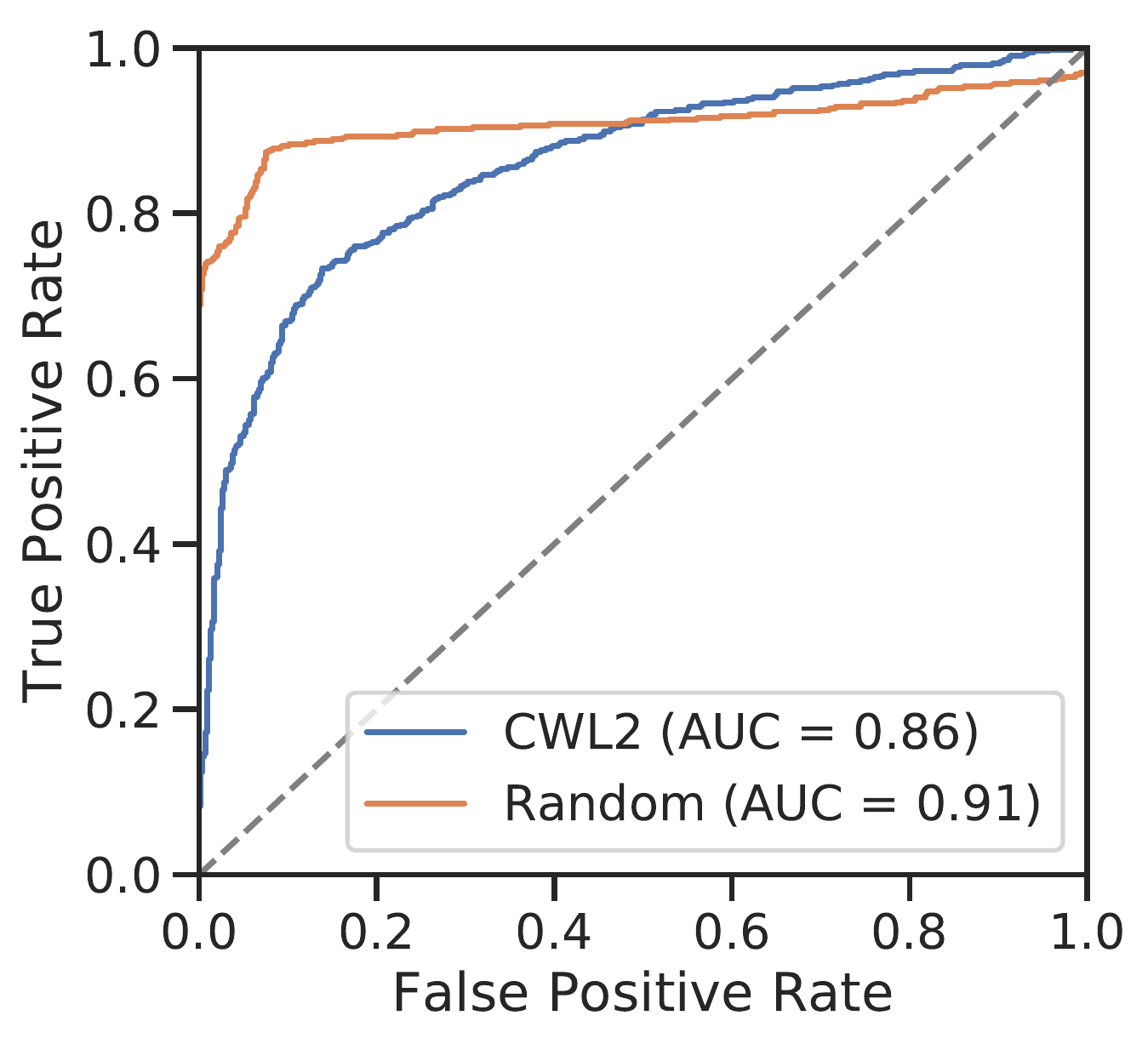}
\label{fig:roc_natural}
}
\vspace*{-0.3cm}
\endgroup
\caption{Linear model detection results for AlexNet on different attack methods.}
\label{fig:imagenet_detection_basic}
\vspace*{-0.3cm}
\end{figure}

\vspace*{-0.5cm}
\paragraph{Targeted Attack}
Targeted attacks are designed to mislead the prediction to a specific target class. \Fig{fig:roc_targeted} shows the result of evaluating targeted C\&W $l_2$ attack for AlexNet. We achieve AUC of 0.94, which is better than the non-targeted version. It is reasonable since the targeted attack's stricter constraint for target class requires larger perturbation, which eases our detection.

\vspace*{-0.5cm}
\paragraph{Universal Perturbation Attack}
Universal perturbation attacks generate perturbations that fool models on a large range of examples. Adversarial Patch~\cite{DBLP:journals/corr/abs-1712-09665} is an attack that generates universal perturbations in the form of image patches, which is robust against patch transformations such as translations, rotations or scaling. The result of adversarial patches in \Fig{fig:roc_universal} indicates that the detection becomes more accurate when the patch becomes larger. Our method can reach AUC of 0.9 when the patch scale relative to image size rises to 0.5.

\vspace*{-0.5cm}
\paragraph{Unrecognizable Examples}
\label{sec:defense-negative}
Adversarial examples are usually human-recognizable, however, unrecognizable images can also fool neural networks~\cite{DBLP:conf/cvpr/NguyenYC15}. We find that effective path can also be used to detect unrecognizable examples by evaluated on LeNet and AlexNet. For LeNet, our detector can recognize 93.85\% randomly generated images. For AlexNet, our method achieves AUC of 0.91 as shown in \Fig{fig:roc_natural}. In this sense, effective path offers the DNNs the ability to identify its recognizable inputs' distribution.

\vspace{0.1cm}
To summarize, the simple linear model constructed with effective path achieves high detection accuracy without requiring attack-specific knowledge.

\subsection{Comparison with Prior Work}

We now compare effective path based detection with prior work CDRP~\cite{wang2018interpret} with the linear model as well as three different kinds of models described in \Sec{sec:defense_model}.

\begin{figure}[t]
    \centering
    \includegraphics[trim=0mm 5mm 0mm 0mm,clip,width=\linewidth]{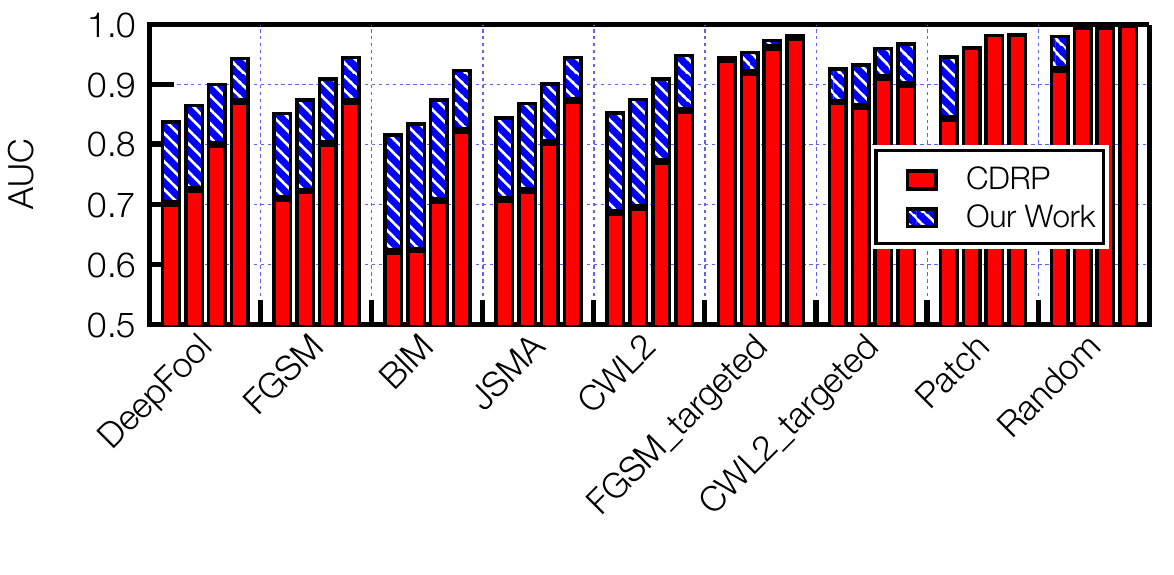}
    \vspace*{-0.3cm}
    \caption{Detection accuracy comparison between effective path and CDRP. Note that the four bars in an attack type represent linear model, AdaBoost, gradient boosting, and random forest in order. The blue bars indicate the AUC delta between our work and CDRP, no matter which is higher. Our work outperforms CDRP except on the linear model for the patch and random attack.}
    \label{fig:cdrp_compare}
    \vspace*{-0.3cm}
\end{figure}

\vspace*{-0.5cm}
\paragraph{Detection Accuracy}
\Fig{fig:cdrp_compare} compares the detection accuracy between effective path based models and CDRP based models.
For both approaches, we find that random forest performs the best while the linear model performs worst among all models.
However,  the effective path approach has a much smaller gap between random forest and linear model than the CDRP approach.
With the exception of patch and random attack, the effective path based linear model can outperform the CDRP based  random forest model.
In particular, the accuracy improvements for our approach are much more significant for the first five attack methods that are non-targeted and smaller for the two targeted attack method in the middle.
Note that the effective path based linear model performs slightly worse on the patch and random attack, which generate much different perturbation patterns (see \Fig{fig:attack_examples}).

\begin{figure}[t]
\vspace*{-0.3cm}
\centering
\begingroup
\captionsetup[subfigure][t]{width=.49\linewidth}
\subfloat[Linear model.]{
    \includegraphics[trim=0mm 2mm 70mm 0mm,clip,height=.36\linewidth]{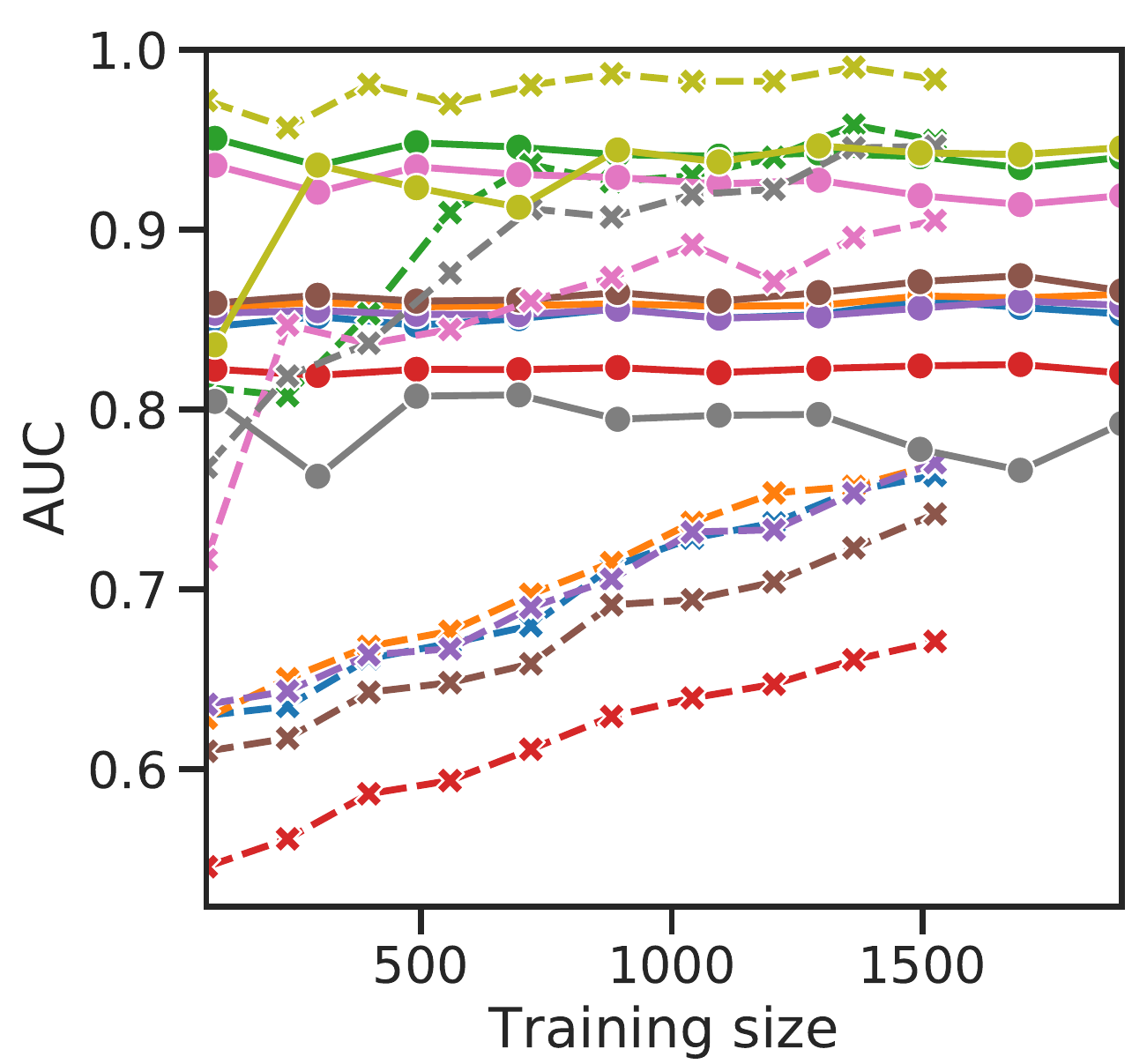}
    \label{fig:training_set_size_linear}
}%
\subfloat[Random forest.]{
    \includegraphics[trim=0mm 2mm 0mm 0mm,clip,height=.36\linewidth]{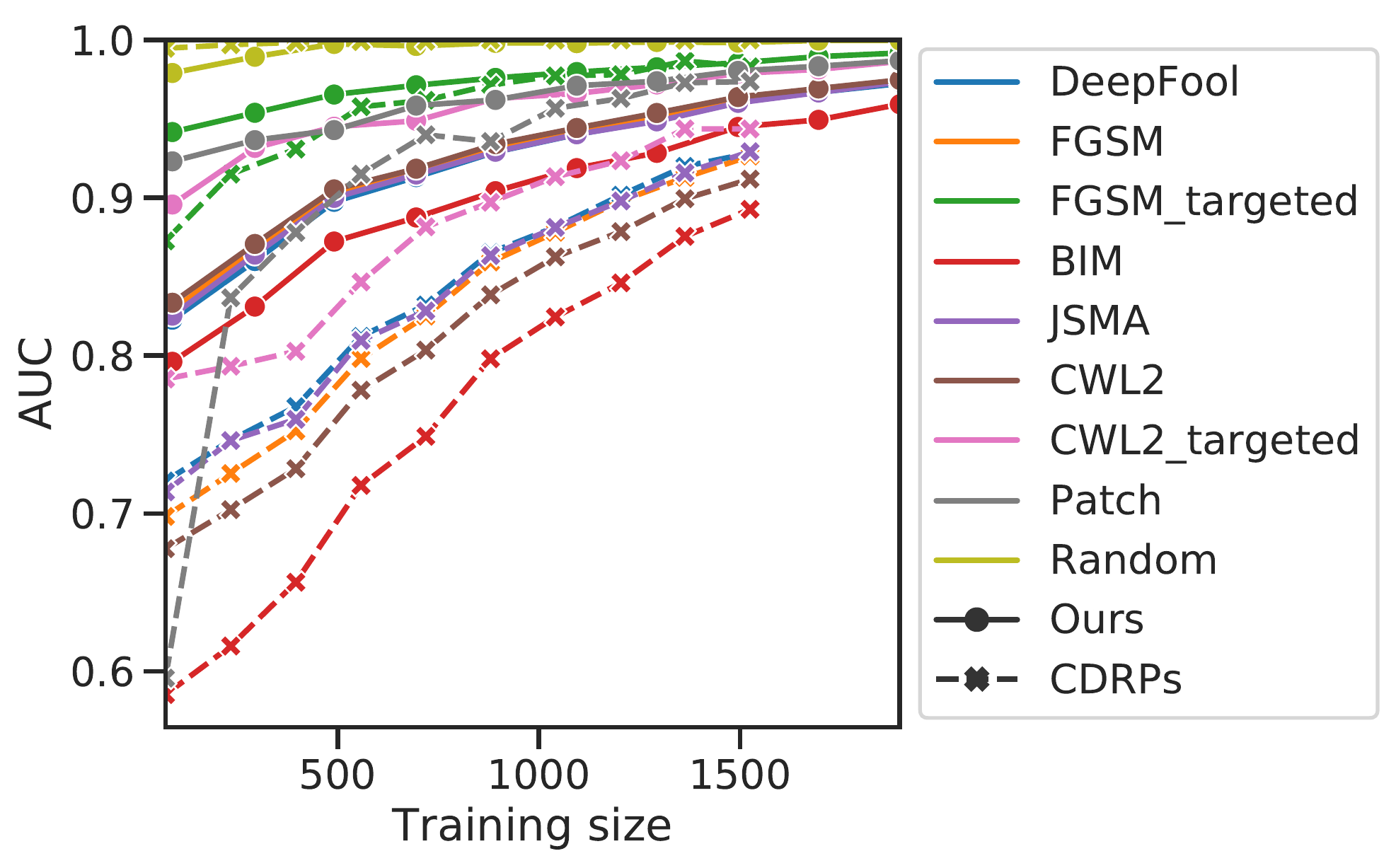}
    \label{fig:training_set_size_rf}
}%
\vspace*{-0.3cm}
\endgroup
\caption{Impact of training set size on the AUC.}
\label{fig:training_set_size}
\vspace*{-0.3cm}
\end{figure}

\vspace*{-0.5cm}
\paragraph{Training Size}
We also study how the size of the training set impacts the detection accuracy. We choose the linear model and random forest model and gradually increase their training set size.
\Fig{fig:training_set_size} compares our approach with the CDRP approach.
For the linear model, our approach stabilizes with a small number of training samples (around 100 images) while the CDRP requires much larger training set size.
For the random forest model, both approaches require a larger training set while our approach is less sensitive because our input feature is low dimensional and effective.

\vspace*{-0.5cm}
\paragraph{Generalizability}
Generalizability measures a defense's ability to withstand unknown attacks.
To study the generalizability of our detection model, we perform a control experiment: we gradually add the adversarial samples from different attack types for training the detection model and observe the model detection accuracy all on attack types.
\Fig{fig:attack_number} shows the experiment results where we add the adversarial samples in the order of legend shown in the right.
For both linear model and random forest model, our work generalizes well to unseen attacks because effective path captures their common behavior.
The CDRP based linear model performs worse for all non-targeted attacks, and its accuracy on targeted attack (FGSM\_targeted and CWL2\_targeted) shows an abrupt increase once incorporating the corresponding samples for training the model.


\begin{figure}[t]
\vspace*{-0.3cm}
\centering
\begingroup
\captionsetup[subfigure][t]{width=.49\linewidth}
\subfloat[Linear model.]{
    \includegraphics[trim=0mm 2mm 70mm 0mm,clip,height=.36\linewidth]{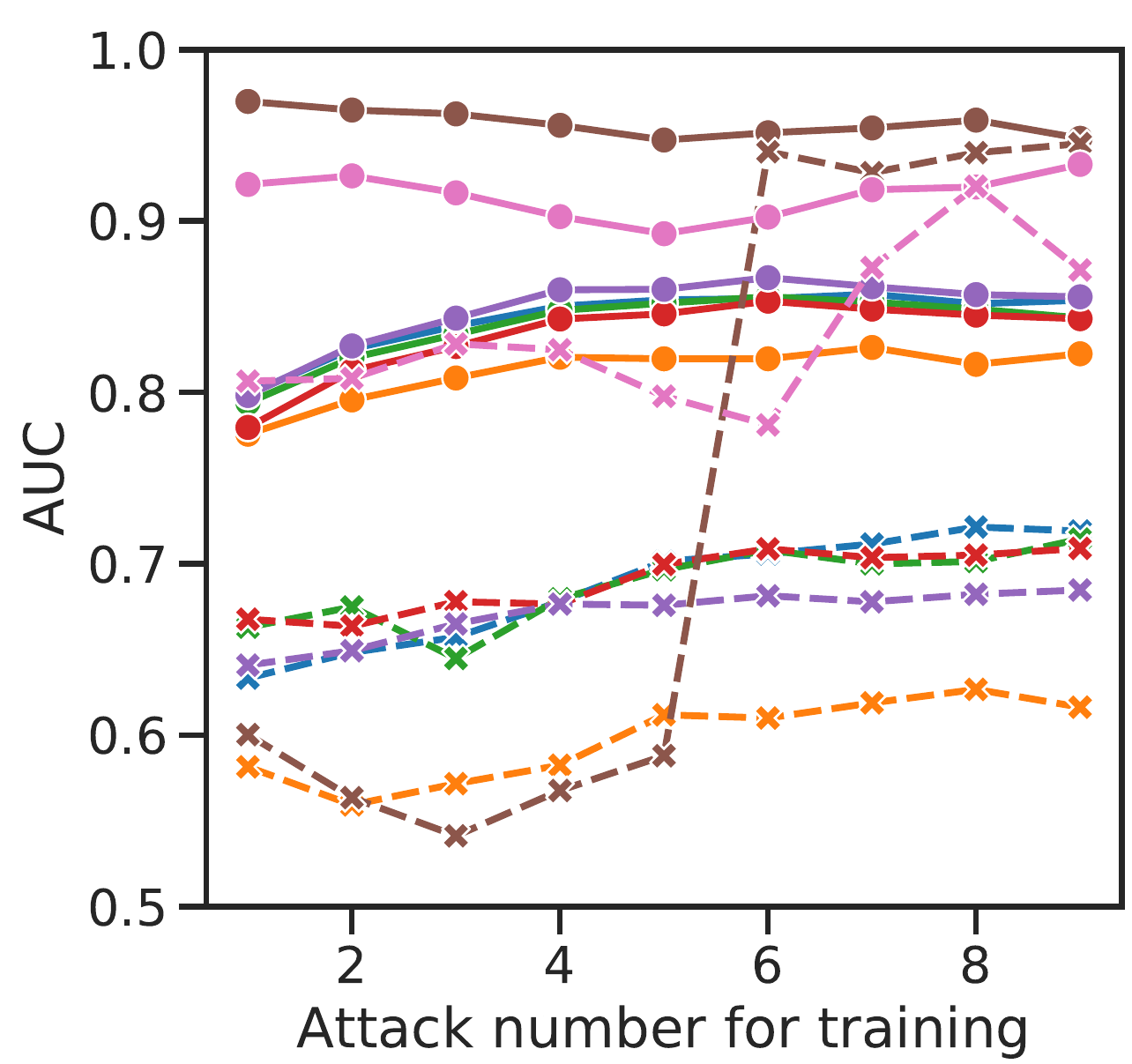}
    \label{fig:attack_number_linear}
}%
\subfloat[Random forest.]{
    \includegraphics[trim=0mm 2mm 0mm 0mm,clip,height=.36\linewidth]{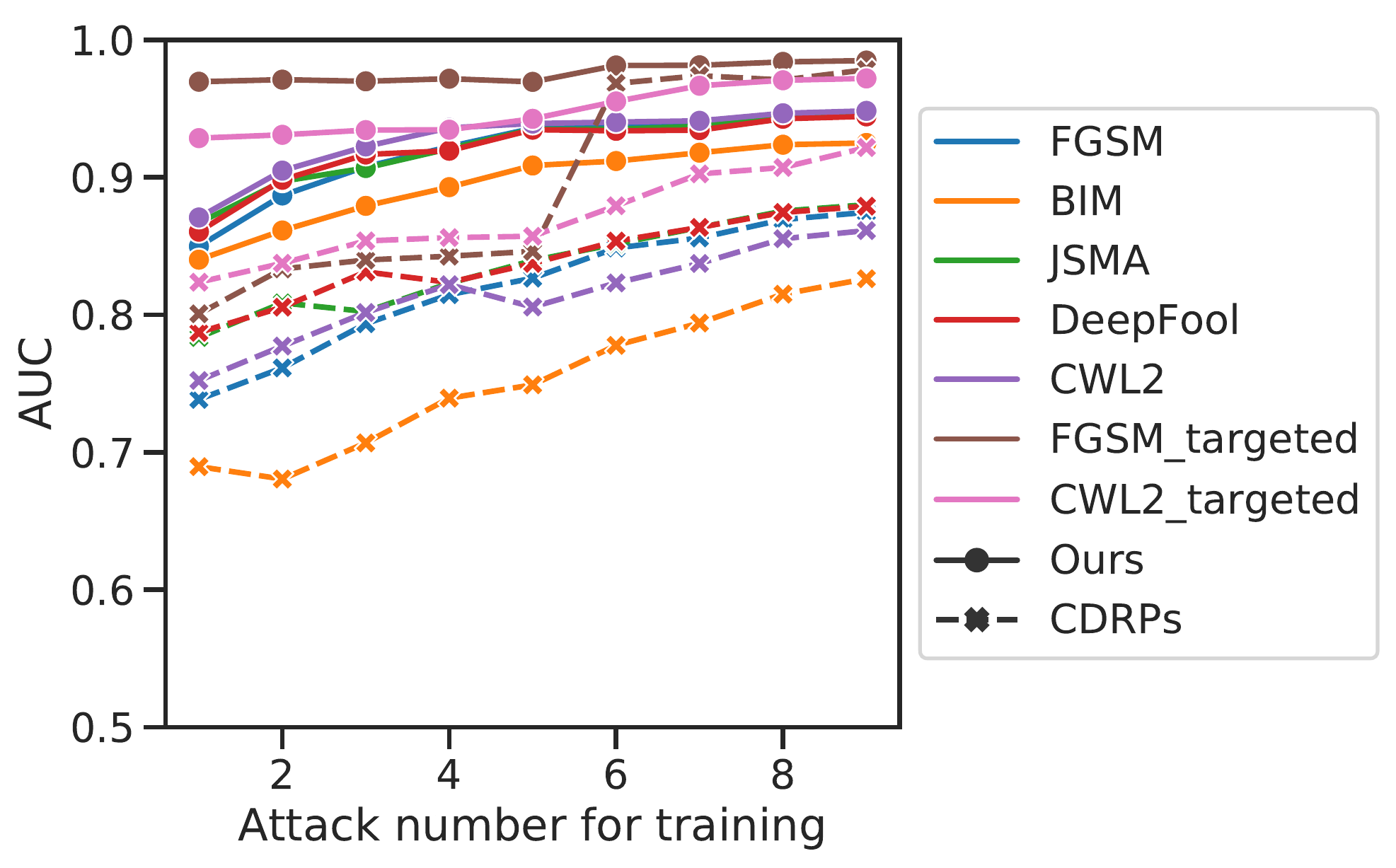}
    \label{fig:attack_number_rf}
}%
\vspace*{-0.3cm}
\endgroup
\caption{Impact of attack number in the training set.}
\label{fig:attack_number}
\vspace*{-0.3cm}
\end{figure}

\begin{figure}[t]
    \centering
    \subfloat[ROC for C\&W $l_2$ attack with different $\theta$.]{
    \includegraphics[width=.4\linewidth]{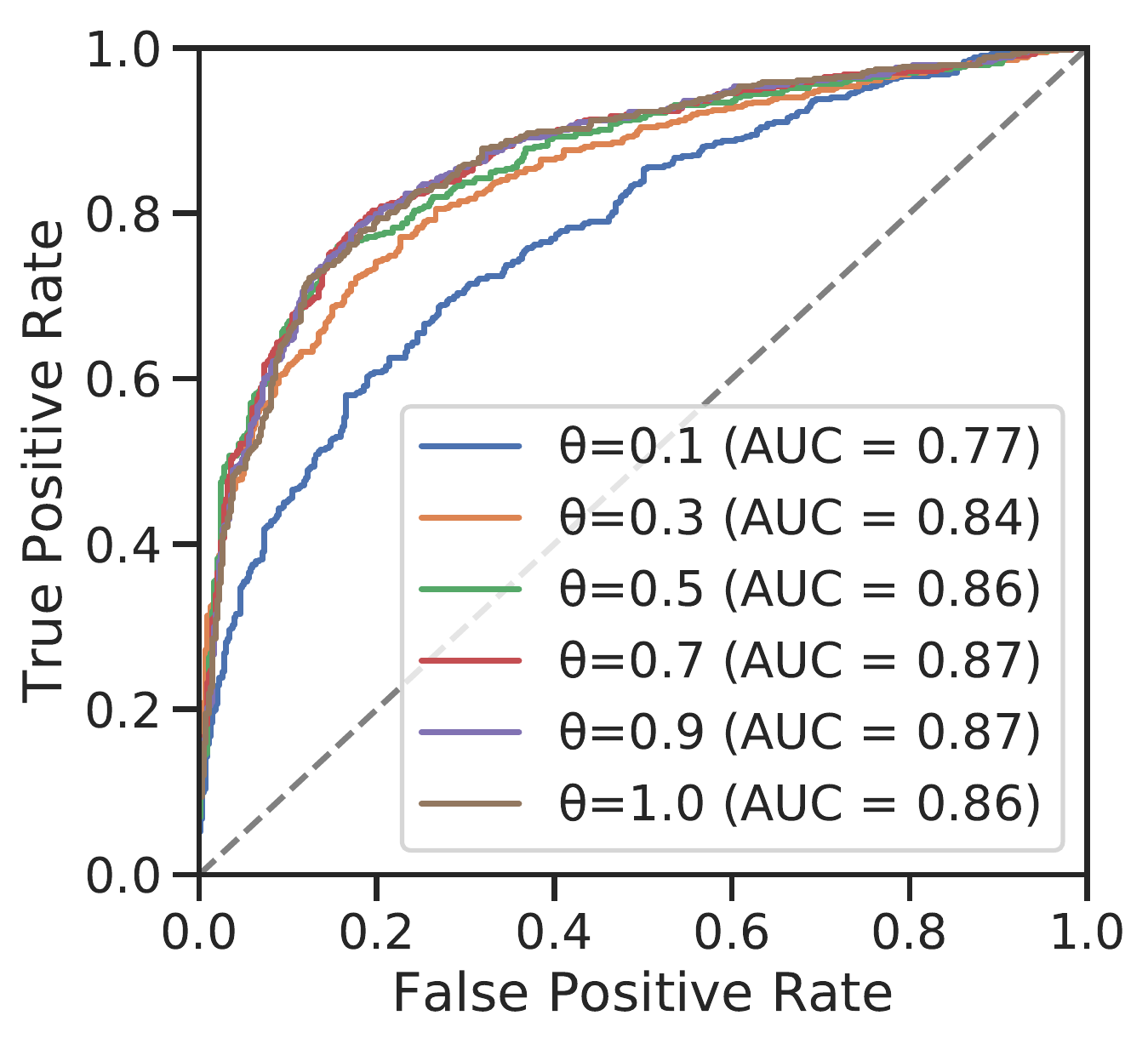}
    \label{fig:roc_theta}
    }
    \hspace{.8cm}
    \subfloat[Effective path size with different $\theta$.]{
    \includegraphics[width=.4\linewidth]{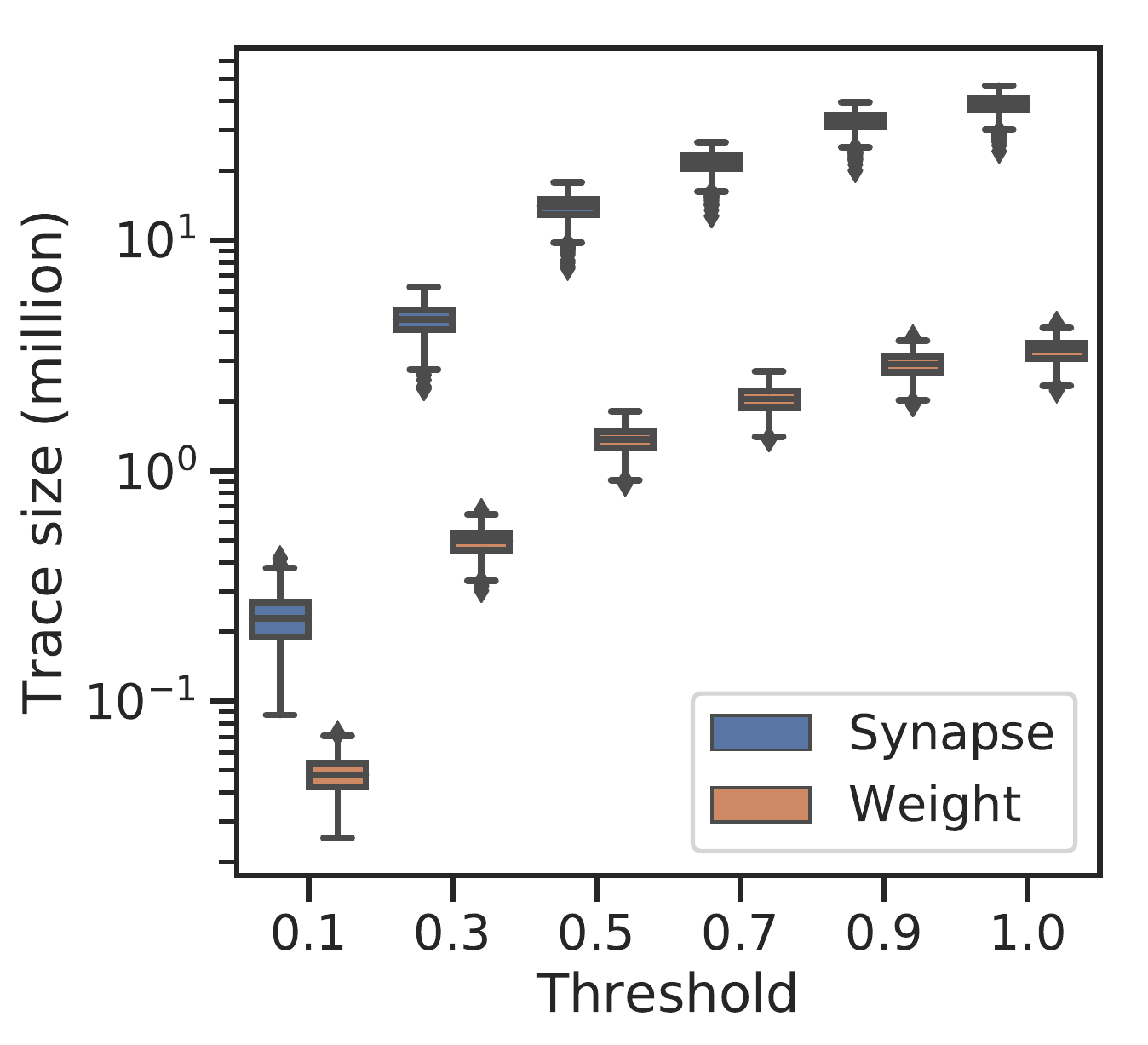}
    \label{fig:path_size}
    }
    \vspace*{-0.3cm}
    \caption{Effective path $\theta$ sensitivity study.}
    \label{fig:attack_number}
    \vspace*{-0.3cm}
\end{figure}

\subsection{Sensitivity Study}

After demonstrating the accuracy of using effective path to detect adversarial samples, we now perform sensitivity study on its parameters, including the $\theta$ in Equation~\ref{eq:eq1} and the number of extracted layers.
Our results further unveil optimization opportunity to make effective path more practical.

\vspace*{-0.5cm}
\paragraph{Parameter Sensitivity}
The first tunable parameter of effective path extraction is $\theta$.
We test C\&W $l_2$ attack with $\theta$ value varying from 0.1 to 1.0 in \Fig{fig:roc_theta}. The detection performance remains almost unchanged when $\theta$ is in range of 0.5 and 1.0, and decreases from $\theta = 0.3$.
\Fig{fig:path_size} shows that the effective path size under $\theta = 0.3$ decrease by one order of magnitude compared with $\theta = 1.0$, with slightly lower detection accuracy. We choose $\theta = 0.5$ as default value to save storage space and improve extraction performance without accuracy loss.

\begin{figure}[t]
\vspace*{-0.3cm}
\centering
\begingroup
\captionsetup[subfigure][t]{width=.49\linewidth}
\subfloat[Linear model.]{
    \includegraphics[trim=0mm 2mm 70mm 0mm,clip,height=.36\linewidth]{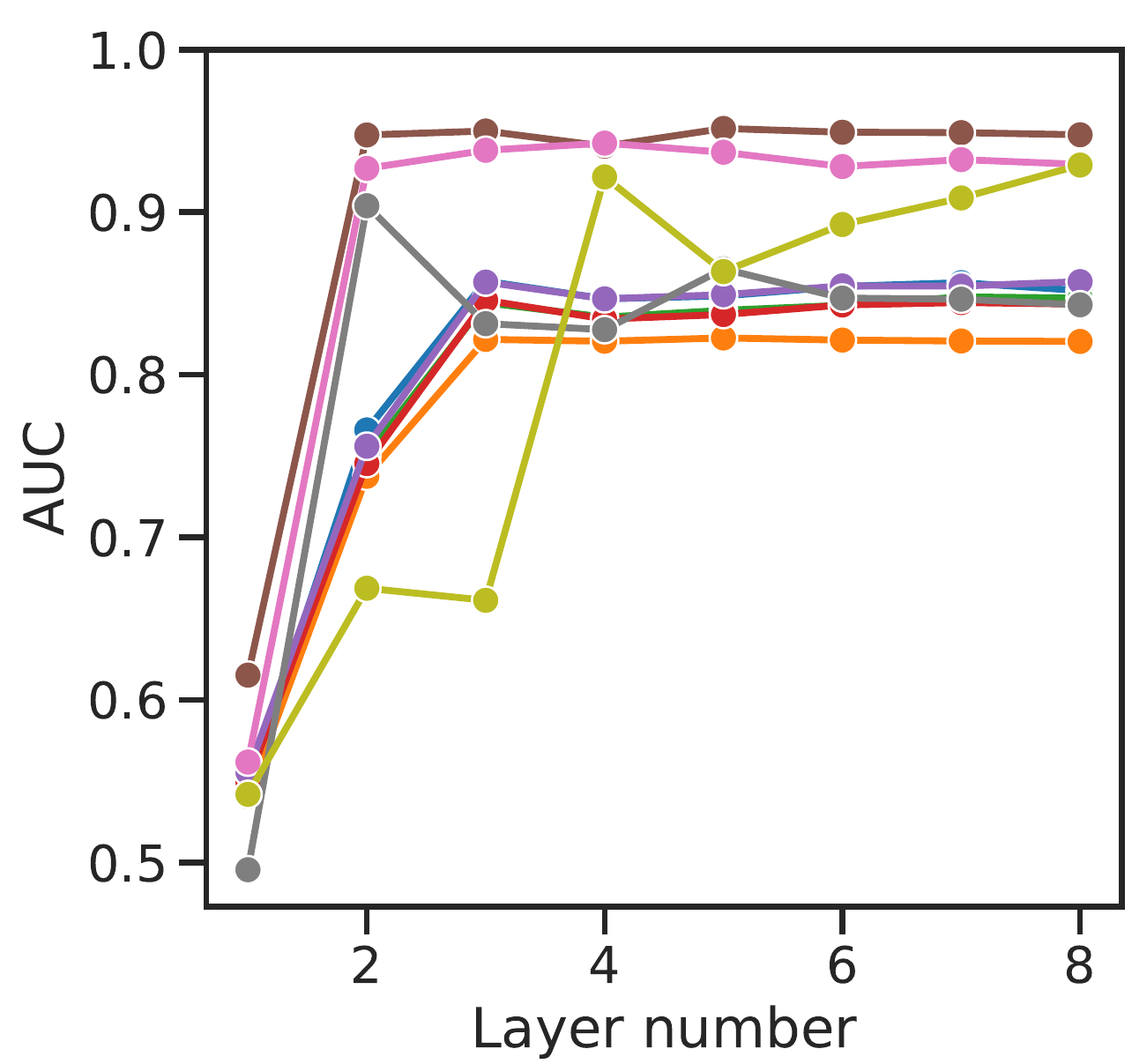}
    \label{fig:layer_number_linear}
}%
\subfloat[Random forest.]{
    \includegraphics[trim=0mm 2mm 0mm 0mm,clip,height=.36\linewidth]{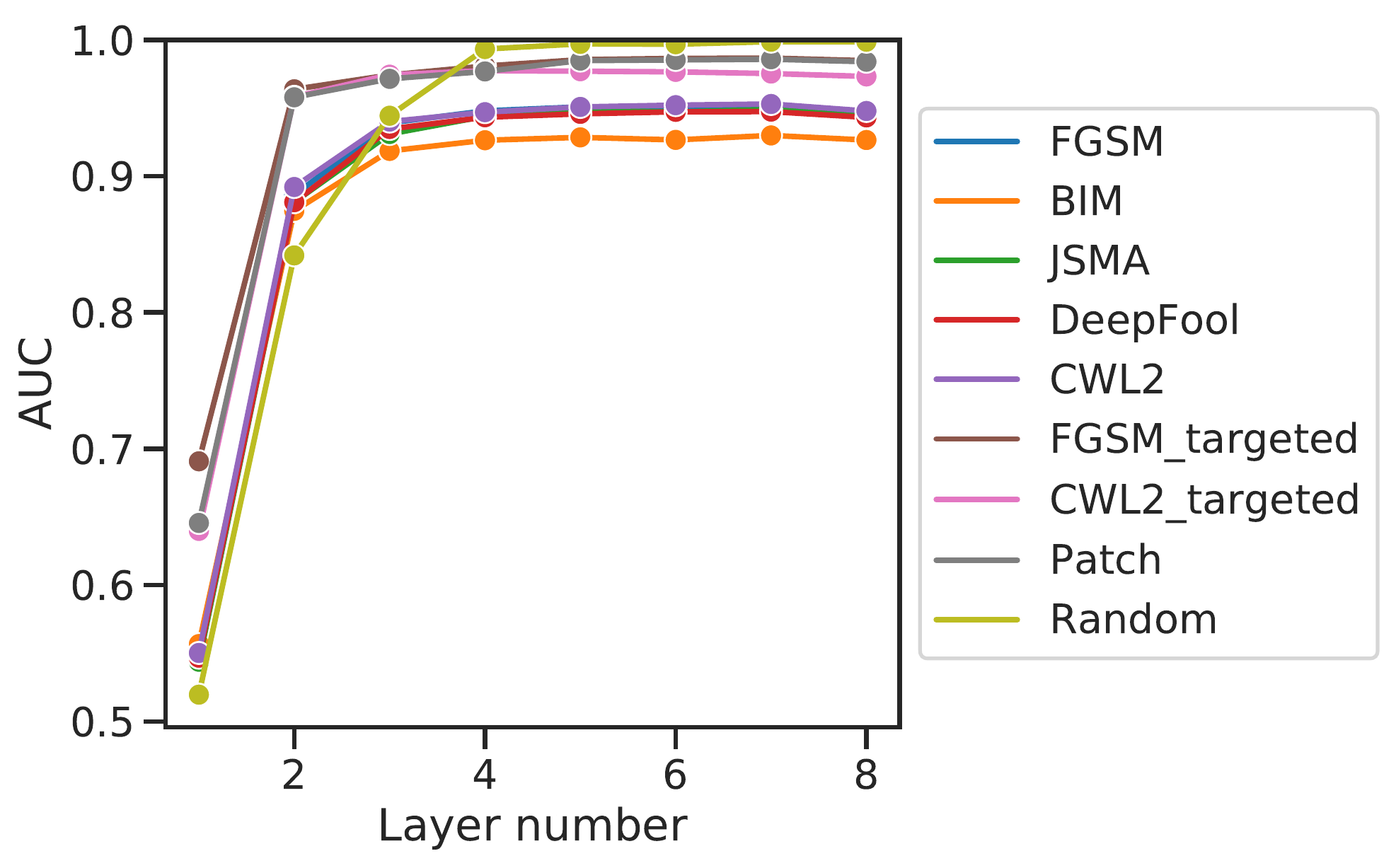}
    \label{fig:layer_number_rf}
}%
\vspace*{-0.3cm}
\endgroup
\caption{Effective path layer number impact on AUC.}
\label{fig:layer_number}
\end{figure}

\vspace*{-0.5cm}
\paragraph{Layer Sensitivity}
Another tunable parameter of effective path extraction is the number of layers as it is extracted layer by layer.
We perform experiments to study the layer number's impact on the adversarial sample detection accuracy and show the result in \Fig{fig:layer_number}.
Note that we extract the effective path starting from the last layer and the last three layers are fully connected layers in AlexNet.
For both linear model and random forest, we observe that the AUC performance for all attacks except random attack saturates after three layers, i.e. FC layers.
The random attack detection accuracy saturates after four layers, i.e. one additional CONV layer.

{
\renewcommand{\arraystretch}{1.2}
\begin{table}[t]
    \caption{Effective path extraction time (second).}
    \label{tab:extraction_time}
    \vspace*{-0.1cm}
    \centering
    \resizebox{\linewidth}{!}{
        \begin{tabular}{c  c c c}
            \Xhline{1.2pt}
            Method    & Effective Path (Full) & Effective Path (Partial) & CDRP            \\\hline
            AlexNet   & 1.43 $\pm$ 0.09       & 0.43 $\pm$ 0.17          & 106.4 $\pm$ 5.2 \\
            ResNet-50 & 68.32 $\pm$ 2.43      & 0.83 $\pm$ 0.21          & 406.3 $\pm$ 6.3 \\
            \Xhline{1.2pt}
        \end{tabular}
    }
    \vspace*{-0.3cm}
\end{table}
}

With the layer sensitivity insight, we can extract effective path for just enough layers instead of the full network, which can significantly reduce the extraction time.
\Tbl{tab:extraction_time} compares the extraction time for AlexNet and ResNet-50. Extracting the full effective path for the entire network is still much less expensive (70$\times$ for AlexNet and 6$\times$ for ResNet-50) than extracting CDRP which requires the retraining process.
Moreover, extracting the partial effective path can lead to even faster process time, which translate to 240$\times$ and 500$\times$ time reduction compared to CDRP extraction.

In summary, effective path enables the use of the highly interpretable linear model to detect a broad range of adversarial attacks, and can achieve great accuracy on different datasets and models.
Compared to prior work CDRP~\cite{wang2018interpret}, our approach achieves better accuracy, requires less training samples, and generalizes well to different adversarial attacks.


%% file: tex/related.tex
\section{Related Work}

{
\renewcommand{\arraystretch}{1.2}
\begin{table}[t]
    \caption{Comparison with other defenses.}
    \label{tab:other_defenses}
  \vspace*{-0.3cm}
    \centering
    \resizebox{\linewidth}{!}{
    \begin{tabular}{c c c c c c c c c c c}
        \Xhline{1.2pt}  
         Type & Defense & $l_0$ & $l_2$ & $l_\infty$ & Attack & Generalizability & Scale \\
        \hline
        \multirow{ 4}{*}{Detector} &
        \textbf{Effective Path} & \textbf{Y} & \textbf{Y} & \textbf{Y} & \textbf{all discussed} & \textbf{strong} & \textbf{ImageNet} \\
        & ~\cite{DBLP:journals/corr/MetzenGFB17} & - & \textbf{Y} & \textbf{Y} & (non-)targeted & weak & CIFAR-10 \\
        & ~\cite{Meng:2017:MTD:3133956.3134057} & - & \textbf{Y} & \textbf{Y} & (non-)targeted & - & CIFAR-10 \\
        & ~\cite{wang2018interpret} & - & - & \textbf{Y} & targeted & - & \textbf{ImageNet} \\
        \hline
        \multirow{ 1}{*}{Adversarial} &
        \cite{DBLP:journals/corr/MadryMSTV17} & - & \textbf{Y} & \textbf{Y} & (non-)targeted & weak & CIFAR-10 \\
        Training& ~\cite{DBLP:journals/corr/abs-1708-02582} & - & - & \textbf{Y} & (non-)targeted & weak & CIFAR-10 \\
        \hline
        \multirow{1}{*}{Input}
        & ~\cite{guo2018countering} & - & \textbf{Y} & \textbf{Y} & (non-)targeted & - & \textbf{ImageNet} \\
      Transformation  & ~\cite{buckman2018thermometer}  & - & - & \textbf{Y} & (non-)targeted & - & CIFAR-100 \\
         \hline
        \multirow{2}{*}{Randomization} 
        & \cite{xie2018mitigating}  & - & \textbf{Y} & \textbf{Y} & (non-)targeted & - & \textbf{ImageNet} \\
        & \cite{s.2018stochastic}  & - & - & \textbf{Y} & (non-)targeted & - & CIFAR-10 \\
        \hline
        \multirow{1}{*}{Generative}
        & \cite{DBLP:journals/corr/abs-1805-09190} & \textbf{Y} & \textbf{Y} & \textbf{Y} & (non-)targeted + random & \textbf{strong} & MNIST \\
        Model & ~\cite{samangouei2018defensegan} & - & \textbf{Y} & \textbf{Y} & (non-)targeted & - & MNIST \\
        \Xhline{1.2pt}
    \end{tabular}
    }
\end{table}
}

To compare our defense method with prior work, we first categorize various defenses methods to the five types listed in \Tbl{tab:other_defenses}. Since almost all the compared work reported a similar detection accuracy (AUC value 0.9 - 0.95), we focus the comparison on the comprehensiveness, attack method, generalizability, and scale of their evaluation.
The "-" in the table indicates that there are not enough details or experimental results to deduce an appropriate conclusion.

\vspace*{-0.5cm}
\paragraph{Detector}
Our work fits in the detector category, which does not require any modification to inputs, models, or training process. Prior work~\cite{DBLP:journals/corr/MetzenGFB17} trained a DNN from network activations to detect adversarial examples. The detector subnetwork doesn't generalize well across different attack parameters or attack types because the activation values are highly attack-specific, which motivates \cite{Meng:2017:MTD:3133956.3134057} to propose MagNet. MagNet uses a reformer to move adversarial examples to normal examples' manifold. However, \cite{DBLP:journals/corr/abs-1711-08478} shows that MagNet can be defeated by a little increase of perturbation. 

The closest work to ours is \cite{wang2018interpret}, which uses the importance coefficients of different channels in the network (named critical data routing paths, abbr. CDRPs) to detect adversarial examples. However, CDRPs do not have aggregation capability as a single channel can have different significance values for different images. As such, CDRPs fail to defend non-targeted attacks and have weak generalizability. In comparison, we use the effective path, which is essentially a binary value for each neuron/synapse, and therefore can be directly aggregated. Our method generalizes well for different attacks and provides strong transferability.

\vspace*{-0.5cm}
\paragraph{Adversarial Training}
Adversarial training requires additional training step to protect the DNNs. It has two known disadvantages: it is difficult to perform in the large-scale dataset like ImageNet~\cite{DBLP:journals/corr/KurakinGB16a}, at the same time easy to overfit to the trained kinds of adversarial examples. Even adversarial training proposed by~\cite{DBLP:journals/corr/MadryMSTV17}, considered as the only effective defense among white-box-secure defenses at ICLR 2018~\cite{DBLP:conf/icml/AthalyeC018}, is found overfitting on the trained $l_\infty$ metric~\cite{DBLP:journals/corr/abs-1710-10733}.

\vspace*{-0.5cm}
\paragraph{Input Transformation}
Many image transformations like rescaling, bit-depth reduction and compression can disturb attacks and increase the perturbation lower bound, with the sacrifice of classification accuracy. This kind of defense method works less well for patch-based attacks and does not provide the ability to filter unrecognizable examples.

\vspace*{-0.5cm}
\paragraph{Randomization}
Randomization-based defense methods apply random modifications to model weights. They can increase required distortion by forcing attacks to generate transferable adversarial examples over a series of possible modified models. However, they also stochastically alter the prediction results, leading to the overhead of more forward passes or retraining steps.

\vspace*{-0.5cm}
\paragraph{Generative Model}
Generative model based defenses change the classification model. They project the inputs onto the manifold before classification. \cite{DBLP:journals/corr/abs-1805-09190} propose a classification model that shows good generality and transferability on MNIST, but its performance on large dataset like ImageNet is still obscure. GAN-based defenses are also hard to apply in ImageNet scale due to its computational cost.

%% file: tex/conclusion.tex
\section{Conclusion and Future Work}

In this work, we propose a novel profiling based method to extract the deep neural network's (DNN) path information when inferring an image.
This method does not modify the DNN structure and can extract meaningful path information that represents the critical dataflow inside the DNN.
We study how to use the extracted path information to decompose a DNN model into different functional blocks corresponding to different inference classes.
Through analysis, we find that adversarial images can activate functional blocks different from normal images to fool the DNN's prediction results because all the blocks are connected.
We propose a defense method that only uses the information from the training set and the image itself, without requiring any knowledge of a specific attack.
The defense method achieves high accuracy and broad coverage of mainstream attacks.



Besides adversarial defense, the effective path can also be used to understand the DNN's working mechanism. In the appendix, we report our preliminary result on how the training process and different DNN topology affects the effective path density and similarity. We believe that the functionality based decomposition is a promising direction for understanding DNNs.



%% file: tex/acknowledgment.tex
\footnotesize{
\paragraph{Acknowledgments.}
This work is sponsored in part by the National Basic Research 973 Program of China (No. 2015CB352403), the National Natural Science Foundation of China under Grant Nos. 61702328 and 61602301, and partially funded by Microsoft Research Asia Collaborative Research Grant.
}

%% file: tex/appendix.tex
\section{Effective Path Extraction for More Network Structures}
\label{sec:path_extend}

For the sake of brevity, we only introduce effective path extraction for networks consist of convolutional layers and FC layers in the submitted paper. We further explain other common network structures' extraction methods in this section.

\paragraph{Skip Connection}
To handle skip connections in ResNet, we need to merge neurons contributed from two different layers. Consider a skip connection from layer $l$ to layer $l+m$, then active neurons in layer $l$ are collected from layer $l+1$ and $l+m$, denoted as $\mathcal{N}^{l} =  \{n_{k}^{l} | k \in \tilde{K}^{l+1} \ or \ k \in \tilde{K}^{l+m} \}$, where $\tilde{K}^{l+1}$ and $\tilde{K}^{l+m}$ are the selected sets of weight indices in layer $l+1$ and $l+m$ respectively.

\paragraph{Pooling Layer}
Pooling layers can be treated as the special case of convolutional layers during extracting. For average pooling layer, we treat it as a convolutional layer with all weights equal to 1; for max pooling layer, we treat it as a convolutional layer that always picks rank-1 weight and input neuron pair when finding the minimum $\tilde{K}_{p}^{l}$.

\section{Effective Path Visualization for CIFAR-100}

\begin{figure}[t]
    \centering
    \includegraphics[width=0.95\columnwidth]{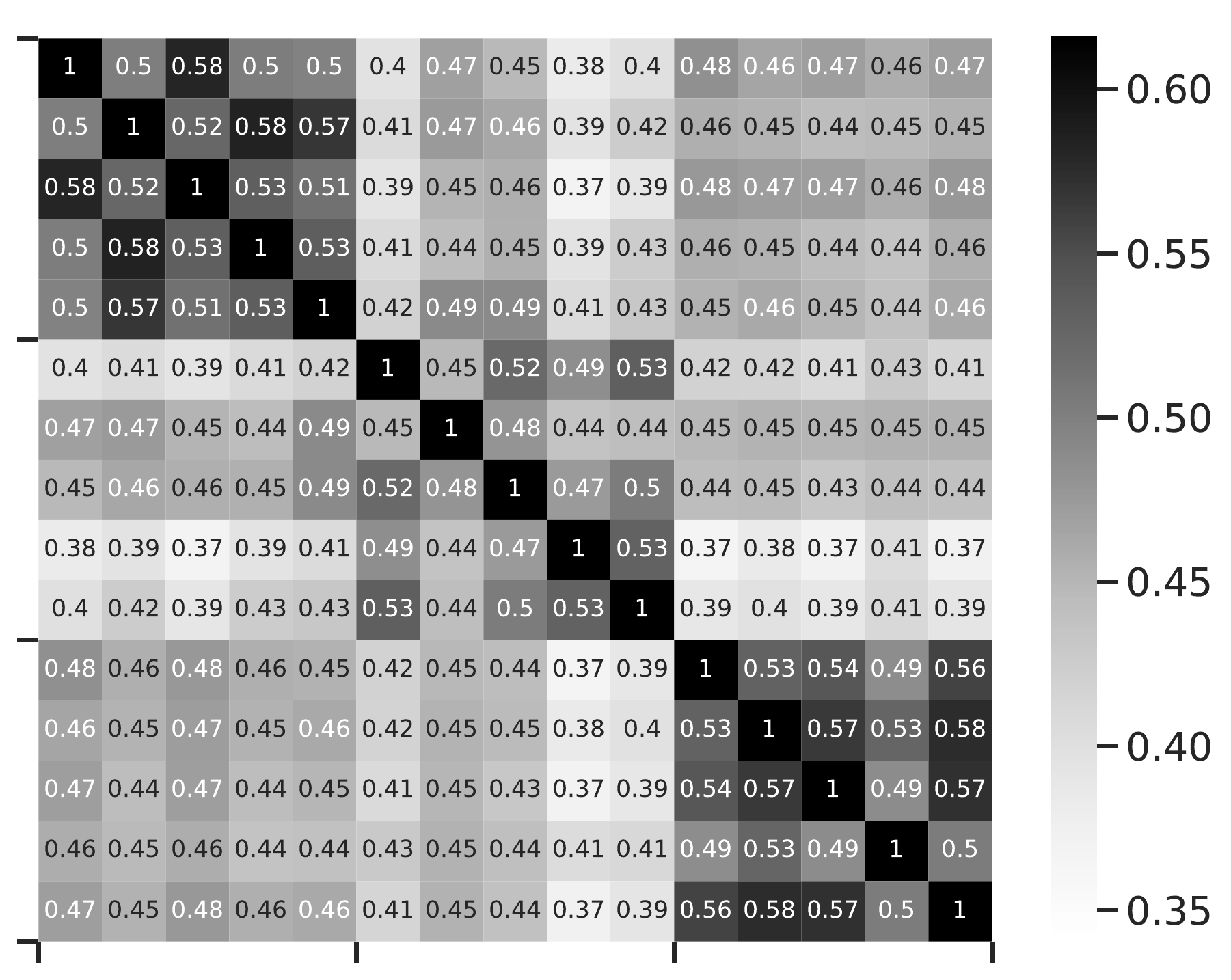}
    \vspace*{-0.3cm}
    \caption{Class-wise path similarity for CIFAR-100.}
    \label{fig:cifar100_similarity}
\end{figure}

To explore path specialization on more realistic dataset than LeNet, we study the similarity of per-class effective paths on CIFAR-100. \Fig{fig:cifar100_similarity} shows the class-wise path similarity of 15 classes in CIFAR-100, which are belonged to three different super classes. We show super classes vehicles 1, large natural outdoor scenes and flowers from left to right, each of which contains 5 basic classes. We can find that classes belong to the same super classes have higher similarity than classes from different super classes, which indicates that effective path discovers the class hierarchy without prior knowledge of super classes.

\section{Adversarial Samples Defense}

\subsection{Adversarial Samples Similarity Analysis for ResNet-50}

\begin{figure*}[t]
    \centering
    \subfloat[Rank-1 similarity.]{
    \includegraphics[width=.95\columnwidth]{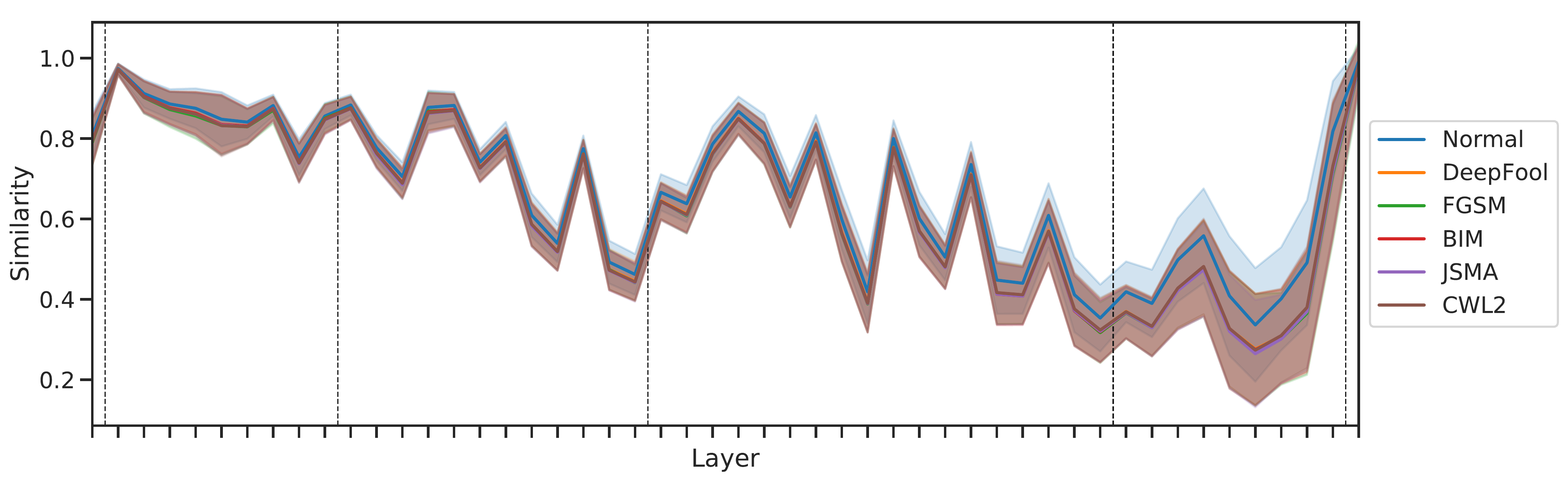}
    \label{fig:resnet_similarity_rank1}
    }
    \hfill
    \subfloat[Rank-1 similarity delta.]{
    \includegraphics[width=.95\columnwidth]{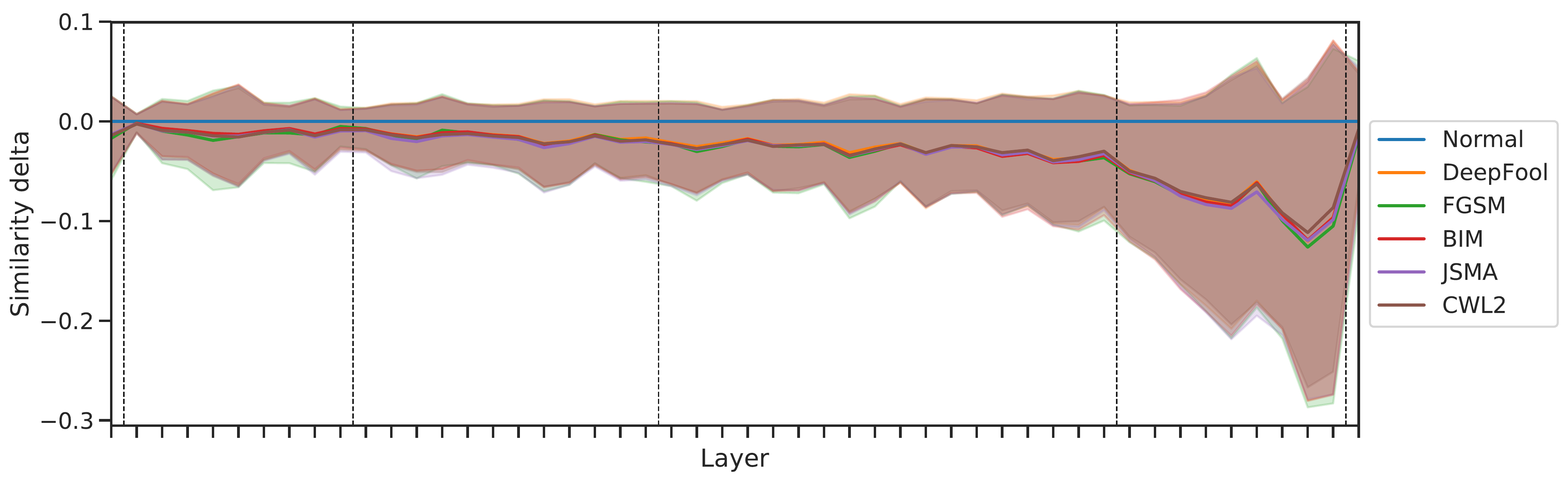}
    \label{fig:resnet_similarity_delta_rank1}
    }
    \hfill
    \subfloat[Rank-2 similarity.]{
    \includegraphics[width=.95\columnwidth]{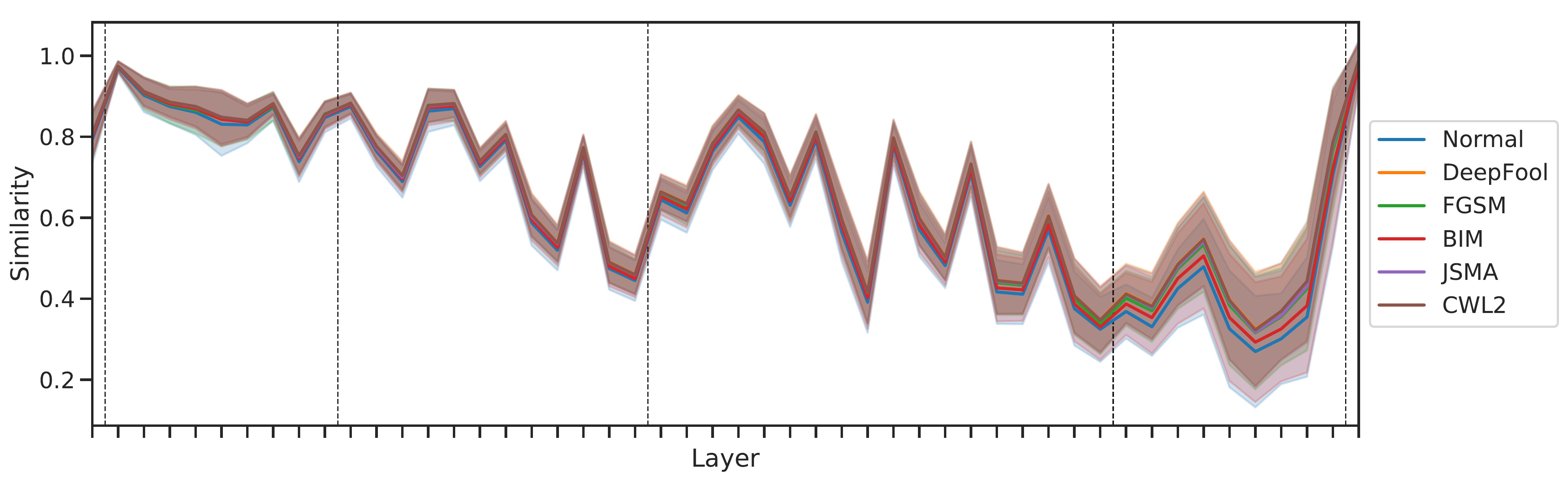}
    \label{fig:resnet_similarity_rank2}
    }
    \hfill
    \subfloat[Rank-2 similarity delta.]{
    \includegraphics[width=.95\columnwidth]{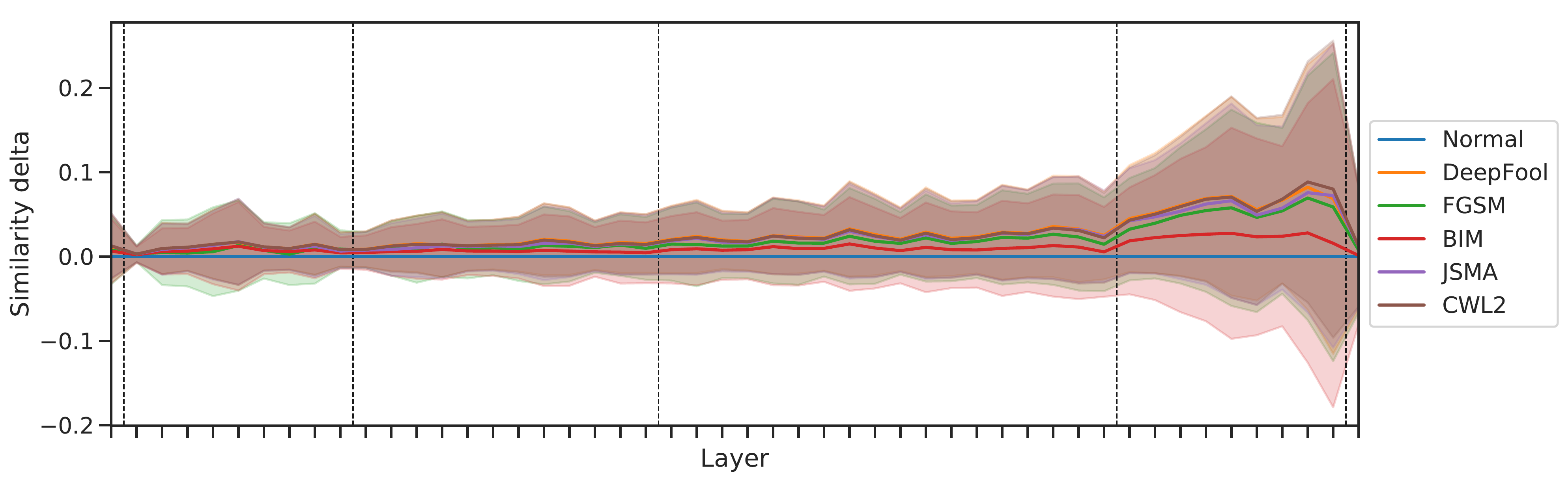}
    \label{fig:resnet_similarity_delta_rank2}
    }
    \caption{Distribution of per-layer similarity for ResNet-50 on ImageNet. Each line plot represents the mean of each kind of adversarial examples' similarity, with the same-color band around to show the standard deviation. The dashed lines indicate that down-sampling is performed in the next layer.}
    \label{fig:imagenet_similarity}
\end{figure*}

Per-layer similarity distribution for ResNet-50 on ImageNet is shown in \Fig{fig:imagenet_similarity}. Similar to AlexNet, adversarial images lead to lower rank-1 similarity and higher rank-2 similarity compared with normal images. Furthermore, corresponding to AlexNet's FC layers, the largest similarity delta is also located in last several layers.

\subsection{Weight-based Defense Model}

\begin{figure}[t]
    \centering
    \begin{minipage}[t]{0.60\linewidth}
        \centering
        \includegraphics[width=.95\columnwidth]{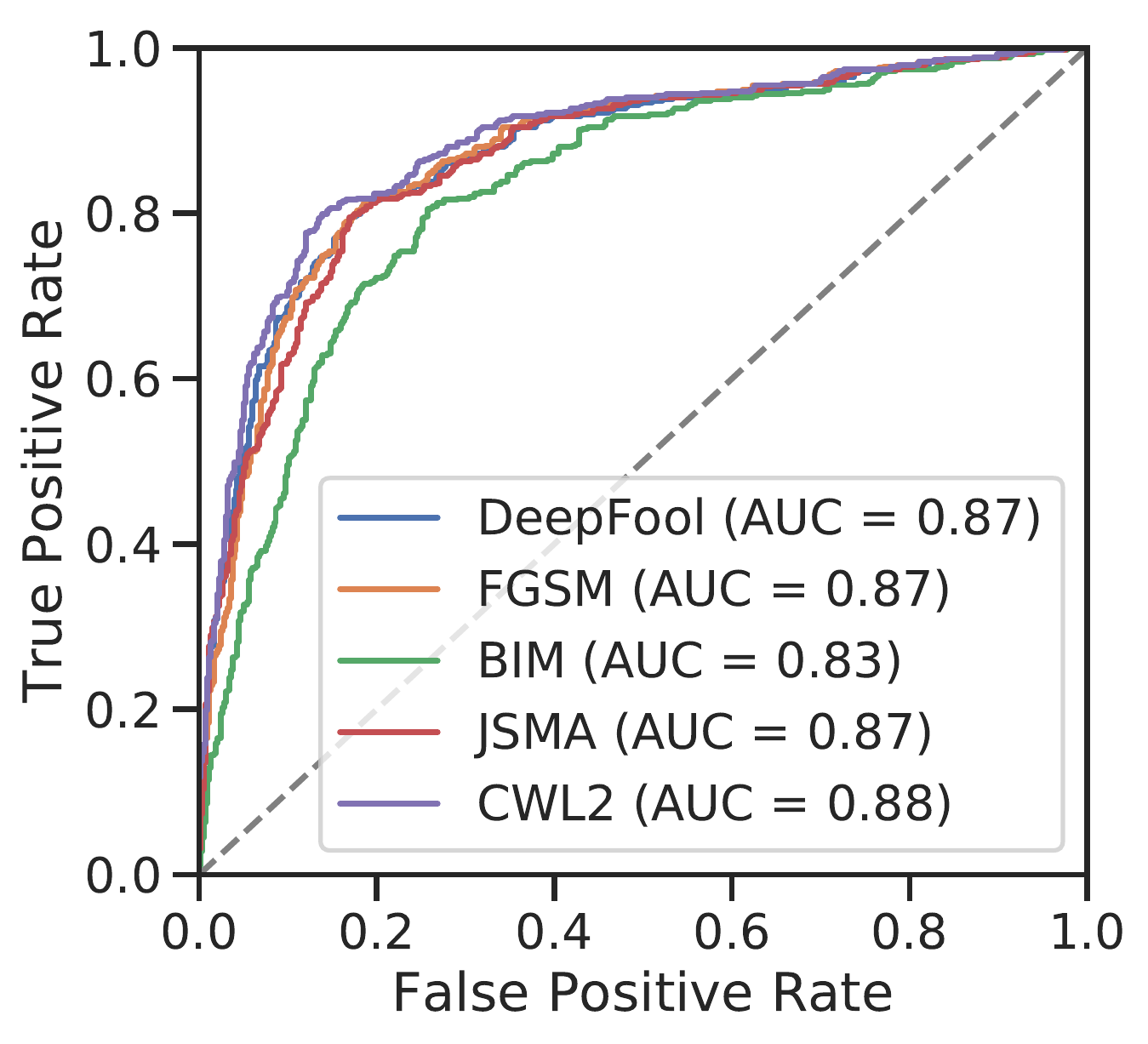}
        \caption{ROC for AlexNet on ImageNet with weight-based joint similarity.}
        \label{fig:roc_confidence}
    \end{minipage}
\end{figure}

For adversarial detection, we can use information from model weights as alternative of synapses. By calculating image-class path similarity from weights in effective path instead, i.e., let $J_{\mathcal{P}}^{l} = |\mathcal{W}^{l} \cap \tilde{\mathcal{W}}_{p}^{l}|/|\mathcal{W}^{l} |$ for layer $l$, we obtain weight-based joint similarity. The detection result using weight-based defense metric for AlexNet is shown in \Fig{fig:roc_confidence}, which indicates that it achieves as high accuracy as the synapse-based metric.

\subsection{Adversarial Sample Visualization}

\begin{figure*}[t]
    \centering
    \subfloat[Effective path similarity.]{
        \includegraphics[trim=40mm 20mm 40mm 20mm,clip,width=.66\columnwidth]{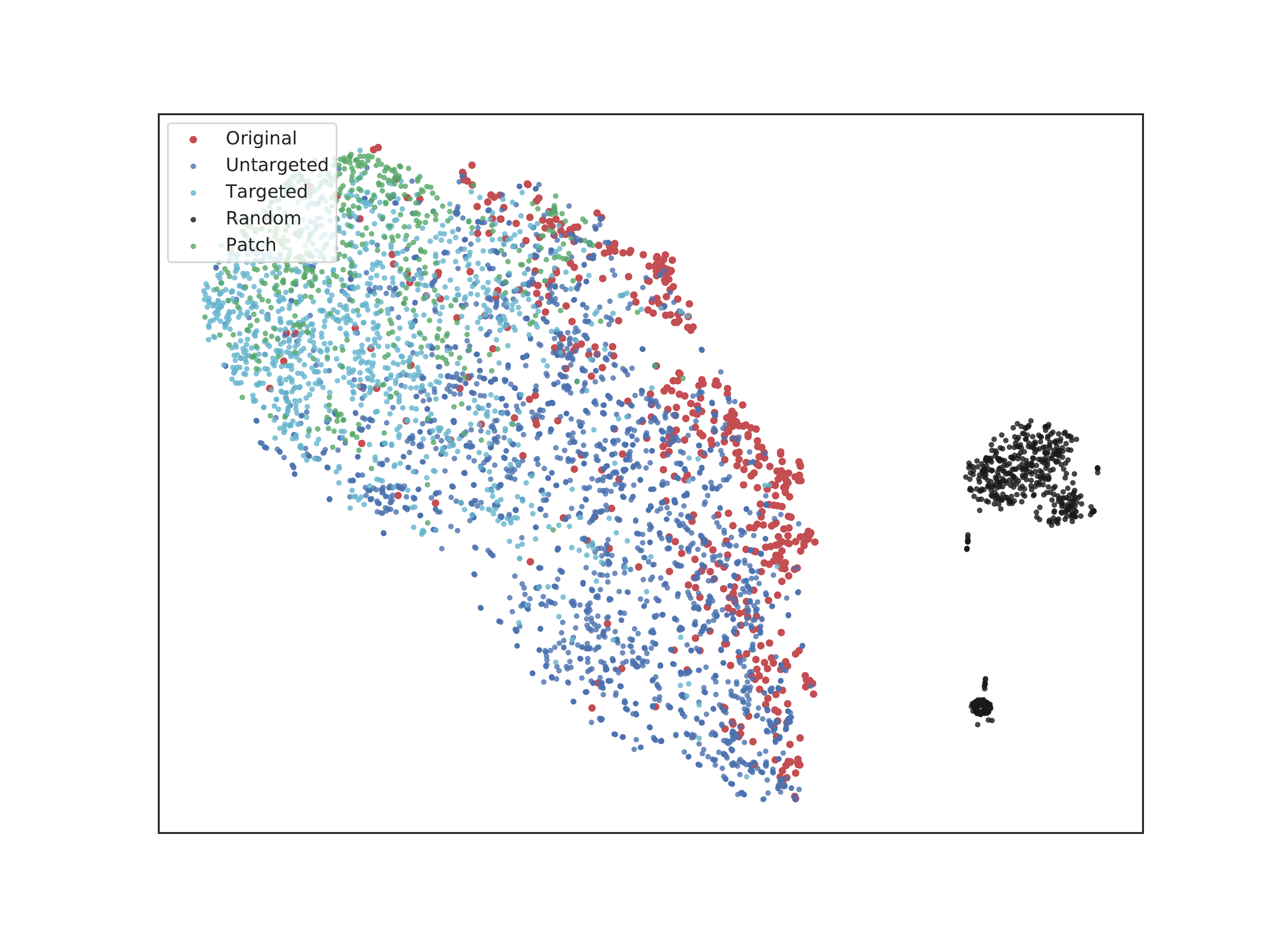}
        \label{fig:visualization_ours}
    }%
    \subfloat[CDRP.]{
        \includegraphics[trim=40mm 20mm 40mm 20mm,clip,width=.66\columnwidth]{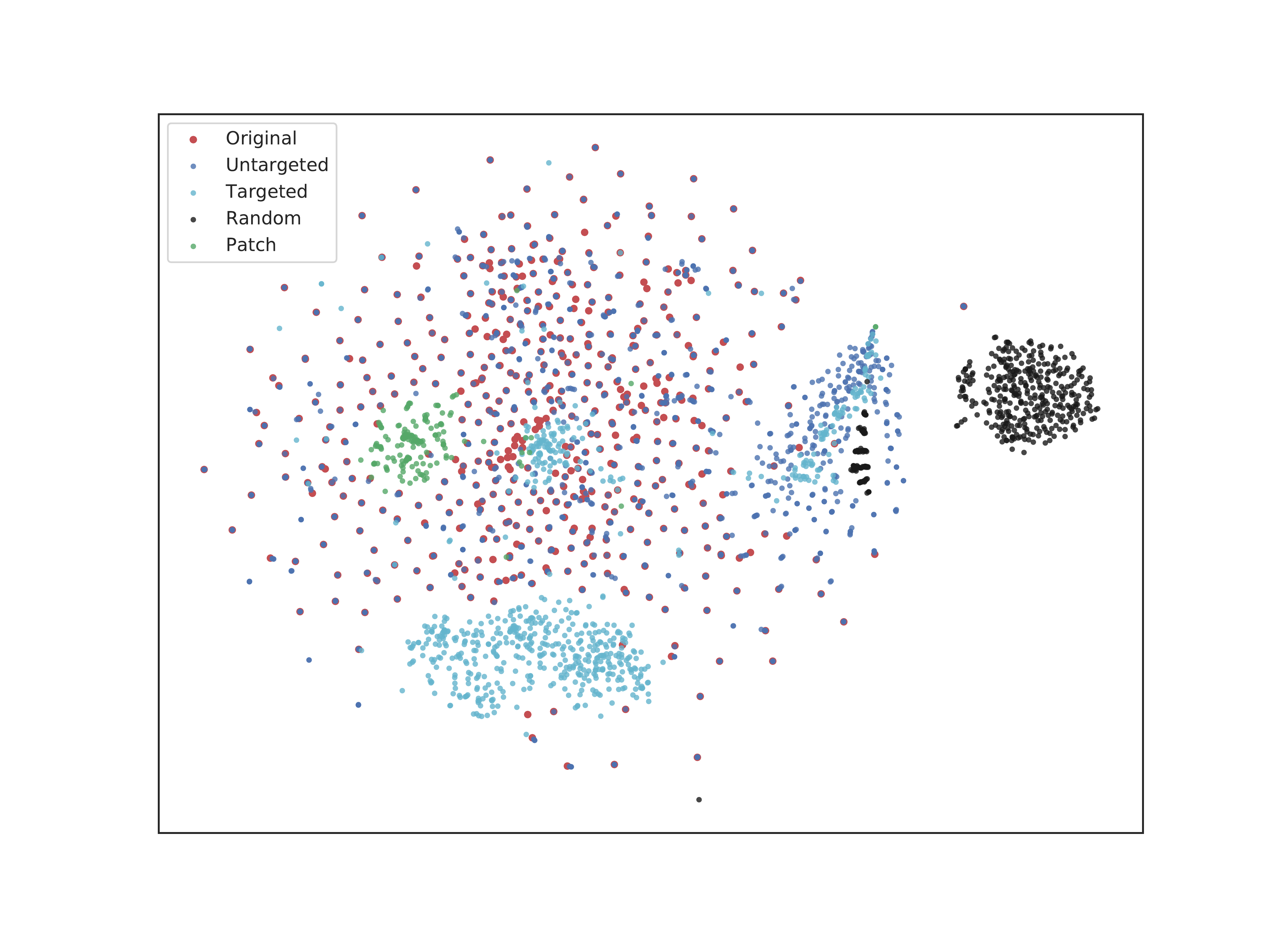}
        \label{fig:visualization_cdrp}
    }
    \subfloat[ResNet-50.]{
        \includegraphics[trim=40mm 20mm 40mm 20mm,clip,width=.66\columnwidth]{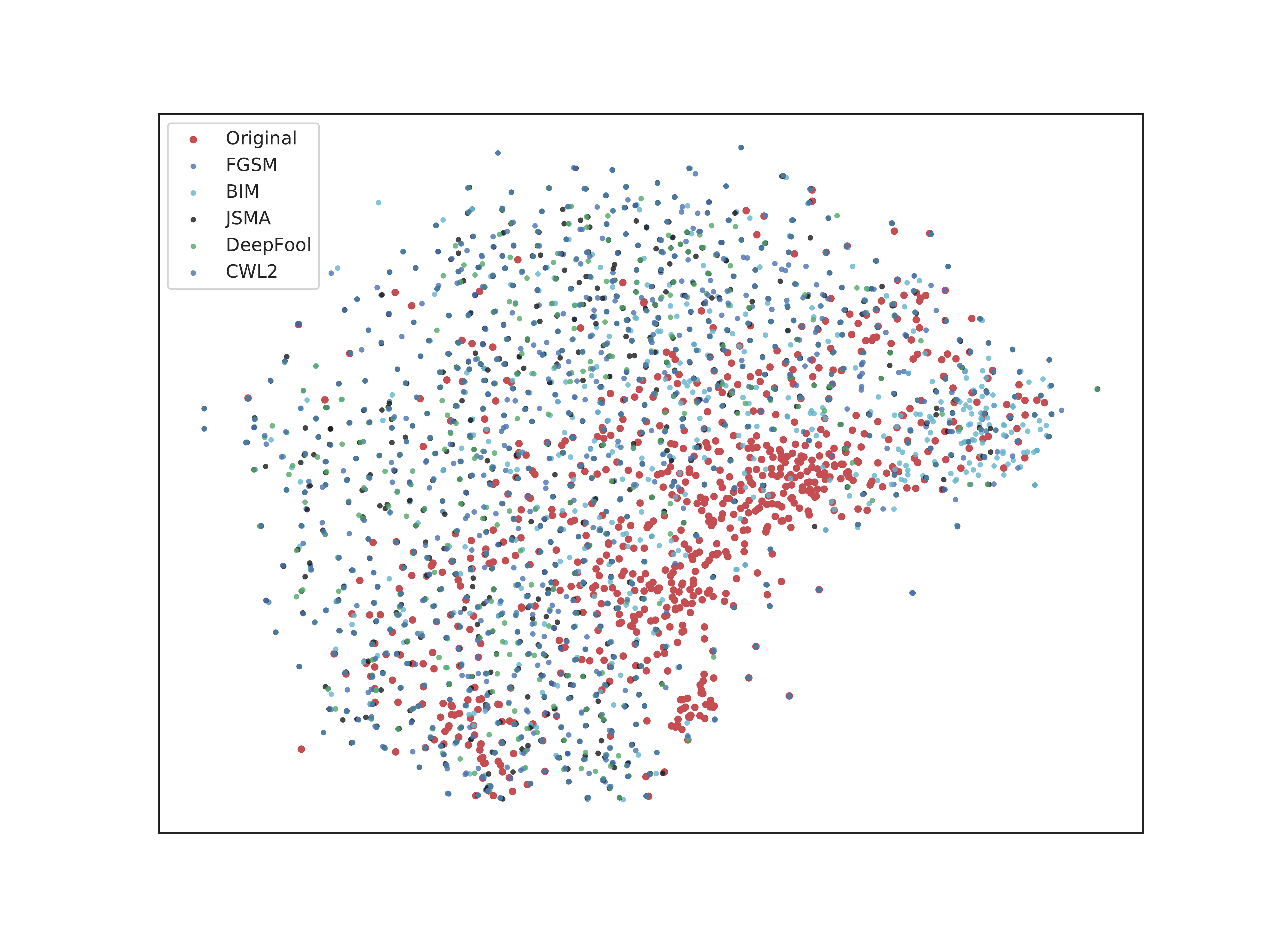}
        \label{fig:visualization_resnet_50}
    }
    \caption{t-SNE 2D embedding of original images and adversarial images from different attacks. Each point stands for an image. The first two pictures show results on AlexNet, while the last one show effective path's result on ResNet-50. For brevity, FGSM, BIM, JSMA, DeepFool, and CWL2 are grouped into untargeted attacks. Targeted version of FGSM and CWL2 are also grouped into targeted attacks.}
    \label{fig:density_similarity_training}
\end{figure*}

In this section, we use t-SNE to visualize effective path similarity and CDRP, which provide an intuitive way to show the adversarial sample detection ability of both methods.
For consistency with our defense model, we use rank-1 and rank-2 effective path similarity as input features of t-SNE. \Fig{fig:visualization_ours} and \Fig{fig:visualization_cdrp} show the t-SNE 2D embedding of effective path similarity and CDRP on AlexNet respectively.
For our method, random images are in two dissociative clusters.
In the cluster contains all other images, original images are located in the edge of right side of the cluster, which is partly separable with other adversarial images.
Adversarial images from untargeted attacks locate in the nearest area beside original examples, which is coincident with the fact that untargeted ones contain the smallest perturbations among these adversarial examples.
On the other hand, adversarial images from targeted attacks and patch attack are far away from original images.
All the above shows that effective path similarity catches the difference between normal images and adversarial images from multiple attacks.
For CDRP, the location of original images and adversarial images from untargeted attacks are almost co-located, which indicates that it fails to catch the difference between them and leads to its low detection accuracy of untargeted attacks.
\Fig{fig:visualization_resnet_50} shows the t-SNE 2D embedding of effective path similarity on ResNet-50.
The boundary between original images and adversarial images is fuzzier than that on AlexNet, but original images still have different distribution compared with adversarial images.

\begin{figure}[t]
    \centering
    \includegraphics[trim=0mm 0mm 0mm 0mm,clip,width=\linewidth]{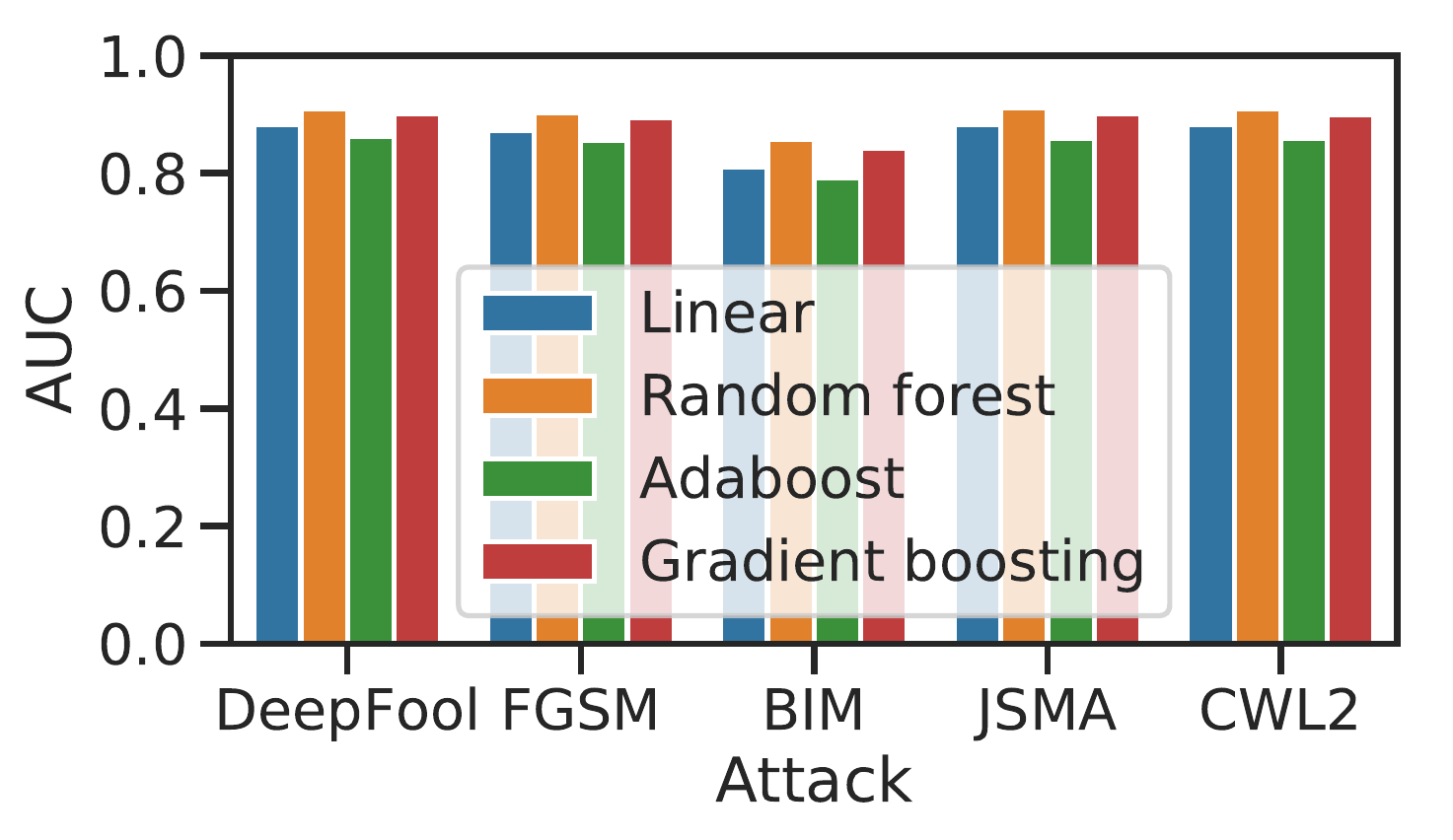}
    \caption{Detection accuracy comparison under different defense models.}
    \label{fig:cdrp_compare}
\end{figure}

\begin{figure}[t]
    \vspace*{-0.3cm}
    \centering
    \begingroup
    \captionsetup[subfigure][t]{width=.49\linewidth}
    \subfloat[Linear model.]{
        \includegraphics[trim=0mm 2mm 52mm 0mm,clip,height=.36\linewidth]{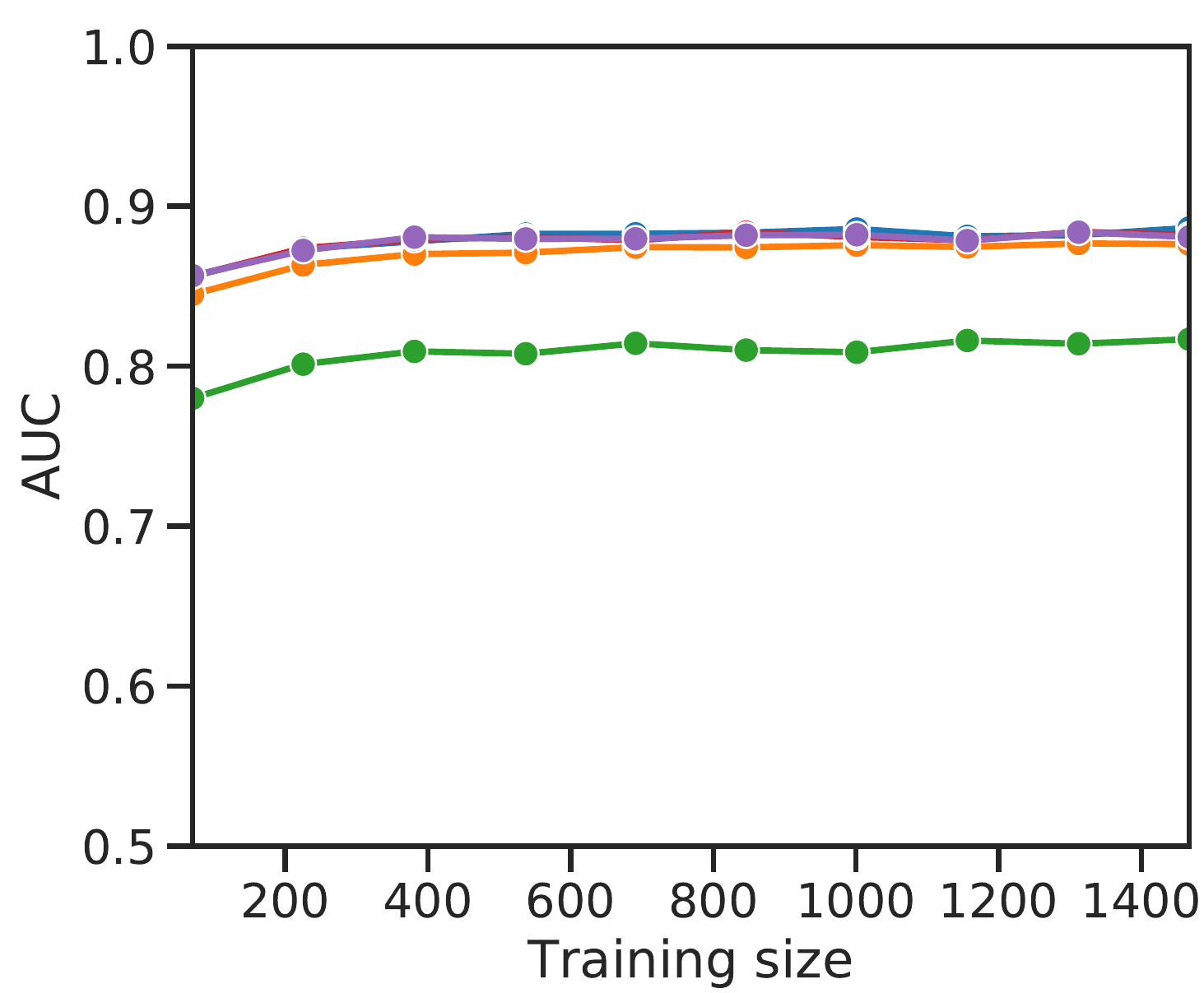}
        \label{fig:training_set_size_linear}
    }%
    \subfloat[Random forest.]{
        \includegraphics[trim=0mm 2mm 0mm 0mm,clip,height=.36\linewidth]{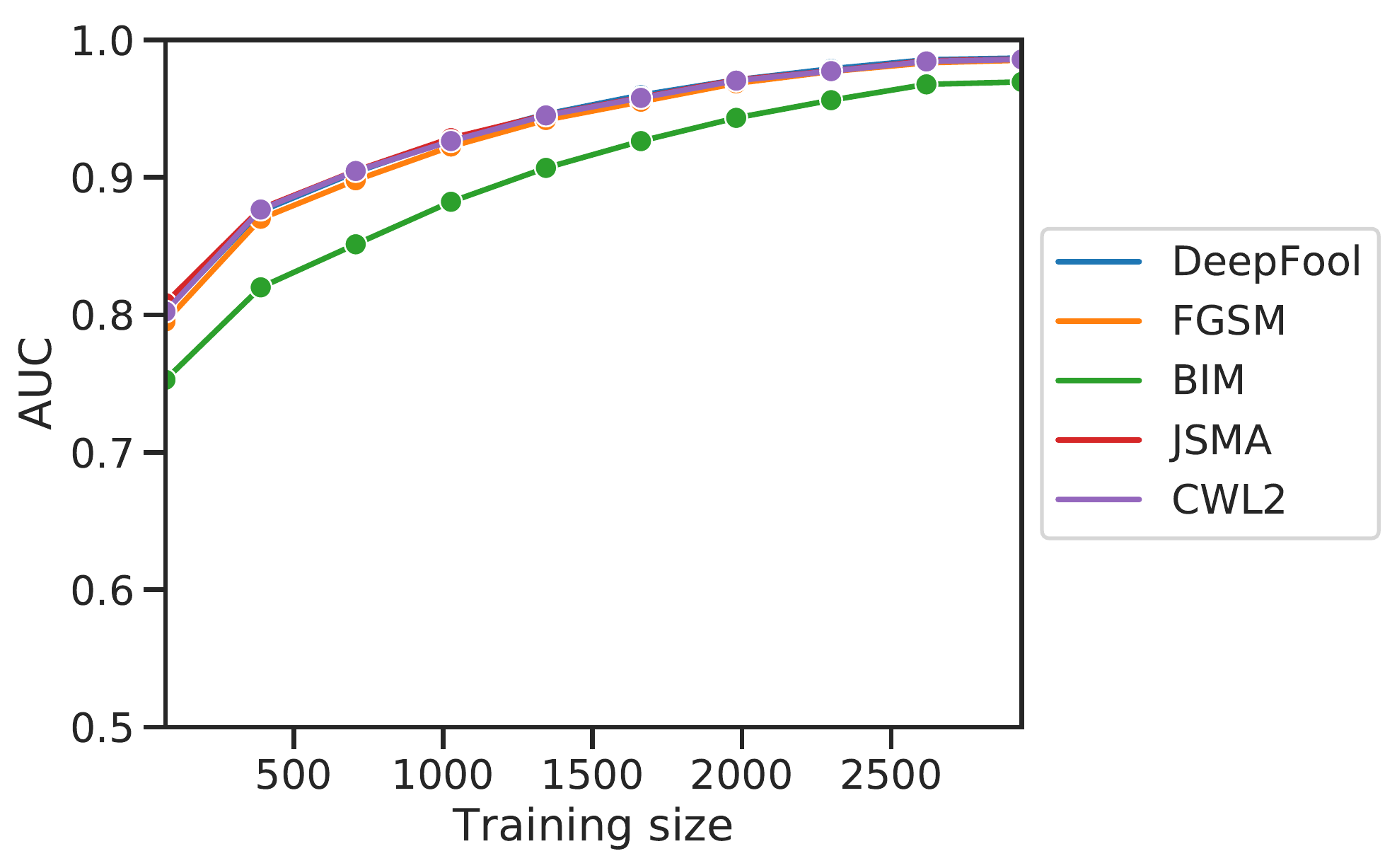}
        \label{fig:training_set_size_rf}
    }%
    \vspace*{-0.3cm}
    \endgroup
    \caption{Impact of training set size on the AUC.}
    \label{fig:training_set_size}
    \vspace*{-0.3cm}
    \end{figure}

    \begin{figure}[t]
        \vspace*{-0.3cm}
        \centering
        \begingroup
        \captionsetup[subfigure][t]{width=.49\linewidth}
        \subfloat[Linear model.]{
            \includegraphics[trim=0mm 2mm 52mm 0mm,clip,height=.36\linewidth]{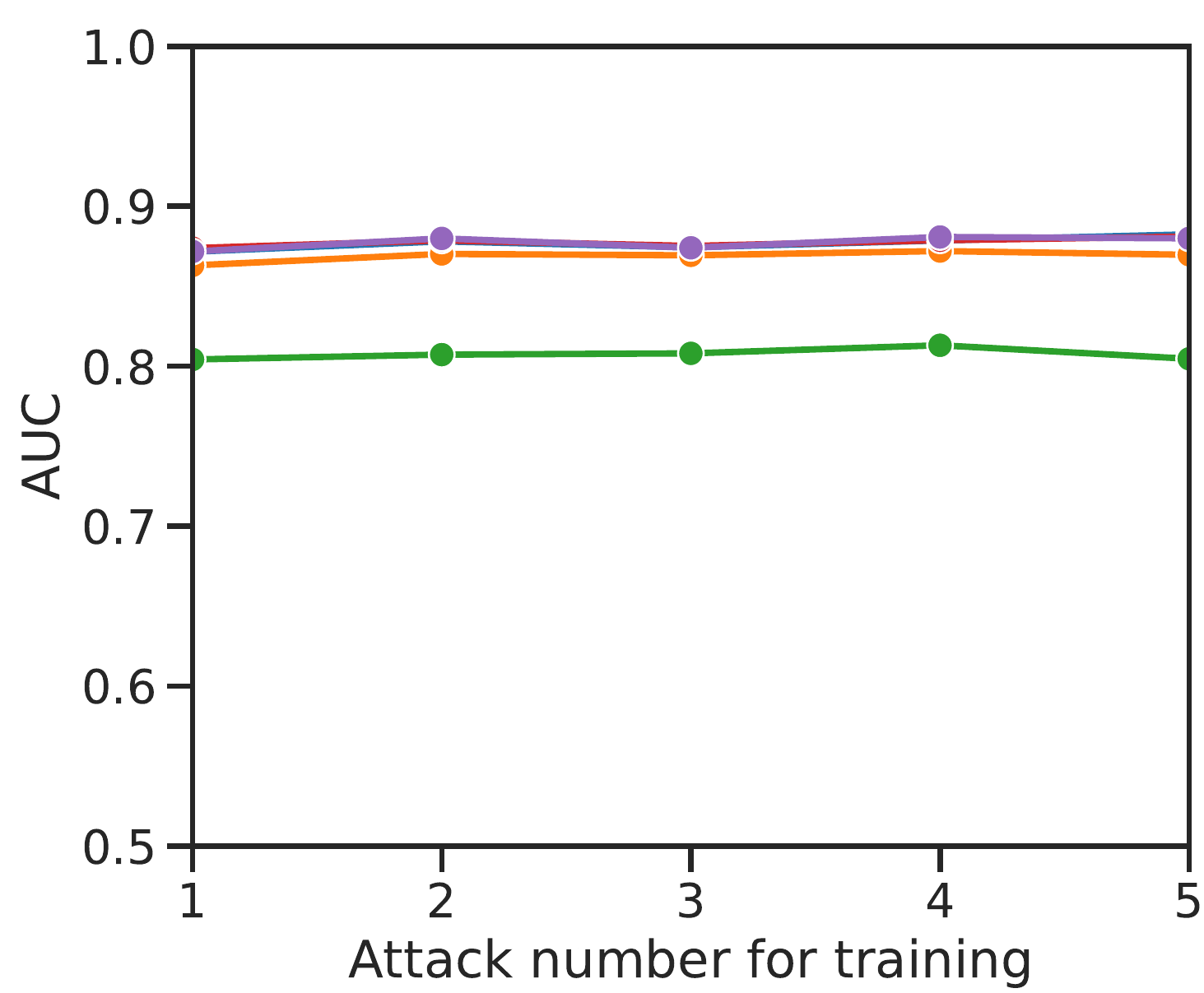}
            \label{fig:attack_number_linear}
        }%
        \subfloat[Random forest.]{
            \includegraphics[trim=0mm 2mm 0mm 0mm,clip,height=.36\linewidth]{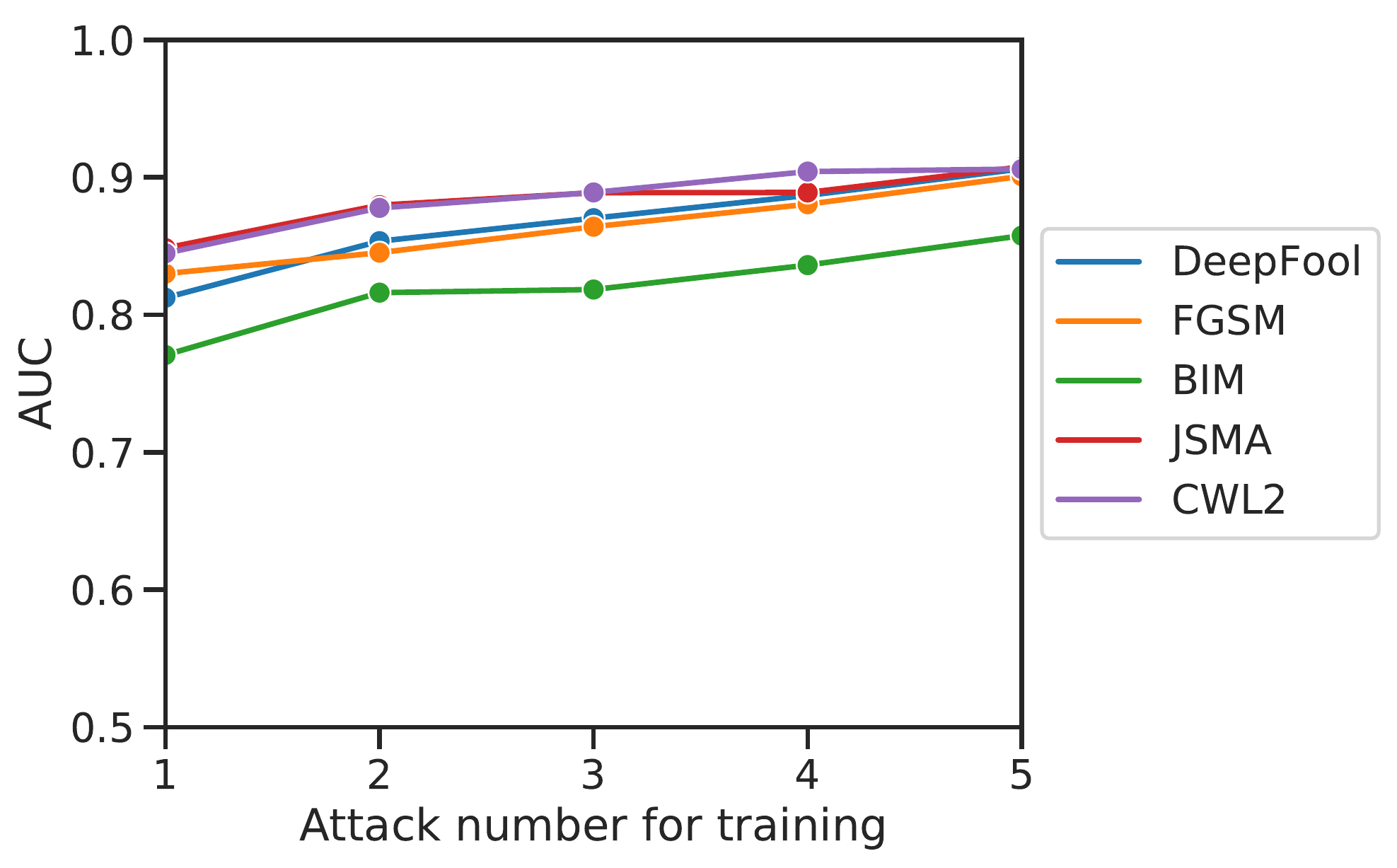}
            \label{fig:attack_number_rf}
        }%
        \vspace*{-0.3cm}
        \endgroup
        \caption{Impact of attack number in the training set.}
        \label{fig:attack_number}
        \vspace*{-0.3cm}
    \end{figure}

    \begin{figure}[t]
        \vspace*{-0.3cm}
        \centering
        \begingroup
        \captionsetup[subfigure][t]{width=.49\linewidth}
        \subfloat[Linear model.]{
            \includegraphics[trim=0mm 2mm 52mm 0mm,clip,height=.36\linewidth]{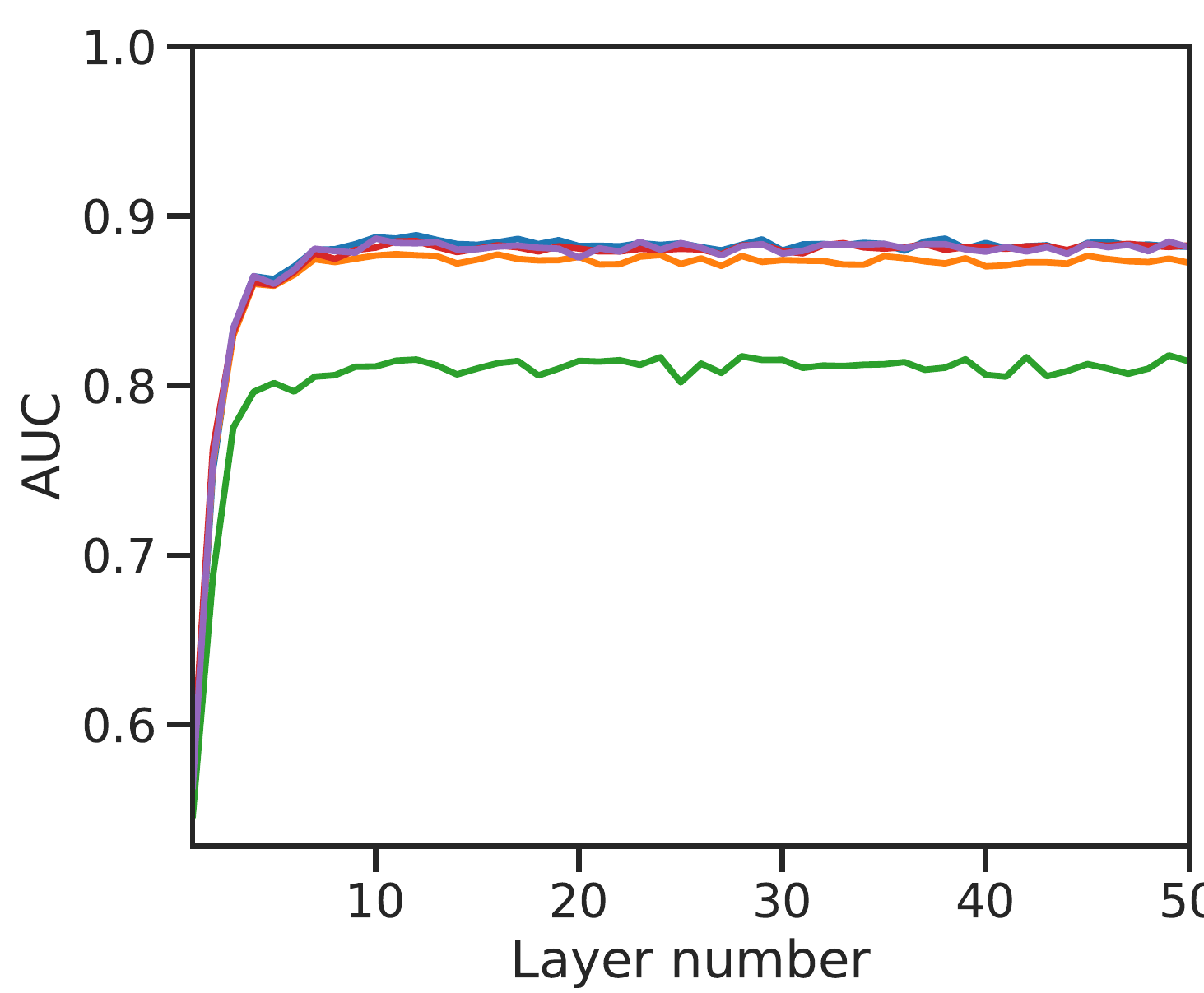}
            \label{fig:layer_number_linear}
        }%
        \subfloat[Random forest.]{
            \includegraphics[trim=0mm 2mm 0mm 0mm,clip,height=.36\linewidth]{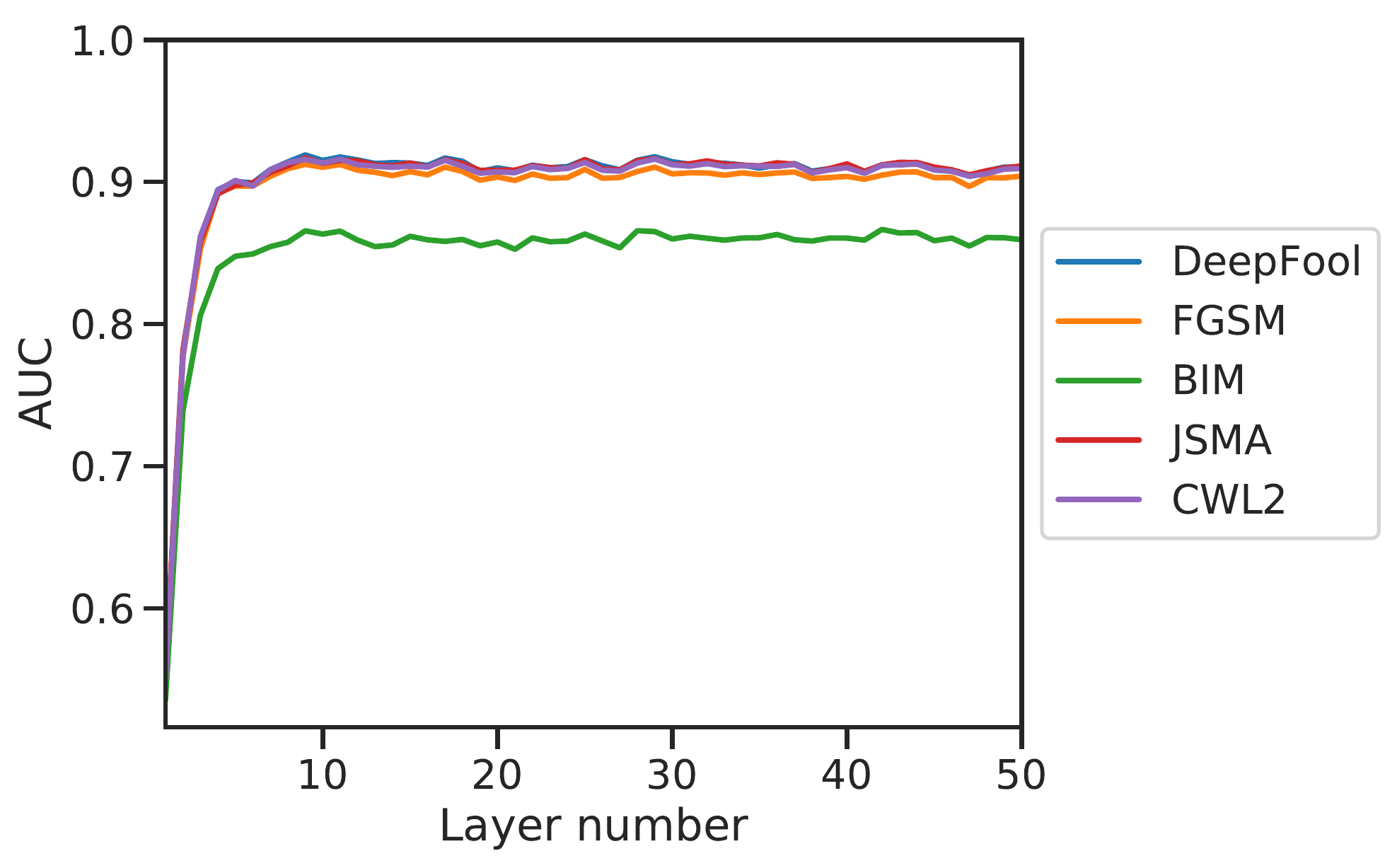}
            \label{fig:layer_number_rf}
        }%
        \vspace*{-0.3cm}
        \endgroup
        \caption{Effective path layer number impact on AUC.}
        \label{fig:layer_number}
    \end{figure}

    \begin{figure*}[t]
        \centering
        \subfloat[Synapses in the path.]{
            \includegraphics[width=.95\columnwidth]{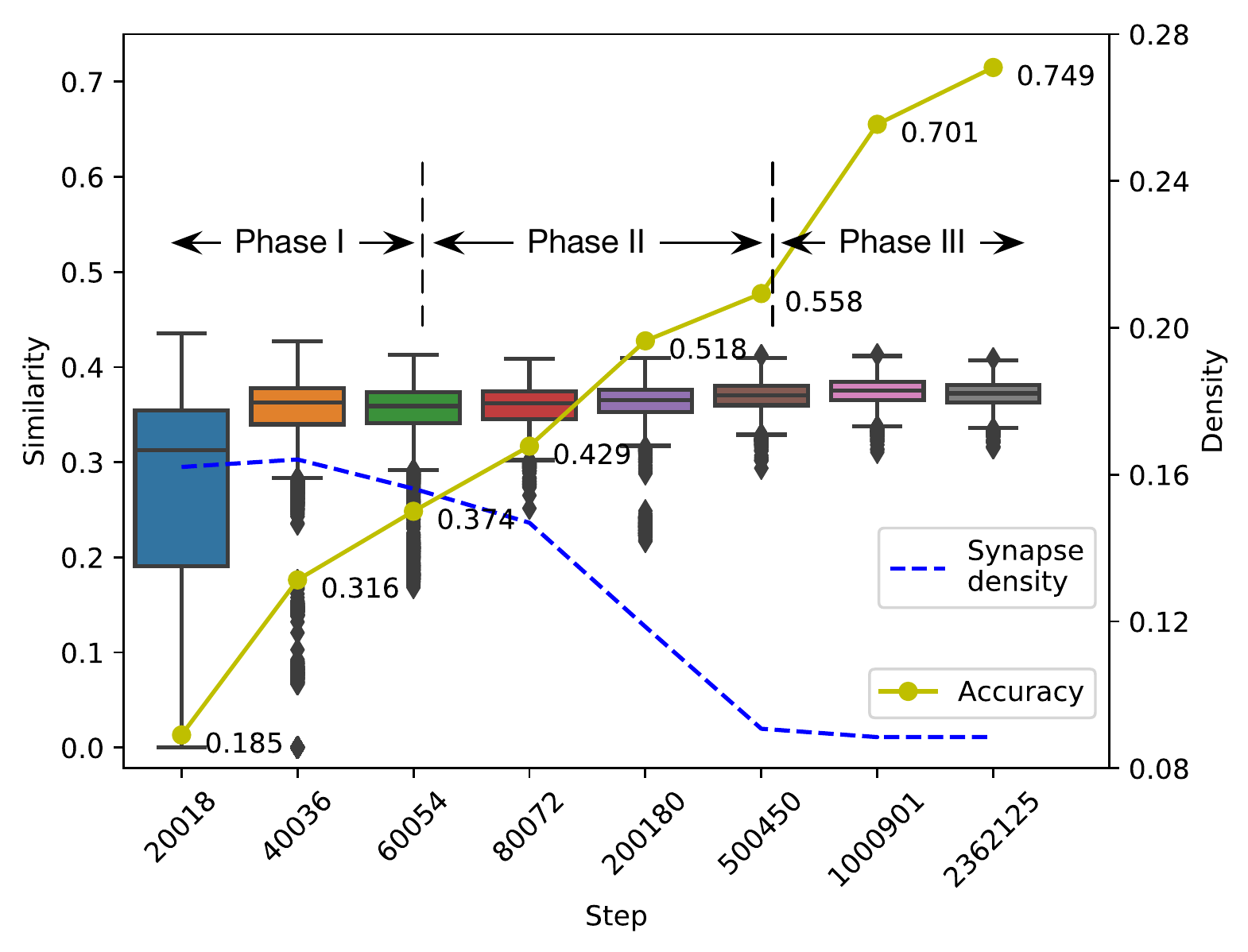}
            \label{fig:synapse_density_similarity_training}
        }%
        \subfloat[Weights in the path.]{
            \includegraphics[width=.95\columnwidth]{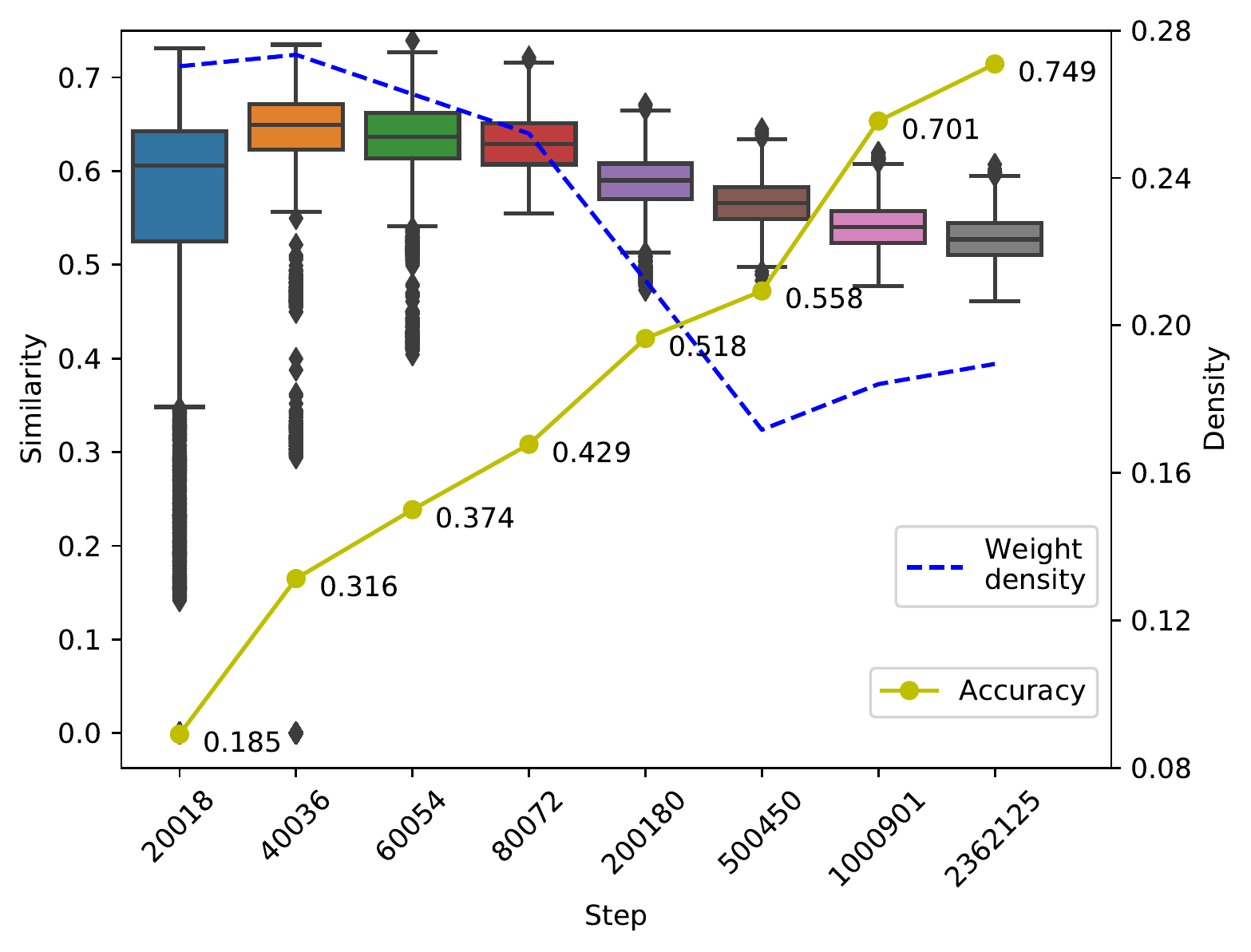}
            \label{fig:weight_density_similarity_training}
        }
        \caption{Effective path's density and class-wise path similarity in the training process.}
        \label{fig:density_similarity_training}
    \end{figure*}
    
    \begin{figure*}[t]
        \begin{minipage}[c]{0.30\linewidth}
            \centering
            \includegraphics[width=.95\columnwidth]{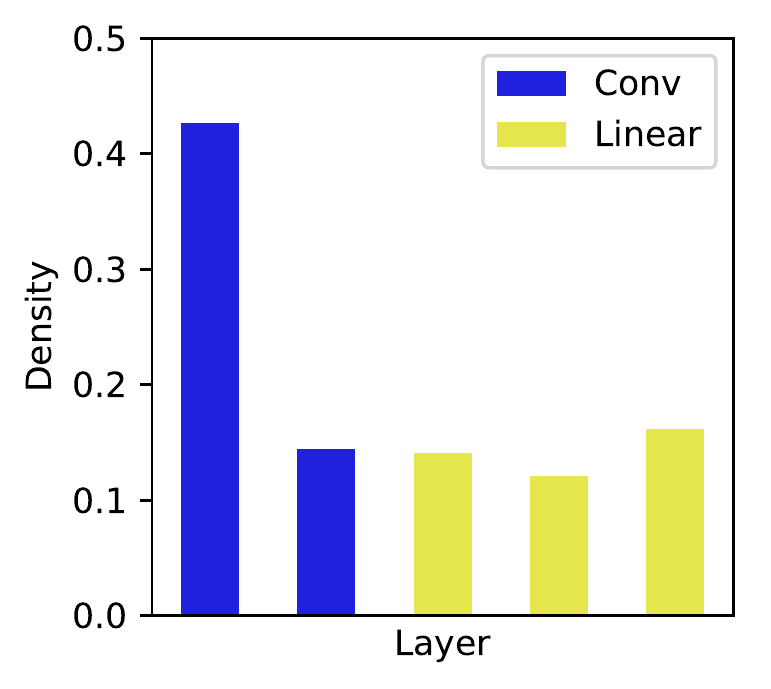}
            \caption{Per-layer density of effective paths in LeNet.}
            \label{fig:per_layer_density_lenet}
        \end{minipage}\hfill
        \begin{minipage}[c]{0.68\linewidth}
            \centering
            \includegraphics[width=.95\columnwidth]{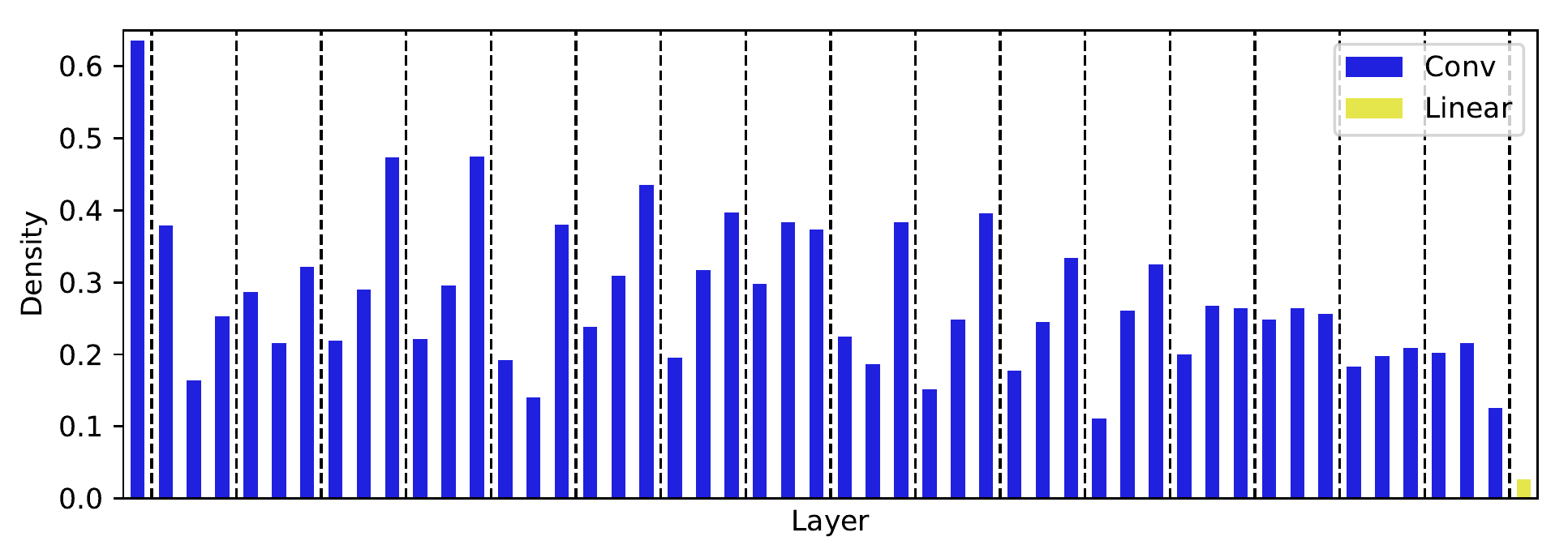}
            \caption{Per-layer density of effective paths in ResNet-50. Layers in ResNet-50 is organized into 3-layer bottleneck blocks, which is split by dashed lines.}
            \label{fig:per_layer_density_resnet_50}
        \end{minipage}
    \end{figure*}

\section{Evaluation on ResNet-50}

In this section, we show further evaluation results on ResNet-50 which is not included in the submitted paper.

\subsection{Detection Accuracy}

\Fig{fig:cdrp_compare} shows the detection accuracy of effective path under different defense models. For all of the mentioned attacks, we find that random forest performs the best while the linear model performs worst among all models. However, the gap between random forest and linear model is very small, which indicates that effective path can achieve comparable detection performance using simple and highly interpretable way on deep and complex networks like ResNet-50.

\subsection{Training Size}

We choose the linear model and random forest model as representations to study how the size of the training set impacts the detection accuracy on ResNet-50. 
\Fig{fig:training_set_size} shows the difference between the simple linear model and complex random forest model.
For the linear model, the detection accuracy stabilizes with a small number of training samples (around 400 images).
For the random forest model, the detector requires much larger training set (around 2500 images), meanwhile achieves much better detection accuracy, which indicates that random forest model can utilize more features in the effective path. Compared with AlexNet, both linear model and random forest model requires more training samples to be stable, which can be attributed to the fact that ResNet-50 has much more layers than AlexNet that provide useful features for detection.

\subsection{Generalizability}

\Fig{fig:attack_number} shows the experiment results of generalizability on ResNet-50 when adding the adversarial samples in the order of legend shown in the right. For both linear model and random forest model, our work generalizes well to unseen attacks because effective path captures their common behavior. Compared with AlexNet, our method achieves the same level of generalizability on much more complex networks.

\subsection{Layer Sensitivity}

We study the layer number's impact on the adversarial sample detection accuracy on ResNet-50 and show the result in \Fig{fig:layer_number}.
For both linear model and random forest, we observe that the AUC performance for all attacks saturates within 10 layers, which is a small portion of the whole 50 layers. This layer sensitivity insight leads to the significant speedup of extraction by only extracting just enough layers instead of all layers, which is described in the submitted paper.

\section{Further Study of Neural Network Interpretability}
\label{sec:further_interpret}

We further study how the training process and different network structure impacts the path specialization.
In the next, we study how the training process and different network structures impact the path specialization.

\subsection{Training Process}

We study how the training process transforms a randomized network to the final state from the perspective of the effective path. Specifically, we extract the effective path for each class at different training stages. Through the analysis, we find that the training process contains three distinctive phases with different path's density and similarity trend, which share similar insights from the previous work using information bottleneck theory to explain training process~\cite{information_bottleneck}.

\Fig{fig:density_similarity_training} shows training process for ResNet. We choose different stages in training and show the class-wise path similarity in the form of box-plot, on top of which we also overlay the path density and prediction accuracy. In the first phase, the density of synapses and weights in the effective paths stays the same while their similarity increases. In the beginning, the network is in a randomized state and simply tries to memorize the input data.

In the second phase, the density of both synapses and weights decrease rapidly. The similarity of the synapses stays relatively the same while the similarity of weights decreases. In this phase, the network mainly performs compression, and the path specialization mainly manifests in the form of weights. In other words, the network tries to use class-specific features extracted by different convolutional filters to increase the specialization degree.

In the third phase, the synapse density stops to decrease but weight density starts to increase. Meanwhile, weight similarity continues to decrease. In this phase, the network compression stops and mainly relies on path specialization (via weight) to increase the prediction accuracy. The path specialization even causes the weight density increases a bit.

In summary, we find that the training process contains mainly three phases, the first two of which conforms to the memorization and compression phase identified by the prior work~\cite{information_bottleneck}. The second phase performs compression (less density) and path specialization (less similarity), while the third phase mainly includes the path specialization. After these phases, the network is transformed into a state with sparse and distinctive paths with great inference capability.

\subsection{Network Structure}

After establishing the effective path as a great indicator of the neural network's inference performance, we study how the network structure affects the effective path characteristics.

\Fig{fig:per_layer_density_lenet} shows the per-layer path density in LeNet. We observe the first convolutional layer has a much higher density compared to the following layers, which matches with the established knowledge that the shallow layers in a CNN extract high-level features that are shared by different classes.

CNN designers have found using a deeper network can increase the prediction accuracy to a certain degree. However, the accuracy stops to increase after a certain number of layers owing to the vanishing gradient in the training process. As such, the ResNet structure with skip connection was proposed to overcome this difficulty. \Fig{fig:per_layer_density_resnet_50} shows the per-layer path density for ResNet-50. Not only the first two layers still have higher density, but also layers before a skip connection also have high density. This suggests that skip connection helps not only the gradient propagation but also the effective paths formation. In the end, ResNet is able to converge and achieve great prediction performance.

%% file: ms.bbl
\begin{thebibliography}{10}\itemsep=-1pt

\bibitem{AbdelHamid2014ConvolutionalNN}
Ossama Abdel-Hamid, Abdel rahman Mohamed, Hui Jiang, Li Deng, Gerald Penn, and
  Dong Yu.
\newblock Convolutional neural networks for speech recognition.
\newblock {\em IEEE/ACM Transactions on Audio, Speech, and Language
  Processing}, 22:1533--1545, 2014.

\bibitem{DBLP:conf/icml/AthalyeC018}
Anish Athalye, Nicholas Carlini, and David~A. Wagner.
\newblock Obfuscated gradients give a false sense of security: Circumventing
  defenses to adversarial examples.
\newblock In {\em Proceedings of the 35th International Conference on Machine
  Learning, {ICML} 2018, Stockholmsm{\"{a}}ssan, Stockholm, Sweden, July 10-15,
  2018}, pages 274--283, 2018.

\bibitem{Ball:1996:EPP:243846.243857}
Thomas Ball and James~R. Larus.
\newblock Efficient path profiling.
\newblock In {\em Proceedings of the 29th Annual ACM/IEEE International
  Symposium on Microarchitecture}, MICRO 29, pages 46--57, Washington, DC, USA,
  1996. IEEE Computer Society.

\bibitem{DBLP:journals/corr/abs-1712-09665}
Tom~B. Brown, Dandelion Man{\'{e}}, Aurko Roy, Mart{\'{\i}}n Abadi, and Justin
  Gilmer.
\newblock Adversarial patch.
\newblock {\em CoRR}, abs/1712.09665, 2017.

\bibitem{buckman2018thermometer}
Jacob Buckman, Aurko Roy, Colin Raffel, and Ian Goodfellow.
\newblock Thermometer encoding: One hot way to resist adversarial examples.
\newblock In {\em International Conference on Learning Representations}, 2018.

\bibitem{DBLP:journals/corr/CarliniW16a}
Nicholas Carlini and David~A. Wagner.
\newblock Towards evaluating the robustness of neural networks.
\newblock {\em CoRR}, abs/1608.04644, 2016.

\bibitem{DBLP:journals/corr/abs-1711-08478}
Nicholas Carlini and David~A. Wagner.
\newblock Magnet and "efficient defenses against adversarial attacks" are not
  robust to adversarial examples.
\newblock {\em CoRR}, abs/1711.08478, 2017.

\bibitem{s.2018stochastic}
Guneet~S. Dhillon, Kamyar Azizzadenesheli, Jeremy~D. Bernstein, Jean Kossaifi,
  Aran Khanna, Zachary~C. Lipton, and Animashree Anandkumar.
\newblock Stochastic activation pruning for robust adversarial defense.
\newblock In {\em International Conference on Learning Representations}, 2018.

\bibitem{adas}
Ivan Dynov.
\newblock {Is Deep Learning Really the Solution for Everything in Self-Driving
  Cars?}
\newblock \url{http://bit.ly/ivan_dynov_adas}, 2016.

\bibitem{eykholt2018robust}
Kevin Eykholt, Ivan Evtimov, Earlence Fernandes, Bo Li, Amir Rahmati, Chaowei
  Xiao, Atul Prakash, Tadayoshi Kohno, and Dawn Song.
\newblock Robust physical-world attacks on deep learning visual classification.
\newblock In {\em Proceedings of the IEEE Conference on Computer Vision and
  Pattern Recognition}, pages 1625--1634, 2018.

\bibitem{fiore2017using}
Ugo Fiore, Alfredo De~Santis, Francesca Perla, Paolo Zanetti, and Francesco
  Palmieri.
\newblock Using generative adversarial networks for improving classification
  effectiveness in credit card fraud detection.
\newblock {\em Information Sciences}, 2017.

\bibitem{tesla_crash}
Jordan Golson.
\newblock {Tesla driver killed in crash with Autopilot active, NHTSA
  investigating}.
\newblock \url{http://bit.ly/tesla_crash}, 2016.

\bibitem{DBLP:journals/corr/GoodfellowSS14}
Ian~J. Goodfellow, Jonathon Shlens, and Christian Szegedy.
\newblock Explaining and harnessing adversarial examples.
\newblock {\em CoRR}, abs/1412.6572, 2014.

\bibitem{guo2018countering}
Chuan Guo, Mayank Rana, Moustapha Cisse, and Laurens van~der Maaten.
\newblock Countering adversarial images using input transformations.
\newblock In {\em International Conference on Learning Representations}, 2018.

\bibitem{DBLP:journals/corr/HeZRS15}
Kaiming He, Xiangyu Zhang, Shaoqing Ren, and Jian Sun.
\newblock Deep residual learning for image recognition.
\newblock {\em CoRR}, abs/1512.03385, 2015.

\bibitem{DBLP:conf/nips/KrizhevskySH12}
Alex Krizhevsky, Ilya Sutskever, and Geoffrey~E. Hinton.
\newblock Imagenet classification with deep convolutional neural networks.
\newblock In {\em Advances in Neural Information Processing Systems 25: 26th
  Annual Conference on Neural Information Processing Systems 2012. Proceedings
  of a meeting held December 3-6, 2012, Lake Tahoe, Nevada, United States.},
  pages 1106--1114, 2012.

\bibitem{Krizhevsky:2012:ICD:2999134.2999257}
Alex Krizhevsky, Ilya Sutskever, and Geoffrey~E. Hinton.
\newblock Imagenet classification with deep convolutional neural networks.
\newblock In {\em Proceedings of the 25th International Conference on Neural
  Information Processing Systems - Volume 1}, NIPS'12, pages 1097--1105, USA,
  2012. Curran Associates Inc.

\bibitem{DBLP:journals/corr/KurakinGB16}
Alexey Kurakin, Ian~J. Goodfellow, and Samy Bengio.
\newblock Adversarial examples in the physical world.
\newblock {\em CoRR}, abs/1607.02533, 2016.

\bibitem{DBLP:journals/corr/KurakinGB16a}
Alexey Kurakin, Ian~J. Goodfellow, and Samy Bengio.
\newblock Adversarial machine learning at scale.
\newblock {\em CoRR}, abs/1611.01236, 2016.

\bibitem{DBLP:journals/neco/LeCunBDHHHJ89}
Yann LeCun, Bernhard~E. Boser, John~S. Denker, Donnie Henderson, Richard~E.
  Howard, Wayne~E. Hubbard, and Lawrence~D. Jackel.
\newblock Backpropagation applied to handwritten zip code recognition.
\newblock {\em Neural Computation}, 1(4):541--551, 1989.

\bibitem{DBLP:journals/corr/MadryMSTV17}
Aleksander Madry, Aleksandar Makelov, Ludwig Schmidt, Dimitris Tsipras, and
  Adrian Vladu.
\newblock Towards deep learning models resistant to adversarial attacks.
\newblock {\em CoRR}, abs/1706.06083, 2017.

\bibitem{Meng:2017:MTD:3133956.3134057}
Dongyu Meng and Hao Chen.
\newblock Magnet: A two-pronged defense against adversarial examples.
\newblock In {\em Proceedings of the 2017 ACM SIGSAC Conference on Computer and
  Communications Security}, CCS '17, pages 135--147, New York, NY, USA, 2017.
  ACM.

\bibitem{DBLP:journals/corr/MetzenGFB17}
Jan~Hendrik Metzen, Tim Genewein, Volker Fischer, and Bastian Bischoff.
\newblock On detecting adversarial perturbations.
\newblock {\em CoRR}, abs/1702.04267, 2017.

\bibitem{DBLP:journals/corr/Moosavi-Dezfooli15}
Seyed{-}Mohsen Moosavi{-}Dezfooli, Alhussein Fawzi, and Pascal Frossard.
\newblock Deepfool: a simple and accurate method to fool deep neural networks.
\newblock {\em CoRR}, abs/1511.04599, 2015.

\bibitem{DBLP:journals/corr/abs-1708-02582}
Taesik Na, Jong~Hwan Ko, and Saibal Mukhopadhyay.
\newblock Cascade adversarial machine learning regularized with a unified
  embedding.
\newblock {\em CoRR}, abs/1708.02582, 2017.

\bibitem{DBLP:conf/cvpr/NguyenYC15}
Anh~Mai Nguyen, Jason Yosinski, and Jeff Clune.
\newblock Deep neural networks are easily fooled: High confidence predictions
  for unrecognizable images.
\newblock In {\em {IEEE} Conference on Computer Vision and Pattern Recognition,
  {CVPR} 2015, Boston, MA, USA, June 7-12, 2015}, pages 427--436, 2015.

\bibitem{papernot2018cleverhans}
Nicolas Papernot, Fartash Faghri, Nicholas Carlini, Ian Goodfellow, Reuben
  Feinman, Alexey Kurakin, Cihang Xie, Yash Sharma, Tom Brown, Aurko Roy,
  Alexander Matyasko, Vahid Behzadan, Karen Hambardzumyan, Zhishuai Zhang,
  Yi-Lin Juang, Zhi Li, Ryan Sheatsley, Abhibhav Garg, Jonathan Uesato, Willi
  Gierke, Yinpeng Dong, David Berthelot, Paul Hendricks, Jonas Rauber, and
  Rujun Long.
\newblock Technical report on the cleverhans v2.1.0 adversarial examples
  library.
\newblock {\em arXiv preprint arXiv:1610.00768}, 2018.

\bibitem{DBLP:journals/corr/PapernotMJFCS15}
Nicolas Papernot, Patrick~D. McDaniel, Somesh Jha, Matt Fredrikson, Z.~Berkay
  Celik, and Ananthram Swami.
\newblock The limitations of deep learning in adversarial settings.
\newblock {\em CoRR}, abs/1511.07528, 2015.

\bibitem{Pei:2017:DAW:3132747.3132785}
Kexin Pei, Yinzhi Cao, Junfeng Yang, and Suman Jana.
\newblock Deepxplore: Automated whitebox testing of deep learning systems.
\newblock In {\em Proceedings of the 26th Symposium on Operating Systems
  Principles}, SOSP '17, pages 1--18, New York, NY, USA, 2017. ACM.

\bibitem{rauber2017foolbox}
Jonas Rauber, Wieland Brendel, and Matthias Bethge.
\newblock Foolbox: A python toolbox to benchmark the robustness of machine
  learning models.
\newblock {\em arXiv preprint arXiv:1707.04131}, 2017.

\bibitem{samangouei2018defensegan}
Pouya Samangouei, Maya Kabkab, and Rama Chellappa.
\newblock Defense-{GAN}: Protecting classifiers against adversarial attacks
  using generative models.
\newblock In {\em International Conference on Learning Representations}, 2018.

\bibitem{DBLP:journals/corr/abs-1805-09190}
Lukas Schott, Jonas Rauber, Wieland Brendel, and Matthias Bethge.
\newblock Robust perception through analysis by synthesis.
\newblock {\em CoRR}, abs/1805.09190, 2018.

\bibitem{DBLP:journals/corr/abs-1710-10733}
Yash Sharma and Pin{-}Yu Chen.
\newblock Attacking the madry defense model with l\({}_{\mbox{1}}\)-based
  adversarial examples.
\newblock {\em CoRR}, abs/1710.10733, 2017.

\bibitem{DBLP:journals/corr/SimonyanZ14a}
Karen Simonyan and Andrew Zisserman.
\newblock Very deep convolutional networks for large-scale image recognition.
\newblock {\em CoRR}, abs/1409.1556, 2014.

\bibitem{DBLP:conf/nips/SutskeverVL14}
Ilya Sutskever, Oriol Vinyals, and Quoc~V. Le.
\newblock Sequence to sequence learning with neural networks.
\newblock In {\em Advances in Neural Information Processing Systems 27: Annual
  Conference on Neural Information Processing Systems 2014, December 8-13 2014,
  Montreal, Quebec, Canada}, pages 3104--3112, 2014.

\bibitem{DBLP:journals/corr/SzegedyIV16}
Christian Szegedy, Sergey Ioffe, and Vincent Vanhoucke.
\newblock Inception-v4, inception-resnet and the impact of residual connections
  on learning.
\newblock {\em CoRR}, abs/1602.07261, 2016.

\bibitem{wang2018interpret}
Yulong Wang, Hang Su, Bo Zhang, and Xiaolin Hu.
\newblock Interpret neural networks by identifying critical data routing paths.
\newblock In {\em Proceedings of the IEEE Conference on Computer Vision and
  Pattern Recognition}, pages 8906--8914, 2018.

\bibitem{information_bottleneck}
Natalie Wolchover.
\newblock {New Theory Cracks Open the Black Box of Deep Learning}.
\newblock \url{http://bit.ly/information_bottleneck}, 2017.

\bibitem{xie2018mitigating}
Cihang Xie, Jianyu Wang, Zhishuai Zhang, Zhou Ren, and Alan Yuille.
\newblock Mitigating adversarial effects through randomization.
\newblock In {\em International Conference on Learning Representations}, 2018.

\end{thebibliography}
